\PassOptionsToPackage{table,xcdraw}{xcolor}
\documentclass[10pt,sigconf,table]{acmart}
\pdfoutput=1
\usepackage{hyperref}
\usepackage{microtype}

\usepackage{flushend}
\usepackage{array}
\usepackage{times}
\usepackage{comment}
\usepackage{algorithm}
\usepackage{environ}
\usepackage[noend]{algpseudocode}
\usepackage{amssymb}
\usepackage{bm}
\usepackage[font=small,labelfont=bf]{caption}
\usepackage[caption=false,font=scriptsize,
]{subfig}
\usepackage{url}
\usepackage{tikz}
\usepackage[T1]{fontenc}

\usepackage{bbm}

\usepackage{multirow}
\usepackage{hhline}
\usepackage{booktabs}

\usepackage{threeparttable}
\usepackage{xspace} 

\usepackage{color,soulutf8}
\setulcolor{red}

\newcounter{RQCounter}

\NewEnviron{myequation}{%
    \vspace{-10 pt}
    \begin{equation*}
    \scalebox{1.0}{$\BODY$}
    \end{equation*}
    }

\usepackage{setspace} 

\newcommand{\prname}{DeepXplore\xspace}
\newcommand{\ie}{i.e., }
\newcommand{\eg}{e.g., }

\begin{document}\sloppy

\vspace{-5in}
\renewcommand{\textrightarrow}{$\rightarrow$}
\vspace{-1in}

\vspace{-1in}

\setcopyright{acmcopyright}
\acmDOI{10.1145/3132747.3132785}
\acmISBN{978-1-4503-5085-3/17/10}
\copyrightyear{2017} 
\acmPrice{15.00}
\acmConference[SOSP '17]{ACM Symposium on Operating Systems Principles}{October 28, 2017}{Shanghai, China}

\title[\prname]{\prname: Automated Whitebox Testing \\ of Deep Learning Systems}

\author{\vspace{-.2cm}Kexin Pei\texorpdfstring{$^\star$}{*}, Yinzhi Cao\texorpdfstring{$^\dag$}{+}, Junfeng Yang\texorpdfstring{$^\star$}{*}, Suman Jana\texorpdfstring{$^\star$}{*}}

\affiliation{
	\institution{\vspace{-.2cm}\texorpdfstring{$^\star$}{*}Columbia University, \texorpdfstring{$^\dag$}{+}Lehigh University}\vspace{.4cm}    
}

\renewcommand{\shortauthors}{K.~Pei, Y.~Cao, J.~Yang, S.~Jana}
\renewcommand{\authors}{Kexin Pei, Yinzhi Cao, Junfeng Yang, Suman Jana}

\begin{abstract}
\label{sec:abst}
Deep learning (DL) systems are increasingly deployed in safety- and security-critical domains including self-driving cars and malware detection, where the correctness and predictability of a system's behavior for corner case inputs are of great importance. 
Existing DL testing depends heavily on manually labeled data and therefore often fails to expose erroneous behaviors for rare inputs. 

We design, implement, and evaluate \prname, the first whitebox framework for systematically testing real-world DL systems. First, we introduce neuron coverage for systematically measuring the parts of a DL system exercised by test inputs. Next, we leverage multiple DL systems with similar functionality as cross-referencing oracles to avoid manual checking. Finally, we demonstrate how finding inputs for DL systems that both trigger many differential behaviors and achieve high neuron coverage can be represented as a joint optimization problem and solved efficiently using gradient-based search techniques.

\prname efficiently finds thousands of incorrect corner case behaviors (e.g., self-driving cars crashing into guard rails and malware masquerading as benign software) in state-of-the-art DL models with thousands of neurons trained on five popular datasets including ImageNet and Udacity self-driving challenge data. For all tested DL models, on average, \prname generated one test input demonstrating incorrect behavior within one second while running only on a commodity laptop. 
We further show that the test inputs generated by \prname can also be used to retrain the corresponding DL model to improve the model's accuracy by up to $3$\%.  
\end{abstract}

\maketitle

\vspace{-0.4cm}
\section{Introduction}
\label{sec:intro}
Over the past few years, Deep Learning (DL) has made tremendous progress, achieving or surpassing human-level performance for a diverse set of tasks including image classification~\cite{he2015deep, simonyan2014very}, speech recognition~\cite{xiong2016achieving}, and playing games such as Go~\cite{silver2016mastering}. These advances have led to widespread adoption and deployment of DL in security- and safety-critical systems such as self-driving cars~\cite{bojarski2016end}, malware detection~\cite{yuan2014droid}, and aircraft collision avoidance systems~\cite{julian2016policy}. 

This wide adoption of DL techniques presents new challenges as the predictability and correctness of such systems are of crucial importance. Unfortunately, DL systems, despite their impressive capabilities, often demonstrate unexpected or incorrect behaviors in corner cases for several reasons such as biased training data, overfitting, and underfitting of the models. In safety- and security-critical settings, such incorrect behaviors can lead to disastrous consequences such as a fatal collision of a self-driving car. For example, a Google self-driving car recently crashed into a bus because it expected the bus to yield under a set of rare conditions but the bus did not~\cite{google-accident}.  A Tesla car in autopilot crashed into a trailer because the autopilot system failed to recognize the trailer as an obstacle due to its ``white color against a brightly lit sky'' and the ``high ride height''~\cite{tesla-accident}. Such corner cases were not part of Google's or Tesla's test set and thus never showed up during testing.

Therefore, safety- and security-critical DL systems, just like traditional software, must be tested systematically for different corner cases to detect and fix ideally any potential flaws or undesired behaviors. This presents a new systems problem as automated and systematic testing of large-scale, real-world DL systems with thousands of neurons and millions of parameters for all corner cases is extremely challenging.

The standard approach for testing DL systems is to gather and manually label as much real-world test data as possible ~\cite{waymo-report, imagenet-crowdsource}. Some DL systems such as Google self-driving cars also use simulation to generate synthetic training data~\cite{waymo_simulation}. However, such simulation is completely unguided as it does not consider the internals of the target DL system. Therefore, for the large input spaces of real-world DL systems (e.g., all possible road conditions for a self-driving car), none of these approaches can hope to cover more than a tiny fraction (if any at all) of all possible corner cases. 

Recent works on adversarial deep learning~\cite{szegedy2013intriguing,goodfellow2014explaining,nguyen2015deep} have demonstrated that carefully crafted synthetic images by adding minimal perturbations to an existing image can fool state-of-the-art DL systems. The key idea is to create synthetic images such that they get classified by DL models differently than the original picture but still look the same to the human eye. While such adversarial images expose some erroneous behaviors of a DL model, the main restriction of such an approach is that it must limit its perturbations to tiny invisible changes or require manual checks. Moreover, just like other forms of existing DL testing, the adversarial images only cover a small part (52.3\%) of DL system's logic as shown in \S~\ref{sec:evaluation}. In essence, the current machine learning testing practices for finding incorrect corner cases are analogous to finding bugs in traditional software by using test inputs with low code coverage and thus are unlikely to find many erroneous cases.

 
The key challenges in automated systematic testing of large-scale DL systems are twofold: (1) how to generate inputs that trigger different parts of a DL system's logic and uncover different types of erroneous behaviors, and (2) how to identify erroneous behaviors of a DL system without manual labeling/checking. This paper describes how we design and build \prname to address both challenges.

First, we introduce the concept of neuron coverage for measuring the parts of a DL system's logic exercised by a set of test inputs based on the number of neurons activated (i.e., the output values are higher than a threshold) by the inputs.  At a high level, neuron coverage of DL systems is similar to code coverage of traditional systems, a standard empirical metric for measuring the amount of code exercised by an input in a traditional software. However, code coverage itself is not a good metric for estimating coverage of DL systems as most rules in DL systems, unlike traditional software, are not written manually by a programmer but rather are learned from training data. In fact, we find that for most of the DL systems that we tested, even a single randomly picked test input was able to achieve 100\% code coverage while the neuron coverage was less than 10\%. 

Next, we show how multiple DL systems with similar functionality (e.g., self-driving cars by Google, Tesla, and GM) can be used as cross-referencing oracles to identify erroneous corner cases without manual checks. For example, if one self-driving car decides to turn left while others turn right for the same input, one of them is likely to be incorrect. Such differential testing techniques have been applied successfully in the past for detecting logic bugs without manual specifications in a wide variety of traditional software ~\cite{frankencert, mucert, mckeeman1998differential, yang2011finding, sfadiff, chen_2016_pldi_difftest_jvm}. In this paper, we demonstrate how differential testing can be applied to DL systems. 

\begin{figure}[!tbp]
\centering
\captionsetup[subfloat]{captionskip=.5pt}
 \vspace{-.3cm}
\subfloat[Input 1]{
\includegraphics[width=0.46\columnwidth, height=0.46\columnwidth]{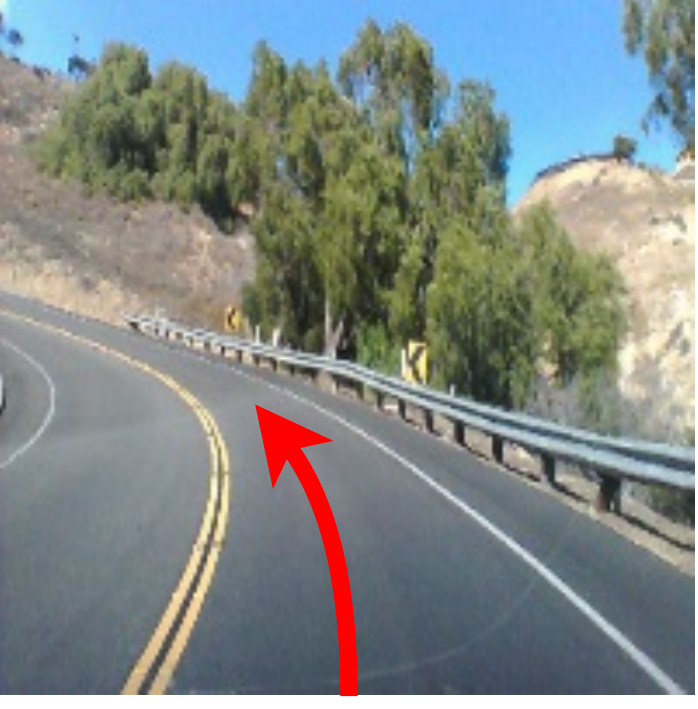}
\label{subfig:wrongcar_orig}}
\hfill
\subfloat[Input 2 (darker version of 1)]{
\includegraphics[width=0.46\columnwidth, height=0.46\columnwidth]{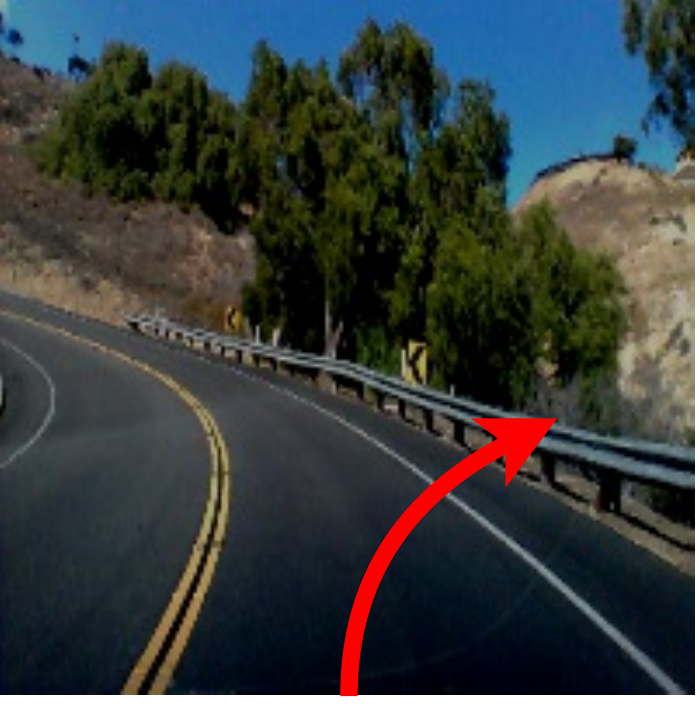}
\label{subfig:wrongcar}}
\caption{An example erroneous behavior found by \prname in Nvidia DAVE-2 self-driving car platform. The DNN-based self-driving car correctly decides to turn left for image (a) but incorrectly decides to turn right and crashes into the guardrail for image (b), a slightly darker version of (a).}
\label{fig:wrongcar}
\vspace{-.2cm}
\end{figure}

Finally, we demonstrate how the problem of generating test inputs that maximize neuron coverage of a DL system while also exposing as many differential behaviors (i.e., differences between multiple similar DL systems) as possible can be formulated as a joint optimization problem. Unlike traditional programs, the functions approximated by most popular Deep Neural Networks (DNNs) used by DL systems are differentiable. Therefore, their gradients with respect to inputs can be calculated accurately given whitebox access to the corresponding model. In this paper, we show how these gradients can be used to efficiently solve the above-mentioned joint optimization problem for large-scale real-world DL systems. 

We design, implement, and evaluate \prname, to the best of our knowledge, the first efficient whitebox testing framework for large-scale DL systems. In addition to maximizing neuron coverage and behavioral differences between DL systems, \prname also supports adding custom constraints by the users for simulating different types of realistic inputs (e.g., different types of lighting and occlusion for images/videos). We demonstrate that \prname efficiently finds thousands of unique incorrect corner case behaviors (e.g., self-driving cars crashing into guard rails) in $15$ state-of-the-art DL models trained using five real-world datasets including Udacity self-driving car challenge data, image data from ImageNet and MNIST, Android malware data from Drebin, and PDF malware data from Contagio/VirusTotal. For all of the tested DL models, on average, \prname generated one test input demonstrating incorrect behavior within one second while running on a commodity laptop. The inputs generated by \prname achieved 34.4\% and 33.2\% higher neuron coverage on average than the same number of randomly picked inputs and adversarial inputs~\cite{szegedy2013intriguing, goodfellow2014explaining,nguyen2015deep} respectively. We further show that the test inputs generated by \prname can be used to retrain the corresponding DL model to improve classification accuracy as well as identify potentially polluted training data. We achieve up to 3\% improvement in classification accuracy by retraining a DL model on inputs generated by \prname compared to retraining on the same number of random or adversarial inputs.

Our main contributions are:
\begin{itemize}
\item
We introduce neuron coverage as the first whitebox testing metric for DL systems that can estimate the amount of DL logic explored by a set of test inputs.
\item
We demonstrate that the problem of finding a large number of behavioral differences between similar DL systems while maximizing neuron coverage can be formulated as a joint optimization problem. We present a gradient-based algorithm for solving this problem efficiently.
\item
We implement all of these techniques as part of \prname, the first whitebox DL-testing framework that exposed thousands of incorrect corner case behaviors (e.g., self-driving cars crashing into guard rails as shown in Figure~\ref{fig:wrongcar}) in $15$ state-of-the-art DL models with a total of $132,057$ neurons trained on five popular datasets containing around $162$ GB of data.  
\item 
We show that the tests generated by \prname can also be used to retrain the corresponding DL systems to improve classification accuracy by up to 3\%. 
\end{itemize}

\section{Background}

\subsection{DL Systems}
\label{subsec:dl_systems}
We define a DL system to be any software system that includes at least one Deep Neural Network (DNN)  component. Note that some DL systems might comprise solely of DNNs (e.g., self-driving car DNNs predicting steering angles without any manual rules) while others may have some DNN components interacting with other traditional software components to produce the final output.

\begin{figure}[!t]
\captionsetup[subfigure]{labelformat=empty}
\centering
\subfloat[Traditional software development]{
\includegraphics[width=0.42\columnwidth,height=3cm]{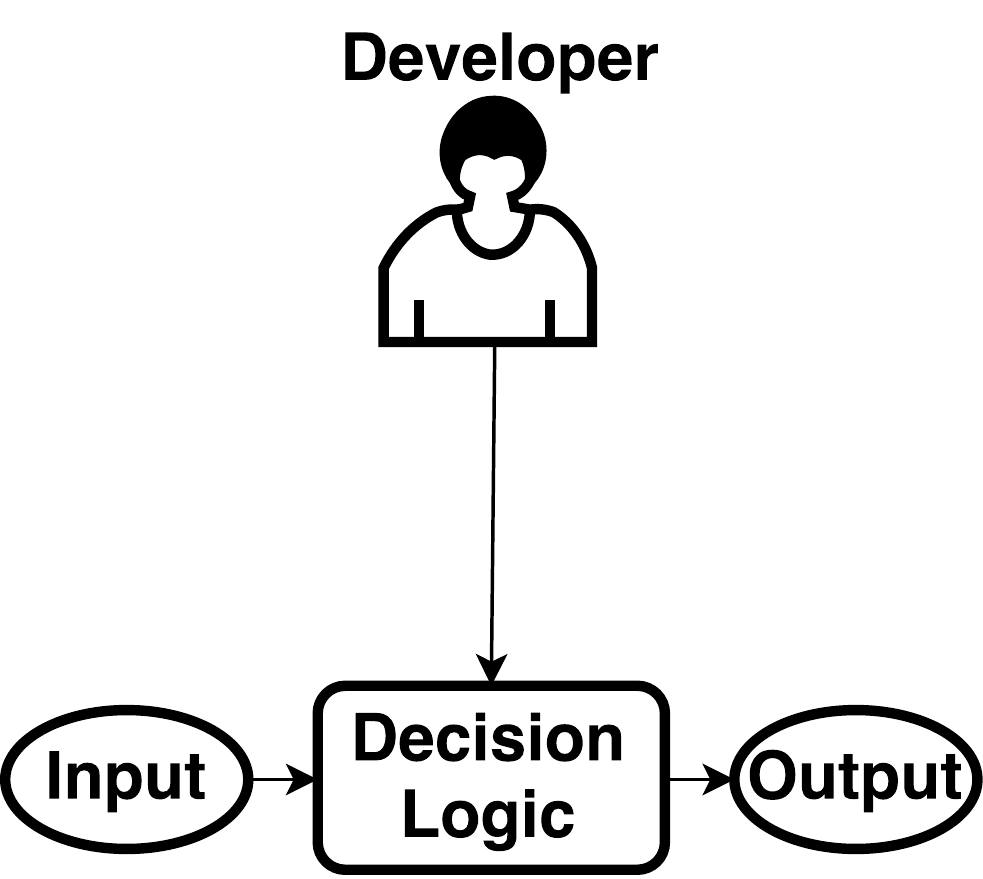}
\label{fig:compare1}}
\subfloat[ML system development]{
\includegraphics[width=0.54\columnwidth,height=3.2cm]{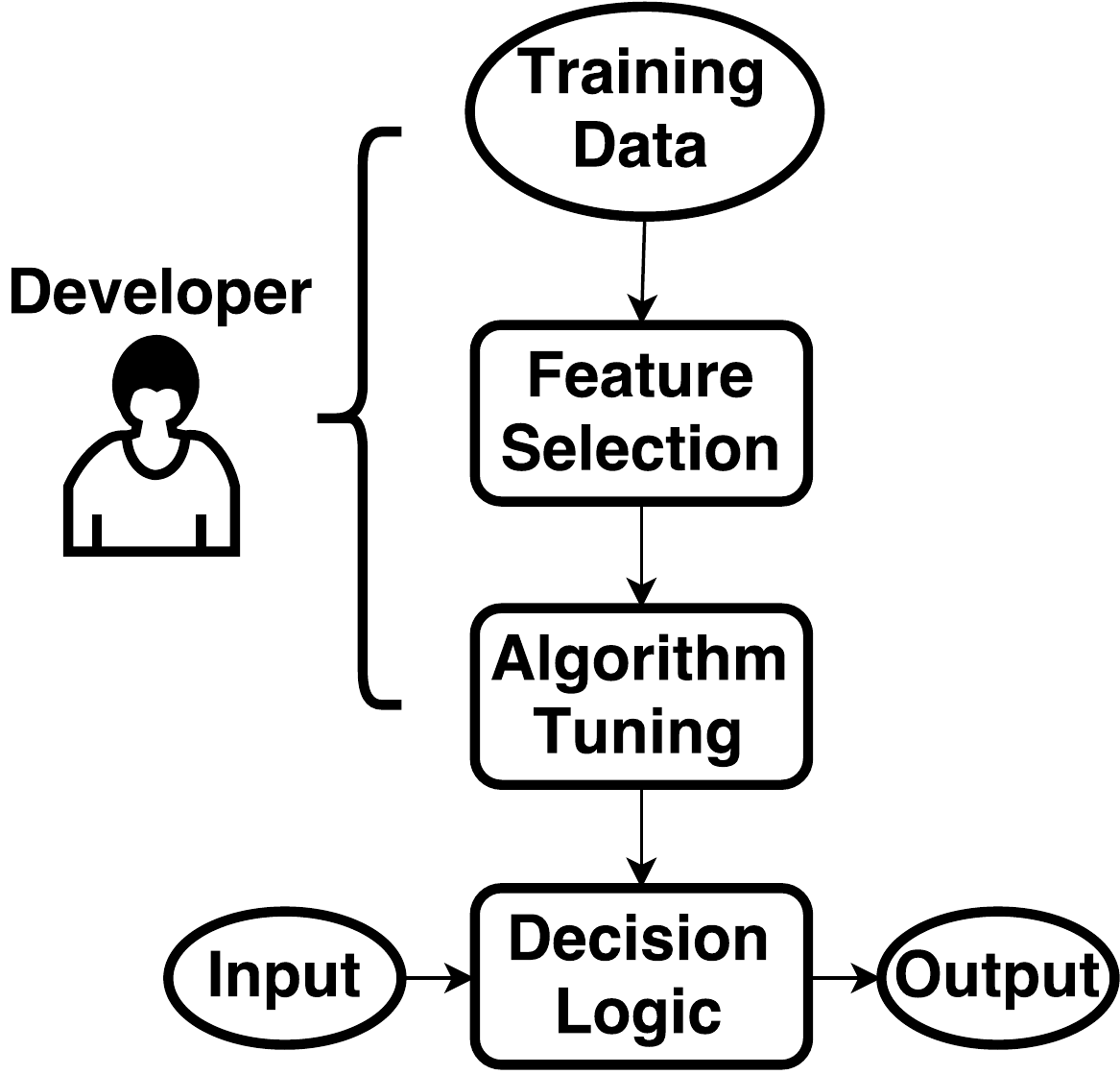}
\label{fig:compare2}}
\caption{Comparison between traditional and ML system development processes.
}
\label{fig:compare}
\vspace{-.1cm}
\end{figure}

The development process of the DNN components of a DL system is fundamentally different from traditional software development.  Unlike traditional software, where the developers directly specify the logic of the system, the DNN components learn their rules automatically from data. The developers of DNN components can indirectly influence the rules learned by a DNN by modifying the training data, features, and the model's architectural details (e.g., number of layers) as shown in Figure~\ref{fig:compare}.  

As a DNN's rules are mostly unknown even to its developers, testing and fixing of erroneous behaviors of DNNs are crucial in safety-critical settings. In this paper, we primarily focus on automatically finding inputs that trigger erroneous behaviors in DL systems and provide preliminary evidence about how these inputs can be used to fix the buggy behavior by augmenting or filtering the training data in \S~\ref{subsec:other_usage}.

\vspace{-.2cm}
\subsection{DNN Architecture}
DNNs are inspired by human brains with millions of interconnected neurons. They are known for  their amazing ability to automatically identify and extract the relevant high-level features from raw inputs without any human guidance besides labeled training data. In recent years, DNNs have surpassed human performance in many application domains due to increasing availability of large datasets~\cite{deng2009imagenet, krizhevsky2009learning, miller1995wordnet}, specialized hardware~\cite{nvidia2008programming, tpu}, and  efficient training algorithms~\cite{krizhevsky2012imagenet, simonyan2014very, szegedy2015going, he2015deep}.

\begin{figure}[!bp]
\centering
\includegraphics[width=0.99\columnwidth]{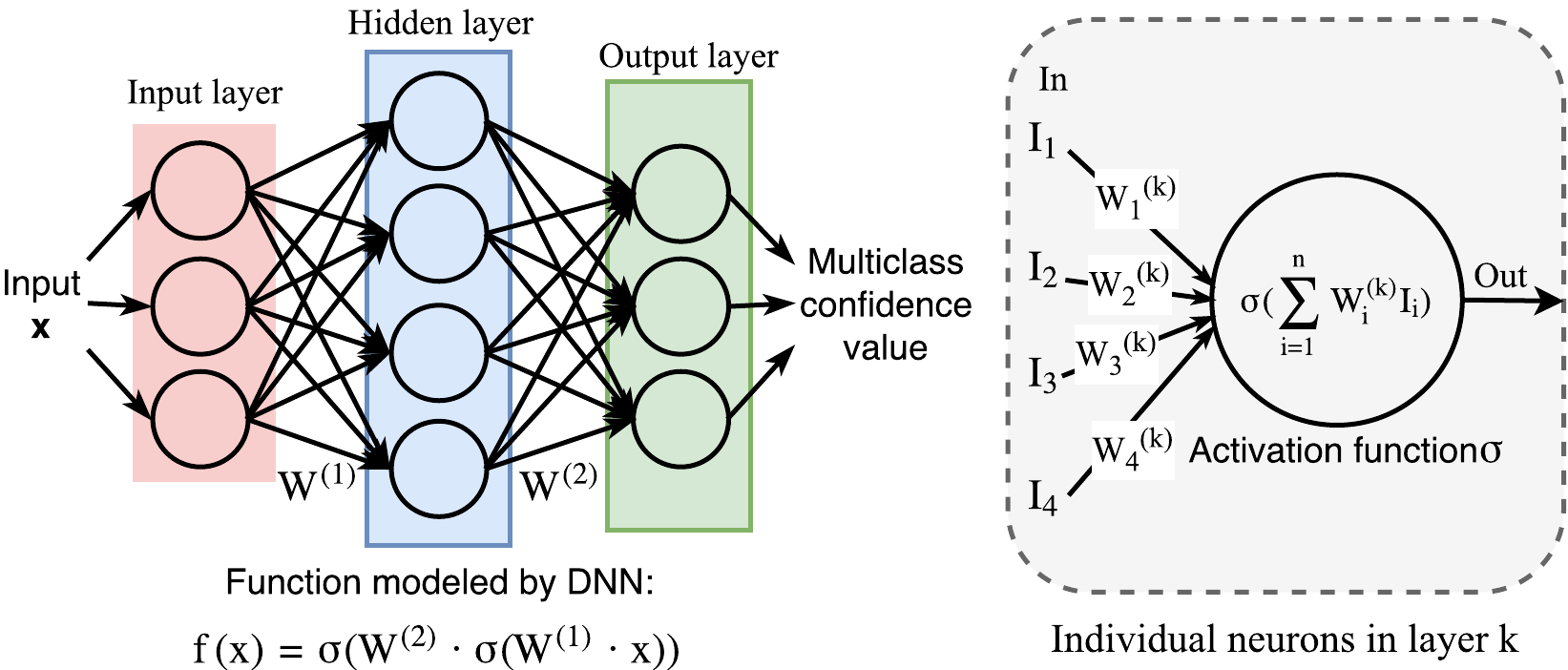}
\caption{A simple DNN and the computations performed by each of its neurons.}
\vspace{-.1cm}
\label{fig:DNN_illustration}
\end{figure}

A DNN consists of multiple \textit{layers}, each containing multiple \textit{neurons} as shown in Figure~\ref{fig:DNN_illustration}.
A \textit{neuron} is an individual computing unit inside a DNN that applies an \textit{activation function} on its inputs and passes the result to other connected neurons (see Figure~\ref{fig:DNN_illustration}). The common activation functions include sigmoid, hyperbolic tangent, or ReLU (Rectified Linear Unit)~\cite{nair2010rectified}. A DNN usually has at least three (often more) layers: one input, one output, and one or more hidden layers. Each neuron in one layer has directed connections to the neurons in the next layer. The numbers of neurons in each layer and the connections between them vary significantly across DNNs. Overall, a DNN can be defined mathematically as a multi-input, multi-output parametric function $F$ composed of many parametric sub-functions representing different neurons.

Each connection between the neurons in a DNN is bound to a \textit{weight} parameter characterizing the strength of the connection between the neurons. For supervised learning, the weights of the connections are learned during training by minimizing a cost function over the training data. DNNs can be trained using different training algorithms, but gradient descent using backpropagation is by far the most popular training algorithm for DNNs~\cite{rumelhart1988learning}.

\begin{figure}[!t]
\centering
\captionsetup[subfloat]{captionskip=2pt}
\subfloat[A program with a rare branch]{
\includegraphics[width=0.47\columnwidth]{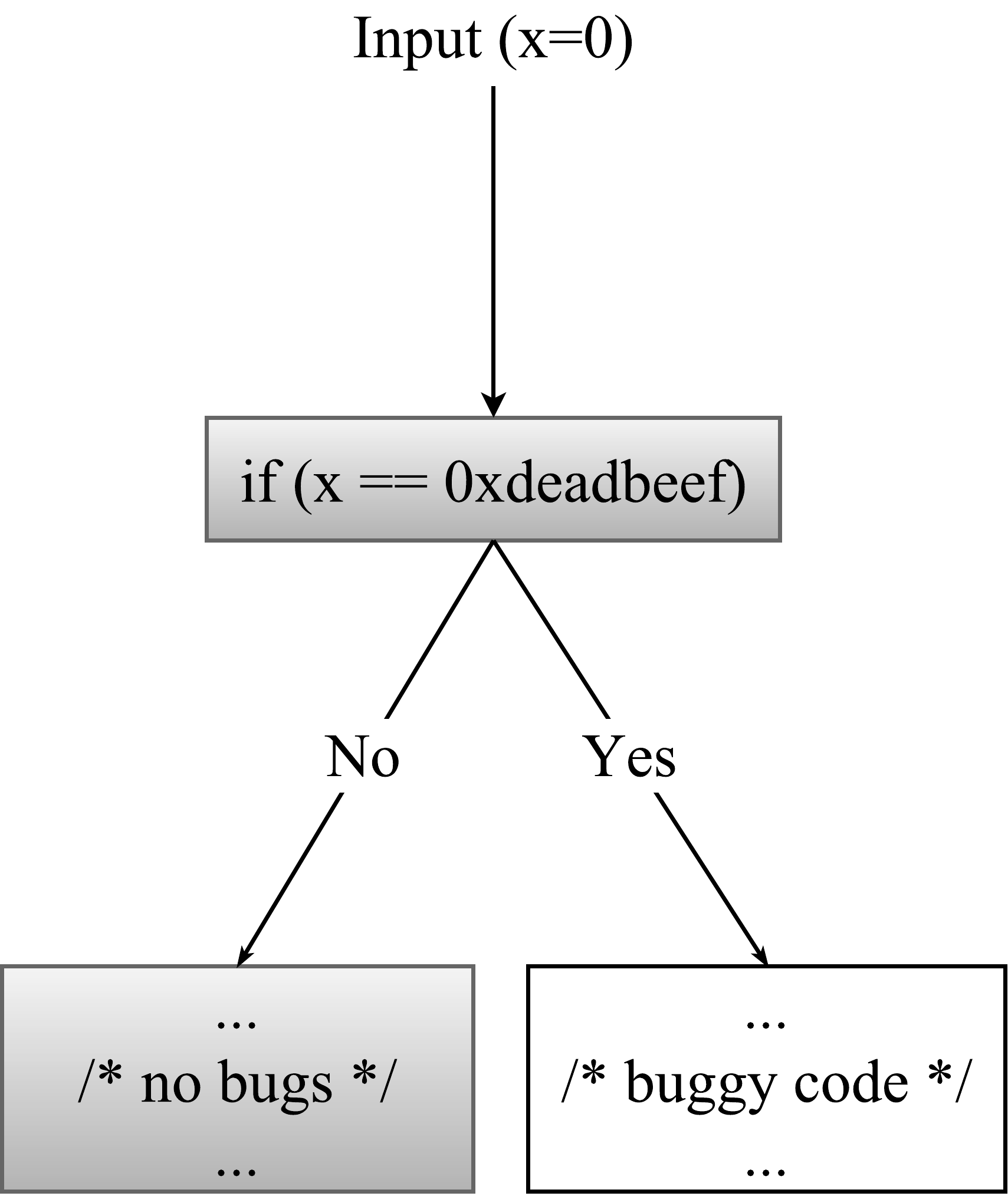}
\label{subfig:s_flow}}
\hfill
\subfloat[A DNN for detecting cars and faces]{
\includegraphics[width=0.47\columnwidth, height=4.8cm]{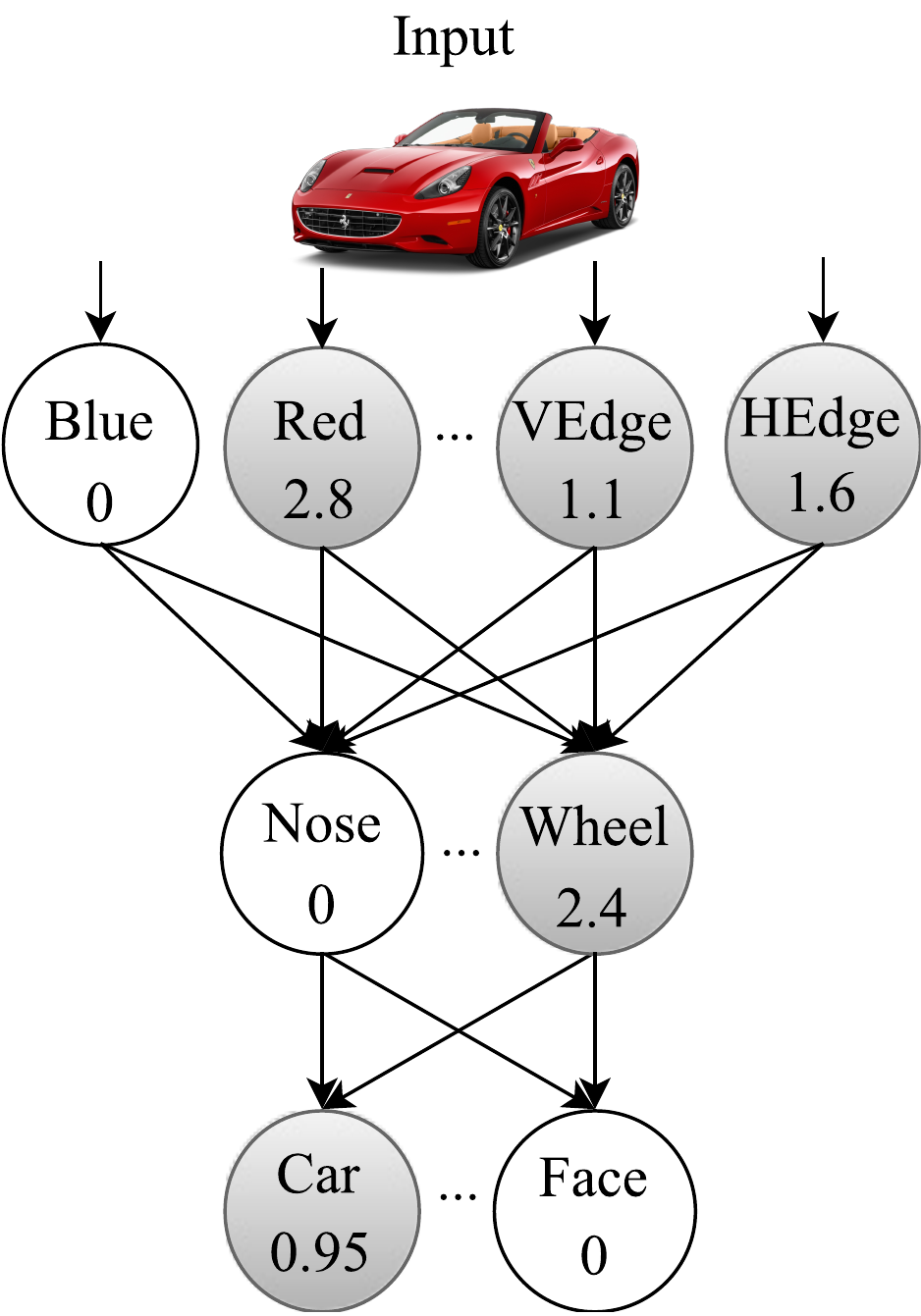}
\label{subfig:nn_flow}}
\caption{Comparison between program flows of a traditional program and a neural network. The nodes in gray denote the corresponding basic blocks or neurons that participated while processing an input.}
\label{fig:comparison}
\vspace{-.4cm}
\end{figure}

Each layer of the network transforms the information contained in its input to a higher-level representation of the data.
For example, consider a pre-trained network shown in Figure~\ref{subfig:nn_flow} for classifying images into two categories: human faces and cars. The first few hidden layers transform the raw pixel values into low-level texture features like edges or colors and feed them to the deeper layers~\cite{yosinskiunderstanding}.
The last few layers, in turn, extract and assemble the meaningful high-level abstractions like noses, eyes, wheels, and headlights to make the classification decision.

\vspace{-.2cm}
\subsection{Limitations of Existing DNN Testing}
\label{sec:limitation}
\noindent \textbf{Expensive labeling effort.} Existing DNN testing techniques require prohibitively expensive human effort to provide correct labels/actions for a target task (e.g., self-driving a car, image classification, and malware detection). For complex and high-dimensional real-world inputs, human beings, even domain experts, often have difficulty in efficiently performing a task correctly for a large dataset. For example, consider a DNN designed to identify potentially malicious executable files.  Even a security professional will have trouble determining whether an executable is malicious or benign without executing it. However, executing and monitoring a malware inside a sandbox incur significant performance overhead and therefore makes manual labeling significantly harder to scale to a large number of inputs. 

\noindent \textbf{Low test coverage.} None of the existing DNN testing schemes even try to cover different rules of the DNN. Therefore, the test inputs often fail to uncover different erroneous behaviors of a DNN.   

For example, DNNs are often tested by simply dividing a whole dataset into two random parts---one for training and the other for testing. The testing set in such cases may only exercise a small subset of all rules learned by a DNN. Recent results involving adversarial evasion attacks against DNNs have demonstrated the existence of some corner cases where DNN-based image classifiers (with state-of-the-art performance on randomly picked testing sets) still incorrectly classify synthetic images generated by adding humanly imperceptible perturbations to a test image ~\cite{adversarial:ccs16,6956565,xu2016automatically,grosse2016adversarial,goodfellow2014explaining, papernot2016limitations}. However, the adversarial inputs, similar to random test inputs, also only cover a small part the rules learned by a DNN as they are not designed to maximize coverage.  Moreover, they are also inherently limited to small imperceptible perturbations around a test input as larger perturbations will visually change the input and therefore will require manual inspection to ensure correctness of the DNN's decision.

\noindent \textbf{Problems with low-coverage DNN tests.} To better understand the problem of low test coverage of rules learned by a DNN, we provide an analogy to a similar problem in testing traditional software.  
 Figure~\ref{fig:comparison} shows a side-by-side comparison of how a traditional program and a DNN handle inputs and produce outputs.  Specifically, the figure shows the \textit{similarity between traditional software and DNNs}: in software program, each statement performs a certain operation to transform the output of previous statement(s) to the input to the following statement(s), while in DNN, each neuron transforms the output of previous neuron(s) to the input of the following neuron(s).  Of course, unlike traditional software,  DNNs do not have explicit branches but a neuron's influence on the downstream neurons
decreases as the neuron's output value gets lower. A lower output value indicates less influence and vice versa. When the output value of a neuron becomes zero, the neuron does not have any influence on the downstream neurons. 
%

As demonstrated in Figure~\ref{subfig:s_flow}, the problem of low coverage in testing traditional software is obvious. In this case, the buggy behavior will never be seen unless the test input is \texttt{0xdeadbeef}. The chances of randomly picking such a value is very small. Similarly, low-coverage test inputs will also leave different behaviors of DNNs unexplored. For example, consider a simplified neural network, as shown in Figure~\ref{subfig:nn_flow}, that takes an image as input and classifies it into two different classes: cars and faces. The text in each neuron (represented as a node) denotes the object or property that the neuron detects\footnote{\scriptsize Note that one cannot always map each neuron to a particular task, i.e., detecting specific objects/properties. Figure~\ref{subfig:nn_flow} simply highlights that different neurons often tend to detect different features.}, and the number in each neuron is the real value outputted by that neuron. The number indicates how confident the neuron is about its output. Note that randomly picked inputs are highly unlikely to set high output values for the unlikely combination of neurons. Therefore, many incorrect DNN behaviors will remain unexplored even after performing a large number of random tests. For example, if an image causes neurons labeled as ``Nose'' and ``Red'' to produce high output values and the DNN misclassifies the input image as a car, such a behavior will never be seen during regular testing as the chances of an image containing a red nose (e.g., a picture of a clown) is very small.

\section{Overview}
\label{sec:overview}





\begin{figure}[!tbp]
\centering
\includegraphics[width=\columnwidth]{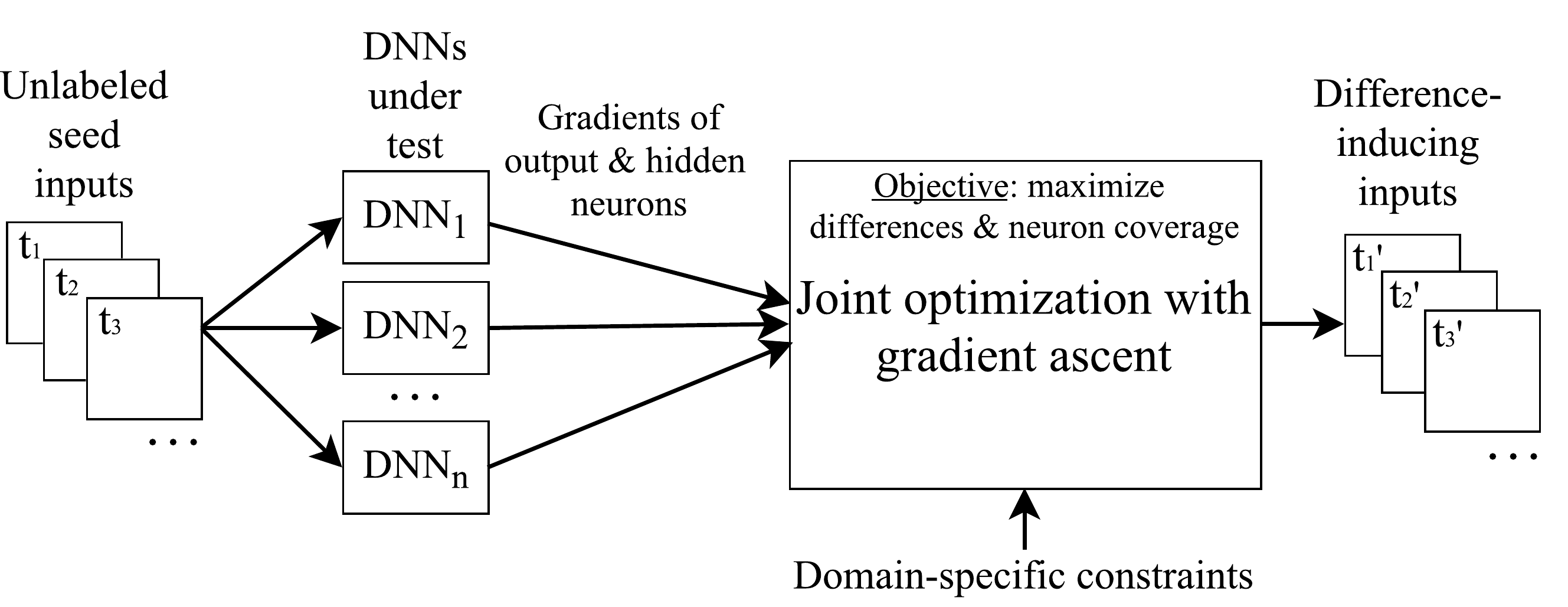}
\caption{\prname workflow.}
\label{fig:workflow}
\end{figure}

In this section, we provide a general overview of \prname, our whitebox framework for systematically testing DNNs for erroneous corner case behaviors. The main components of \prname are shown in Figure~\ref{fig:workflow}. \prname takes unlabeled test inputs as seeds and generates new tests that cover a large number of neurons (\ie activates them to a value above a customizable threshold) while causing the tested DNNs to behave differently.  Specifically, \prname solves a joint optimization problem that maximizes both differential behaviors and neuron coverage. Note that both goals are crucial for thorough testing of DNNs and finding diverse erroneous corner case behaviors. High neuron coverage alone may not induce many erroneous behaviors while just maximizing different behaviors might simply identify different manifestations of the same underlying root cause.

\prname also supports enforcing of custom domain-specific constraints as part of the joint optimization process. For example, the value of an image pixel has to be between 0 and 255. Such domain-specific constraints can be specified by the users of \prname to ensure that the generated test inputs are valid and realistic.

We designed an algorithm for efficiently solving the joint optimization problem mentioned above using gradient ascent. First, we compute the gradient of the \textit{outputs} of the neurons in both the output and hidden layers with the \textit{input value} as a variable and the \textit{weight parameter} as a constant. Such gradients can be computed efficiently for most DNNs. Note that \prname is designed to operate on pre-trained DNNs. The gradient computation is efficient because our whitebox approach has access to the pre-trained DNNs' weights and the intermediate neuron values. Next, we iteratively perform gradient ascent to modify the test input toward maximizing the objective function of the joint optimization problem described above. Essentially, we perform a gradient-guided local search starting from the seed inputs and find new inputs that maximize the desired goals. Note that, at a high level, our gradient computation is similar to the backpropagation performed during the training of a DNN, but the key difference is that, unlike our algorithm, backpropagation treats the \textit{input value} as a constant and the \textit{weight parameter} as a variable. 

\begin{figure}[!bp]
\centering
\captionsetup[subfloat]{captionskip=3pt}
\subfloat[DNNs produce same output]{
\includegraphics[width=0.46\columnwidth]{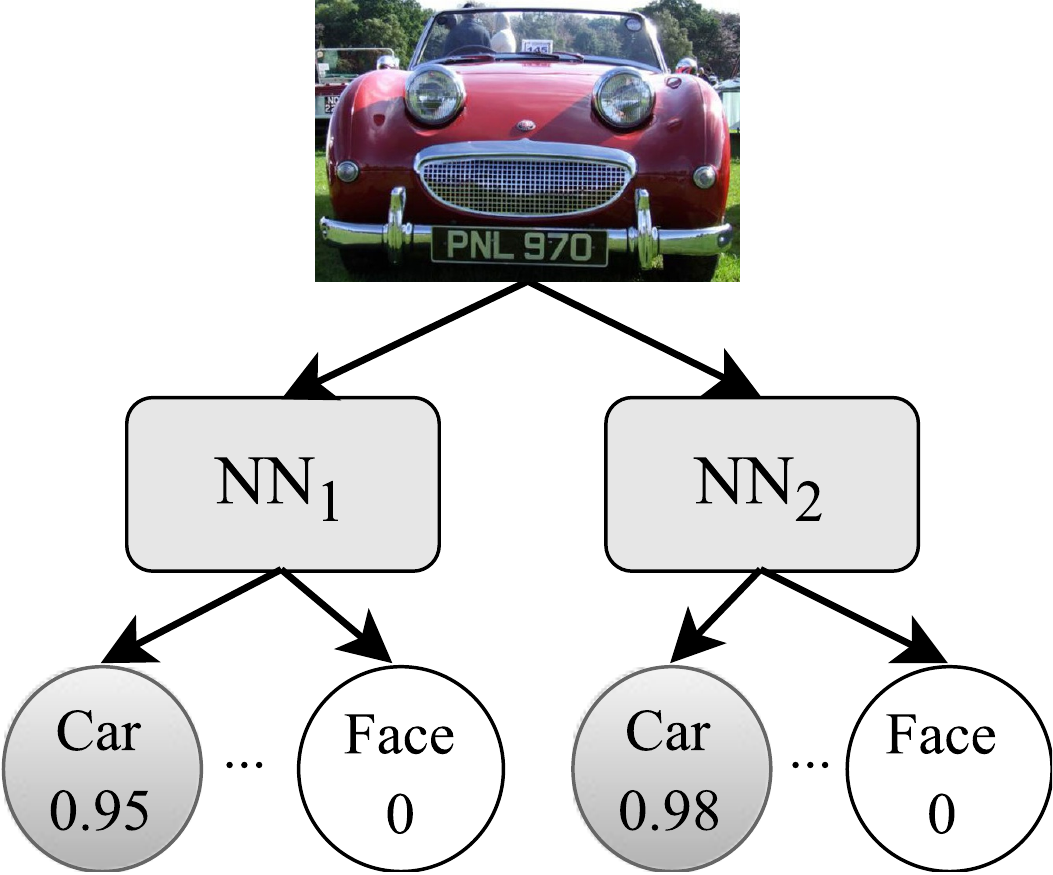}
\label{subfig:same_output}}
\hfill
\subfloat[DNNs produce different output]{
\includegraphics[width=0.46\columnwidth]{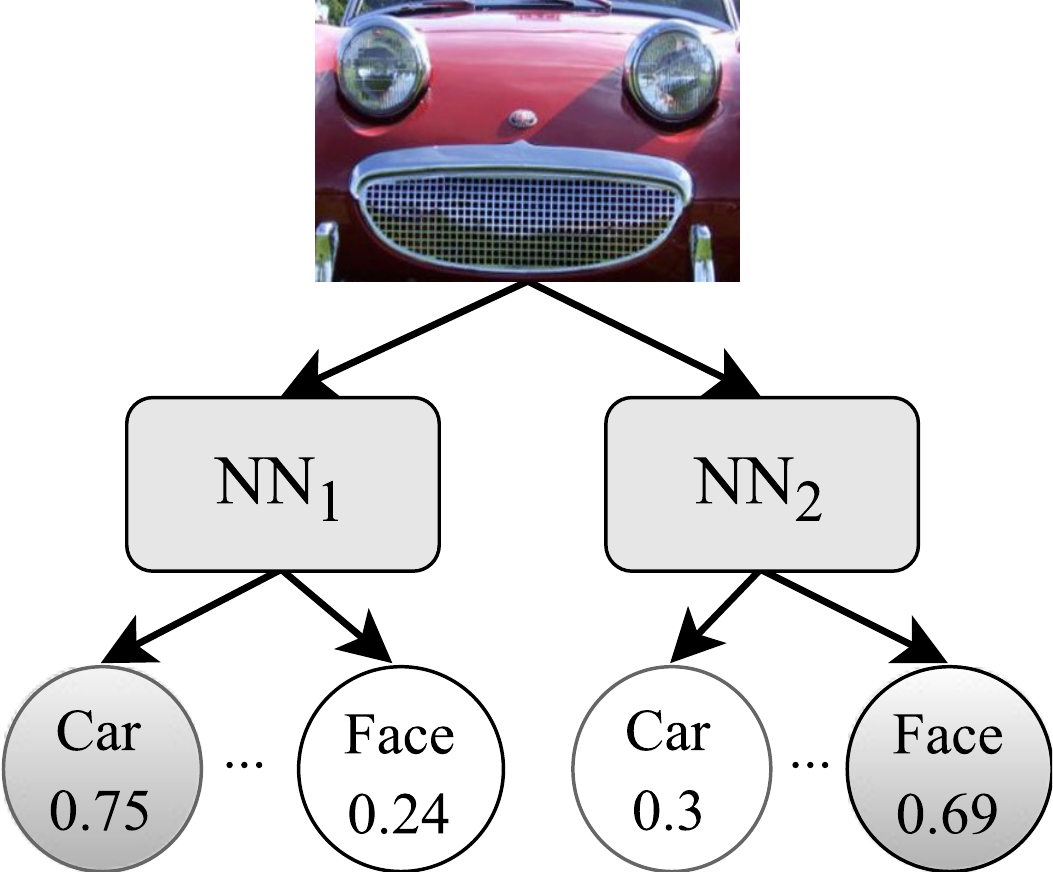}
\label{subfig:diff_output}}
\caption{Inputs inducing different behaviors in two similar DNNs.}
\label{fig:comparison_differential}
\end{figure}

\noindent{\bf A working example.}  We use Figure~\ref{fig:comparison_differential} as an example to show how \prname generates test inputs. Consider that we have two DNNs to test---both perform similar tasks, i.e., classifying images into cars or faces, as shown in Figure~\ref{fig:comparison_differential}, but they are trained independently with different datasets and parameters. Therefore, the DNNs will learn similar but slightly different classification rules. Let us also assume that we have a seed test input, the image of a red car, which both DNNs identify as a car as shown in Figure~\ref{subfig:same_output}. 



\prname tries to maximize the chances of finding differential behavior by modifying the input, \ie the image of the red car, towards maximizing its probability of being classified as a car by one DNN but minimizing corresponding probability of the other DNN. \prname also tries to cover as many neurons as possible by activating (\ie causing a neuron's output to have a value greater than a threshold) inactive neurons in the hidden layer. We further add domain-specific constraints (\eg ensure the pixel values are integers within 0 and 255 for image input) to make sure that the modified inputs still represent real-world images. The joint optimization algorithm will iteratively perform a gradient ascent to find a modified input that satisfies all of the goals described above. \prname will eventually generate a set of test inputs where the DNNs' outputs differ, \eg one DNN thinks it is a car while the other thinks it is a face as shown in Figure~\ref{subfig:diff_output}. 

\begin{figure}[!t]
\centering
\includegraphics[width=0.8\columnwidth]{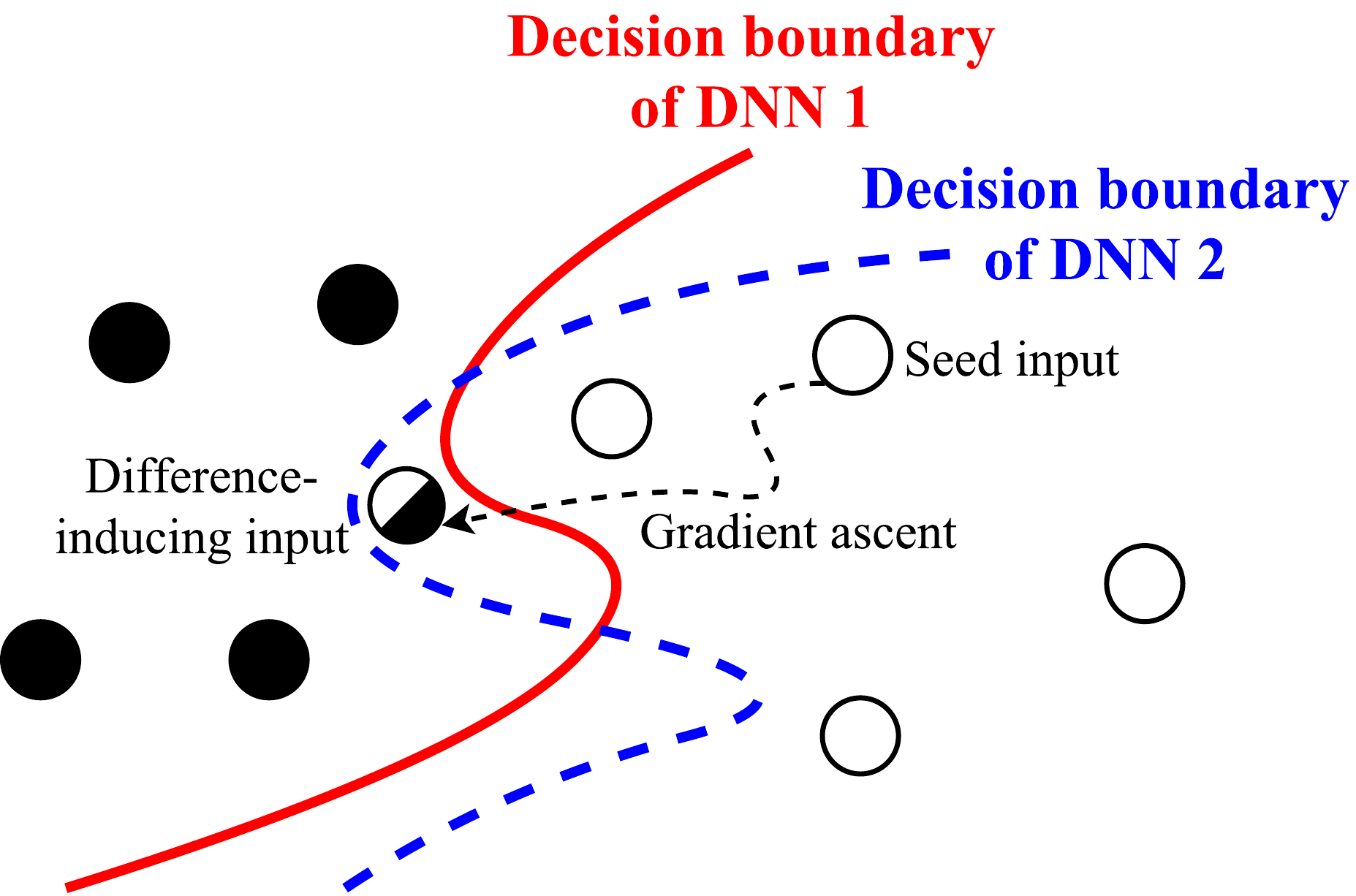}
\caption{Gradient ascent starting from a seed input and gradually finding the difference-inducing test inputs.}
\label{fig:gradient_ascent}
\end{figure}

Figure~\ref{fig:gradient_ascent} illustrates the basic concept of our technique using gradient ascent. Starting from a seed input, \prname performs the guided search by the gradient in the input space of two similar DNNs supposed to perform the same task such that it finally uncovers the test inputs that lie between the decision boundaries of these DNNs. Such test inputs will be classified differently by the two DNNs. Note that while the gradient provides the rough direction toward reaching the goal (\eg finding difference-inducing inputs), it does not guarantee the fastest convergence. Thus as shown in Figure~\ref{fig:gradient_ascent}, the gradient ascent process often does not follow a straight path towards reaching the target.

\section{Methodology}
\label{sec:approach}
In this section, we provide a detailed technical description of our algorithm. First, we define and explain the concepts of neuron coverage and gradient for DNNs. Next, we describe how the testing problem can be formulated as a joint optimization problem. Finally, we provide the gradient-based algorithm for solving the joint optimization problem.

\vspace{-0.2cm}
\subsection{Definitions}
\label{subsec:def}
\noindent
{\bf Neuron coverage.}
We define neuron coverage of a set of test inputs as the ratio of the number of unique activated neurons for all test inputs and the total number of neurons in the DNN. We consider a neuron to be activated if its output is higher than a threshold value (\eg $0$).

More formally, let us assume that all neurons of a DNN are represented by the set $N=\{n_1,n_2,...\}$, all test inputs are represented by the set $T=\{\bm{x_1},\bm{x_2},...\}$, and $out(n,\bm{x})$ is a function that returns the output value of neuron $n$ in the DNN for a  given test input $\bm{x}$. 
Note that the bold $\bm{x}$ signifies that $\bm{x}$ is a vector.
Let $t$ represent the threshold for considering a neuron to be activated. In this setting, neuron coverage can be defined as follows.

\begin{myequation}\footnotesize
    NCov(T, \bm{x}) = \frac{|\{n \mid \forall \bm{x} \in T, out(n,\bm{x})>t\}|}{|N|} 
\end{myequation}

To demonstrate how neuron coverage is calculated in practice, consider the DNN showed in Figure~\ref{subfig:nn_flow}. The neuron coverage (with threshold $0$) for the input picture of the red car shown in Figure~\ref{subfig:nn_flow} will be $5/8=0.625$. 

\noindent{\bf Gradient.}
The gradients or forward derivatives of the outputs of neurons of a DNN with respect to the input are well known in deep learning literature. They have been extensively used both for crafting adversarial examples~\cite{grosse2016adversarial, goodfellow2014explaining, papernot2016limitations, szegedy2013intriguing} and visualizing/understanding DNNs~\cite{yosinskiunderstanding, simonyan2013deep, mahendran2015understanding}. We provide a brief definition here for completeness and refer interested readers to~\cite{yosinskiunderstanding} for more details.  

Let $\bm{\theta}$ and $\bm{x}$ represent the parameters and the test input of a DNN respectively. The parametric function performed by a neuron can be represented as $y=f(\bm{\theta}, \bm{x})$ where $f$ is a function that takes $\bm{\theta}$ and $\bm{x}$ as input and output $y$. Note that $y$ can be the output of any neuron defined in the DNN (\eg neuron from output layer or intermediate layers). The gradient of $f(\bm{\theta}, \bm{x})$ with respect to input $\bm{x}$ can be defined as:
\begin{equation}\footnotesize
\label{eq:jcobian}
\begin{split}
    \bm{G}=\nabla_{\bm{x}}f(\bm{\theta}, \bm{x})={\partial y}/{\partial \bm{x}}
\end{split}
\end{equation}

The computation inside $f$ is essentially a sequence of stacked functions that compute the input from previous layers and forward the output to next layers. Thus, $\bm{G}$ can be calculated by utilizing the chain rule in calculus, \ie by computing the layer-wise derivatives starting from the layer of the neuron that outputs $y$ until reaching the input layer that takes $\bm{x}$ as input. Note that the dimension of the gradient $\bm{G}$ is identical to that of the input $\bm{x}$.

\vspace{-0.2cm}
\subsection{\prname algorithm}
\label{subsec:algo}
The main advantage of the test input generation process for a DNN over traditional software is that the test generation process, once defined as an optimization problem, can be solved efficiently using gradient ascent. In this section, we describe the details of the formulation and finding solutions to the optimization problem. Note that solutions to the optimization problem can be efficiently found for DNNs as the gradients of the objective functions of DNNs, unlike traditional software, can be easily computed. 

As discussed earlier in \S~\ref{sec:overview}, the objective of the test generation process is to maximize both the number of observed differential behaviors and the neuron coverage while preserving domain-specific constraints provided by the users. Algorithm~\ref{alg:testgen} shows the algorithm for generating test inputs by solving this joint optimization problem. Below, we define the objectives of our joint optimization problem formally and explain the details of the algorithm for solving it.

\vspace{2pt}
\noindent
{\bf Maximizing differential behaviors.}
The first objective of the optimization problem is to generate test inputs that can induce different behaviors in the tested DNNs, \ie different DNNs will classify the same input into different classes.
Suppose we have $n$ DNNs $F_{k\in 1..n}: \bm{x} \rightarrow \bm{y}$, where $F_k$ is the function modeled by the $k$-th neural network. $\bm{x}$ represents the input and $\bm{y}$ represents the output class probability vectors. Given an arbitrary $\bm{x}$ as seed that gets classified to the same class by all DNNs, our goal is to modify $\bm{x}$ such that the modified input $\bm{x'}$ will be classified differently by at least one of the $n$ DNNs.

\begin{algorithm}
\renewcommand{\arraystretch}{.8}
\setlength{\tabcolsep}{2pt}
\caption{Test input generation via joint optimization}
\footnotesize
\label{alg:testgen}
\begin{tabular}{|lp{2.9in}|}\hline
\textbf{Input}:
    & \textbf{seed\_set} $\leftarrow$ unlabeled inputs as the seeds \\
    & \textbf{dnns} $\leftarrow$ multiple DNNs under test \\
	& \boldsymbol{$\lambda_{1}$} $\leftarrow$ parameter to balance output differences of DNNs (Equation~\ref{eq:objective 1}) \\
    & \boldsymbol{$\lambda_{2}$} $\leftarrow$ parameter to balance coverage and differential behavior \\
    & \textbf{s} $\leftarrow$ step size in gradient ascent\\
    & \textbf{t} $\leftarrow$ threshold for determining if a neuron is activated\\
    & \textbf{p} $\leftarrow$ desired neuron coverage\\
    & \textbf{cov\_tracker} $\leftarrow$ tracks which neurons have been activated\\
\hline
\end{tabular}

\begin{spacing}{0.7}
\begin{algorithmic}[1]
    \State \textit{/* main procedure */}
    \State gen\_test := empty set
    \For{cycle(x $\in$ seed\_set)}\ \ // \textit{infinitely cycling through seed\_set}
        \State \textit{/* all dnns should classify the seed input to the same class */}
        \State c = dnns[0].predict(x)
    	\State d = randomly select one dnn from dnns
    	\While{True}
	    \State obj1 = COMPUTE\_OBJ1(x, d, c, dnns, $\lambda_{1}$)
	    \State obj2 = COMPUTE\_OBJ2(x, dnns, cov\_tracker)            
	    \State obj = obj1 + $\lambda_{2}$ $\cdot$ obj2
            \State grad = $\partial$obj / $\partial$x
            \State \textit{/*apply domain specific constraints to gradient*/}
            \State grad = DOMAIN\_CONSTRNTS(grad)
            \State x = x + s $\cdot$ grad\ \ //\textit{gradient ascent}
            \If{d.predict(x) $\neq$ (dnns-d).predict(x)}
                \State \textit{/* dnns predict x differently */}
            	\State gen\_test.add(x)
                \State update cov\_tracker 
                \State break
            \EndIf
        \EndWhile
	\If{DESIRED\_COVERAGE\_ACHVD(cov\_tracker)}
        	\State return gen\_test
        \EndIf
    \EndFor
    \State \textit{/* utility functions for computing obj1 and obj2 */}
	\Procedure{compute\_obj1}{x, d, c, dnns, $\lambda_{1}$}
    	\State rest = dnns - d
        \State loss1 := 0
        \For{dnn in rest}
        	\State loss1 += dnn$_c$(x)\ \ //\textit{confidence score of x being in class c}
        \EndFor
        \State loss2 := d$_c$(x)\ \  //\textit{d's output confidence score of x being in class c}
	\State return (loss1 - $\lambda_{1} \cdot$loss2)
    \EndProcedure
    
    \Procedure{compute\_obj2}{x, dnns, cov\_tracker}
    	\State loss := 0
    	\For{dnn $\in$ dnns}
        	\State select a neuron $n$ inactivated so far using cov\_tracker
            \State loss += n(x) \ //\textit{the neuron n's output when x is the dnn's input}
        \EndFor
        \State return loss
    \EndProcedure
     
\end{algorithmic}
\end{spacing}

\end{algorithm}

Let $F_k(\bm{x})[c]$ be the class probability that $F_k$ predicts $\bm{x}$ to be $c$.
We randomly select one neural network $F_j$ (Algorithm~\ref{alg:testgen} line 6) and maximize the following objective function:
\begin{equation}\footnotesize
\label{eq:objective 1}
\begin{split}
    obj_1(\bm{x}) = \Sigma_{k\neq j} F_k(\bm{x})[c] - \lambda_{1}\cdot F_j(\bm{x})[c]
\end{split}
\end{equation}
where $\lambda_{1}$ is a parameter to balance the objective terms between the DNNs $F_{k\neq j}$ that maintain the same class outputs as before and the DNN $F_j$ that produce different class outputs.
As all of $F_{k\in 1..n}$ are differentiable, Equation~\ref{eq:objective 1} can be easily maximized using gradient ascent by iteratively changing $\bm{x}$ based on the computed gradient: $\frac{\partial obj_1(\bm{x})}{\partial \bm{x}}$ (Algorithm~\ref{alg:testgen} line 8-14 and procedure \texttt{COMPUTE\_OBJ1}). 

\vspace{2pt}
\noindent
{\bf Maximizing neuron coverage.}
The second objective is to generate inputs that maximize neuron coverage. We achieve this goal by iteratively picking inactivated neurons and modifying the input such that output of that neuron goes above the neuron activation threshold. Let us assume that we want to maximize the output of a neuron $n$, \ie we want to maximize $obj_2(\bm{x})=f_n(\bm{x})$ such that $f_n(\bm{x}) > t$, where $t$ is the neuron activation threshold, and we write $f_n(\bm{x})$ as the function modeled by neuron $n$ that takes $\bm{x}$ (the original input to the DNN) as input and produce the output of neuron $n$ (as defined in Equation~\ref{eq:jcobian}). 
 We can again leverage the gradient ascent mechanism as $f_n(\bm{x})$ is a differentiable function whose gradient is $\frac{\partial f_n(\bm{x})}{\partial \bm{x}}$. 

Note that we can also potentially jointly maximize multiple neurons simultaneously, but we choose to activate one neuron at a time in this algorithm for clarity (Algorithm~\ref{alg:testgen} line 8-14 and procedure \texttt{COMPUTE\_OBJ2}).

\noindent
{\bf Joint optimization.}
We jointly maximize $obj_1$ and $f_n$ described above and maximize the following function:
\begin{equation}\footnotesize
\label{eq:objectives}
\begin{split}
    obj_{joint} = (\Sigma_{i\neq j} F_i(\bm{x})[c] - \lambda_{1} F_j(\bm{x})[c]) + \lambda_{2}\cdot f_n(\bm{x})
 \end{split}
\end{equation}
where $\lambda_{2}$ is a parameter for balancing between the two objectives of the joint optimization process and $n$ is the inactivated neuron that we randomly pick at each iteration (Algorithm~\ref{alg:testgen} line 33). As all terms of $obj_{joint}$ are differentiable, we jointly maximize them using gradient ascent by modifying $\bm{x}$ (Algorithm~\ref{alg:testgen} line 14).

\noindent
{\bf Domain-specific constraints.}
One important aspect of the optimization process is that the generated test inputs need to satisfy several domain-specific constraints to be physically realistic~\cite{adversarial:ccs16}. In particular, we want to ensure that the changes applied to $\bm{x_i}$ during the $i$-th iteration of gradient ascent process satisfy all the domain-specific constraints for all $i$.
For example, for a generated test image $\bm{x}$ the pixel values must be within a certain range (e.g., $0$ to $255$). 

While some such constraints can be efficiently embedded into the joint optimization process using the Lagrange Multipliers similar to those used in support vector machines~\cite{vapnik1998statistical}, we found that the majority of them cannot be easily handled by the optimization algorithm. Therefore, we designed a simple rule-based method to ensure that the generated tests satisfy the custom domain-specific constraints. As the seed input $\bm{x_{seed}}=\bm{x_0}$ always satisfy the constraints by definition, our technique must ensure that after $i$-th ($i>0$) iteration of gradient ascent, $\bm{x_i}$ still satisfies the constraints. Our algorithm ensures this property by modifying the gradient $\bm{grad}$ (line 13 in Algorithm~\ref{alg:testgen}) such that $\bm{x_{i+1}}=\bm{x_i}+s\cdot \bm{grad}$ still satisfies the constraints ($s$ is the step size in the gradient ascent).

For discrete features, we round the gradient to an integer. For DNNs handling visual input (\eg images), we add different spatial restrictions such that only part of the input images is modified. A detailed description of the domain-specific constraints that we implemented can be found in \S~\ref{subsubsec:physical-realizablity}.

\noindent\textbf{Hyperparameters in Algorithm~\ref{alg:testgen}.}
To summarize, there are four major hyperparameters that control different aspects of \prname as described below. 
(1) $\lambda_1$ balances the objectives between minimizing one DNN's prediction for a certain label and maximizing the rest of DNNs' predictions for the same label. Larger $\lambda_1$ puts higher priority on lowering the prediction value/confidence of a particular DNN while smaller $\lambda_1$ puts more weight on maintaining the other DNNs' predictions. 
(2) $\lambda_2$ provides balance between finding differential behaviors and neuron coverage. Larger $\lambda_2$ focuses more on covering different neurons while smaller $\lambda_2$ generates more difference-inducing test inputs. 
(3) $s$ controls the step size used during iterative gradient ascent. Larger $s$ may lead to oscillation around the local optimum while smaller $s$ may need more iterations to reach the objective. 
(4) $t$ is the threshold to determine whether each individual neuron is activated or not. Finding inputs that activate a neuron become increasingly harder as $t$ increases.

\section{Implementation}
\label{sec:implementation}
We implement \prname\ using TensorFlow 1.0.1~\cite{abadi2016tensorflow} and Keras 2.0.3~\cite{chollet2015keras} DL frameworks. Our implementation consists of around $7,086$ lines of Python code. Our code is built on TensorFlow/Keras but does not require any modifications to these frameworks. We leverage TensorFlow's efficient implementation of gradient computations in our joint optimization process. TensorFlow also supports creating sub-DNNs by marking any arbitrary neuron's output as the sub-DNN's output while keeping the input same as the original DNN's input. We use this feature to intercept and record the output of neurons in the intermediate layers of a DNN and compute the corresponding gradients with respect to the DNN's input. All our experiments were run on a Linux laptop running Ubuntu 16.04 (one Intel i7-6700HQ 2.60GHz processor with 4 cores, 16GB of memory, and a NVIDIA GTX 1070 GPU).

%
%

\section{Experimental Setup}
\label{sec:evaluation}



\subsection{Test datasets and DNNs}
\label{subsec:dataset-neuralnet}
We adopt five popular public datasets with different types of data\textemdash MNIST, ImageNet, Driving, Contagio/VirusTotal, and Drebin---and then evaluate \prname on three DNNs for each dataset (\ie a total of fifteen DNNs). 
We provide a summary of the five datasets and the corresponding DNNs in Table~\ref{tab:architecture}.
All the evaluated DNNs are either pre-trained (\ie we use public weights reported by previous researchers) or trained by us using public real-world architectures to achieve comparable performance to that of the state-of-the-art models for the corresponding dataset. 
 %
For each dataset, we used \prname to test three DNNs with different architectures as described in  Table~\ref{tab:architecture}. 

\begin{table*}
\setlength{\tabcolsep}{4pt}
\footnotesize
\renewcommand{\arraystretch}{1.1}
\centering
\caption{Details of the DNNs and datasets used to evaluate \prname}
\label{tab:architecture}
\begin{tabular}{llllllll}
\hline
\multicolumn{1}{|l|}{\textbf{Dataset}} & \multicolumn{1}{l|}{\begin{tabular}[l]{@{}l@{}}\textbf{Dataset}\\ \textbf{Description}\end{tabular}} & \multicolumn{1}{l|}{\begin{tabular}[l]{@{}l@{}}\textbf{DNN}\\ \textbf{Description}\end{tabular}} & \multicolumn{1}{l|}{\textbf{DNN Name}} & \multicolumn{1}{l|}{\begin{tabular}[l]{@{}l@{}}\textbf{\# of}\\ \textbf{Neurons}\end{tabular}} & \multicolumn{1}{l|}{\textbf{Architecture}} & \multicolumn{1}{l|}{\begin{tabular}[l]{@{}l@{}}\textbf{Reported}\\ \textbf{Acc.}\end{tabular}} & \multicolumn{1}{l|}{\begin{tabular}[c]{@{}l@{}}\textbf{Our}\\ \textbf{Acc.}\end{tabular}} \\ \hline
\multicolumn{1}{|l|}{\multirow{3}{*}{MNIST}} & \multicolumn{1}{l|}{\multirow{3}{*}{Hand-written digits}} & \multicolumn{1}{l|}{\multirow{3}{*}{LeNet variations}} & \multicolumn{1}{l|}{MNI\_C1} & \multicolumn{1}{l|}{52} & \multicolumn{1}{l|}{LeNet-1, LeCun et al.~\cite{lecun1998gradient,lecun2010mnist}} & \multicolumn{1}{l|}{98.3\%} & \multicolumn{1}{l|}{98.33\%} \\ \cline{4-8} 
\multicolumn{1}{|l|}{} & \multicolumn{1}{l|}{} & \multicolumn{1}{l|}{} & \multicolumn{1}{l|}{MNI\_C2} & \multicolumn{1}{l|}{148} & \multicolumn{1}{l|}{LeNet-4, LeCun et al.~\cite{lecun1998gradient,lecun2010mnist}} & \multicolumn{1}{l|}{98.9\%} & \multicolumn{1}{l|}{98.59\%} \\ \cline{4-8} 
\multicolumn{1}{|l|}{} & \multicolumn{1}{l|}{} & \multicolumn{1}{l|}{} & \multicolumn{1}{l|}{MNI\_C3} & \multicolumn{1}{l|}{268} & \multicolumn{1}{l|}{LeNet-5, LeCun et al.~\cite{lecun1998gradient,lecun2010mnist}} & \multicolumn{1}{l|}{99.05\%} & \multicolumn{1}{l|}{98.96\%} \\ \hline
\multicolumn{1}{|l|}{\multirow{3}{*}{Imagenet}} & \multicolumn{1}{l|}{\multirow{3}{*}{General images}} & \multicolumn{1}{l|}{\multirow{3}{*}{\begin{tabular}[c]{@{}l@{}}State-of-the-art\\ image classifiers\\ from ILSVRC\end{tabular}}} & \multicolumn{1}{l|}{IMG\_C1} & \multicolumn{1}{l|}{14,888} & \multicolumn{1}{l|}{VGG-16, Simonyan et al.~\cite{simonyan2014very}} & \multicolumn{1}{l|}{92.6\%$^{**}$} & \multicolumn{1}{l|}{92.6\%$^{**}$} \\ \cline{4-8} 
\multicolumn{1}{|l|}{} & \multicolumn{1}{l|}{} & \multicolumn{1}{l|}{} & \multicolumn{1}{l|}{IMG\_C2} & \multicolumn{1}{l|}{16,168} & \multicolumn{1}{l|}{VGG-19, Simonyan et al.~\cite{simonyan2014very}} & \multicolumn{1}{l|}{92.7\%$^{**}$} & \multicolumn{1}{l|}{92.7\%$^{**}$} \\ \cline{4-8} 
\multicolumn{1}{|l|}{} & \multicolumn{1}{l|}{} & \multicolumn{1}{l|}{} & \multicolumn{1}{l|}{IMG\_C3} & \multicolumn{1}{l|}{94,059} & \multicolumn{1}{l|}{ResNet50, He et al.~\cite{he2015deep}} & \multicolumn{1}{l|}{96.43\%$^{**}$} & \multicolumn{1}{l|}{96.43\%$^{**}$} \\ \hline
\multicolumn{1}{|l|}{\multirow{3}{*}{Driving}} & \multicolumn{1}{l|}{\multirow{3}{*}{Driving video frames}} & \multicolumn{1}{l|}{\multirow{3}{*}{\begin{tabular}[c]{@{}l@{}}Nvidia DAVE \\ self-driving systems\end{tabular}}} & \multicolumn{1}{l|}{DRV\_C1} & \multicolumn{1}{l|}{1,560} & \multicolumn{1}{l|}{Dave-orig~\cite{autopilot:dave}, Bojarski et al.~\cite{bojarski2016end}} & \multicolumn{1}{l|}{N/A} & \multicolumn{1}{l|}{99.91\%$^\#$} \\ \cline{4-8} 
\multicolumn{1}{|l|}{} & \multicolumn{1}{l|}{} & \multicolumn{1}{l|}{} & \multicolumn{1}{l|}{DRV\_C2} & \multicolumn{1}{l|}{1,560}  & \multicolumn{1}{l|}{Dave-norminit~\cite{visualize:dave}} & \multicolumn{1}{l|}{N/A} & \multicolumn{1}{l|}{99.94\%$^\#$} \\ \cline{4-8} 
\multicolumn{1}{|l|}{} & \multicolumn{1}{l|}{} & \multicolumn{1}{l|}{} & \multicolumn{1}{l|}{DRV\_C3} & \multicolumn{1}{l|}{844} & \multicolumn{1}{l|}{Dave-dropout~\cite{clone:dave}} & \multicolumn{1}{l|}{N/A} & \multicolumn{1}{l|}{99.96\%$^\#$} \\ \hline
\multicolumn{1}{|l|}{\multirow{3}{*}{Contagio/Virustotal}} & \multicolumn{1}{l|}{\multirow{3}{*}{PDFs}} & \multicolumn{1}{l|}{\multirow{3}{*}{\begin{tabular}[c]{@{}l@{}}PDF\\ malware detectors\end{tabular}}} & \multicolumn{1}{l|}{PDF\_C1} & \multicolumn{1}{l|}{402} & \multicolumn{1}{l|}{<200, 200>$^+$} & \multicolumn{1}{l|}{98.5\%$^-$} & \multicolumn{1}{l|}{96.15\%} \\ \cline{4-8} 
\multicolumn{1}{|l|}{} & \multicolumn{1}{l|}{} & \multicolumn{1}{l|}{} & \multicolumn{1}{l|}{PDF\_C2} & \multicolumn{1}{l|}{602} & \multicolumn{1}{l|}{<200, 200, 200>$^+$} & \multicolumn{1}{l|}{98.5\%$^-$} & \multicolumn{1}{l|}{96.25\%} \\ \cline{4-8} 
\multicolumn{1}{|l|}{} & \multicolumn{1}{l|}{} & \multicolumn{1}{l|}{} & \multicolumn{1}{l|}{PDF\_C3} & \multicolumn{1}{l|}{802} & \multicolumn{1}{l|}{<200, 200, 200, 200>$^+$} & \multicolumn{1}{l|}{98.5\%$^-$} & \multicolumn{1}{l|}{96.47\%} \\ \hline
\multicolumn{1}{|l|}{\multirow{3}{*}{Drebin}} & \multicolumn{1}{l|}{\multirow{3}{*}{Android apps}} & \multicolumn{1}{l|}{\multirow{3}{*}{\begin{tabular}[c]{@{}l@{}}Android app\\ malware detectors\end{tabular}}} & \multicolumn{1}{l|}{APP\_C1} & \multicolumn{1}{l|}{402} & \multicolumn{1}{l|}{<200, 200>$^+$, Grosse et al.~\cite{grosse2016adversarial}} & \multicolumn{1}{l|}{98.92\%} & \multicolumn{1}{l|}{98.6\%} \\ \cline{4-8} 
\multicolumn{1}{|l|}{} & \multicolumn{1}{l|}{} & \multicolumn{1}{l|}{} & \multicolumn{1}{l|}{APP\_C2} & \multicolumn{1}{l|}{102} & \multicolumn{1}{l|}{<50, 50>$^+$, Grosse et al.~\cite{grosse2016adversarial}} & \multicolumn{1}{l|}{96.79\%} & \multicolumn{1}{l|}{96.82\%} \\ \cline{4-8} 
\multicolumn{1}{|l|}{} & \multicolumn{1}{l|}{} & \multicolumn{1}{l|}{} & \multicolumn{1}{l|}{APP\_C3} & \multicolumn{1}{l|}{212} & \multicolumn{1}{l|}{<200, 10>$^+$, Grosse et al.~\cite{grosse2016adversarial}} & \multicolumn{1}{l|}{92.97\%} & \multicolumn{1}{l|}{92.66\%} \\ \hline
\multicolumn{7}{l}{\scriptsize\begin{tabular}[c]{@{}l@{}}** top-5 test accuracy; we exactly match the reported performance as we use the pretrained networks\\ \# we report $1$-MSE (Mean Squared Error) as the accuracy because steering angle is a continuous value\\ + <x,y,...> denotes three hidden layers with $x$ neurons in first layer, $y$ neurons in second layer and so on\\ - accuracy using SVM as reported by \v{S}rndic et al.~\cite{6956565} 
\end{tabular}}
\end{tabular}
\vspace{-0.2 cm}
\end{table*}

\noindent \textbf{MNIST}~\cite{lecun1998mnist} is a large handwritten digit dataset containing 28x28 pixel images with class labels from $0$ to $9$ 
The dataset includes $60,000$ training samples and $10,000$ testing samples. We follow Lecun et al.~\cite{lecun1998gradient} and construct three different neural networks based on the LeNet family~\cite{lecun1998gradient}, \ie the LeNet-1, LeNet-4, and LeNet-5. 

\noindent \textbf{ImageNet}~\cite{deng2009imagenet} is a large image dataset with over $10,000,000$ hand-annotated images that are crowdsourced and labeled manually. We test three well-known pre-trained DNNs: VGG-16~\cite{simonyan2014very}, VGG-19~\cite{simonyan2014very}, and ResNet50~\cite{he2015deep}. All three DNNs achieved competitive performance in the ILSVRC~\cite{ILSVRC15} competition. 

\noindent \textbf{Driving}~\cite{udacity:challenge} is the Udacity self-driving car challenge dataset that contains images captured by a camera mounted behind the windshield of a driving car and the simultaneous steering wheel angle applied by the human driver for each image.  The dataset has 101,396 training and 5,614 testing samples. We then used three DNNs~\cite{autopilot:dave, visualize:dave, clone:dave} based on the DAVE-2 self-driving car architecture from Nvidia~\cite{bojarski2016end} with slightly different configurations, which are called DAVE-orig, DAVE-norminit, and DAVE-dropout respectively. 
%
Specifically, DAVE-orig~\cite{autopilot:dave} fully replicates the original architecture from the Nvidia's paper~\cite{bojarski2016end}. DAVE-norminit~\cite{visualize:dave} removes the first batch normalization layer~\cite{ioffe2015batch} and normalizes the randomly initialized network weights. DAVE-dropout~\cite{clone:dave} simplifies DAVE-orig by cutting down the numbers of convolutional layers and fully connected layers. DAVE-dropout also adds two dropout layer~\cite{srivastava2014dropout} between the final three fully-connected layers. We trained all three implementations with the Udacity self-driving car challenge dataset mentioned above.

\noindent \textbf{Contagio/VirusTotal}~\cite{contagio,virustotal} is a dataset containing different benign and malicious PDF documents. We use $5,000$ benign and $12,205$ malicious PDF documents from Contagio database as the training set, and then use $5,000$ malicious PDFs collected by VirusTotal~\cite{virustotal} and $5,000$ benign PDFs crawled from Google as the test set. To the best of our knowledge, there is no publicly available DNN-based PDF malware detection system. Therefore, we define and train three different DNNs using 135 static features from PDFrate~\cite{pdfrate,smutz2012malicious}, an online service for PDF malware detection. Specifically, we construct neural networks with one input layer, one softmax output layer, and $N$ fully-connected hidden layers with 200 neurons where $N$ ranges from 2 to 4 for the three tested DNNs. All our models achieve similar performance to the ones reported by a prior work using SVM models on the same dataset~\cite{6956565}.

\noindent \textbf{Drebin}~\cite{arp2014drebin,spreitzenbarth2013mobile} is a dataset with $129,013$ Android applications among which $123,453$ are benign and $5,560$ are malicious. There is a total of $545,333$ binary features categorized into eight sets including the features captured from manifest files (\eg requested permissions and intents) and disassembled code (\eg restricted API calls and network addresses).  We adopt the architecture of 3 out of 36 DNNs constructed by Grosse et al.~\cite{grosse2016adversarial}. As the DNNs' weights are not available, we train these three DNNs with 66\% randomly picked Android applications from the dataset and use the rest as the test set. 


\vspace{-0.1cm}
\subsection{Domain-specific constraints}
\label{subsubsec:physical-realizablity}

As discussed earlier, to be useful in practice, we need to ensure that the generated tests are valid and realistic by applying domain-specific constraints. For example, generated images should be physically producible by a camera. Similarly, generated PDFs need to follow the PDF specification to ensure that a PDF viewer can open the test file. Below we describe two major types of domain-specific constraints (\ie image and file constraints) that we use in this paper. 


\begin{figure*}
\captionsetup[subfloat]{captionskip=.4pt, labelformat=empty, font=scriptsize, belowskip=-1pt}
{\bf \scriptsize \hspace{0.025in} Different lighting conditions:
\vspace{-0.1in}}

\subfloat[all:right]{
\includegraphics[width=0.1\textwidth, height=0.1\textwidth]{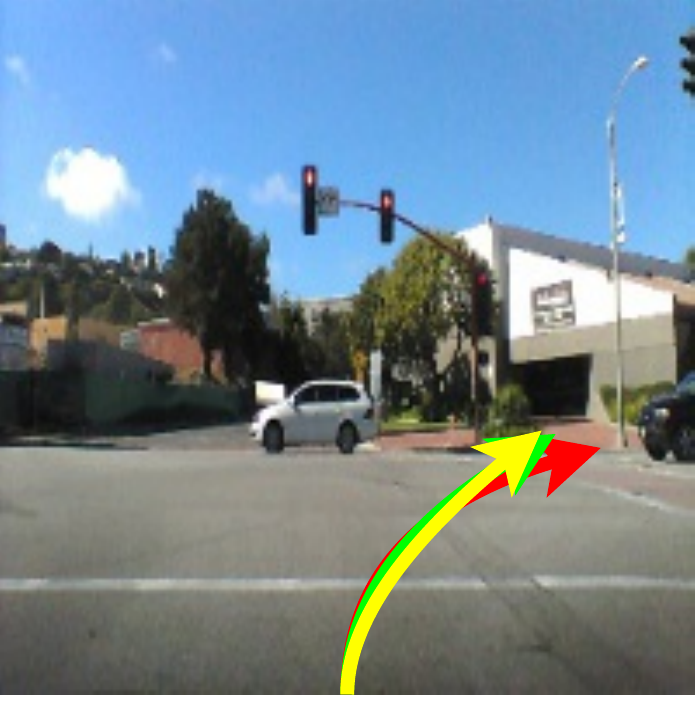}
\label{subfig:light_car_rect4_orig}}
\subfloat[all:right]{
\includegraphics[width=0.1\textwidth, height=0.1\textwidth]{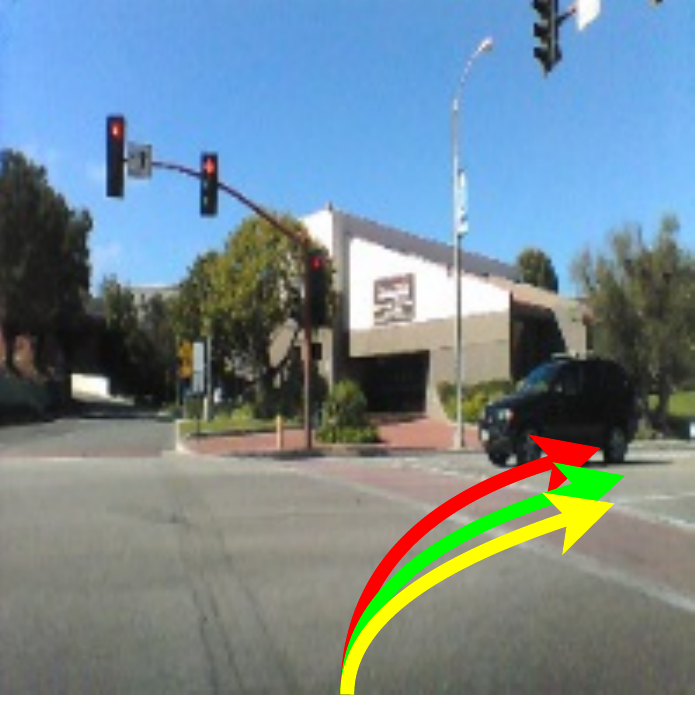}
\label{subfig:light_car_rect2_orig}}
\subfloat[all:right]{
\includegraphics[width=0.1\textwidth, height=0.1\textwidth]{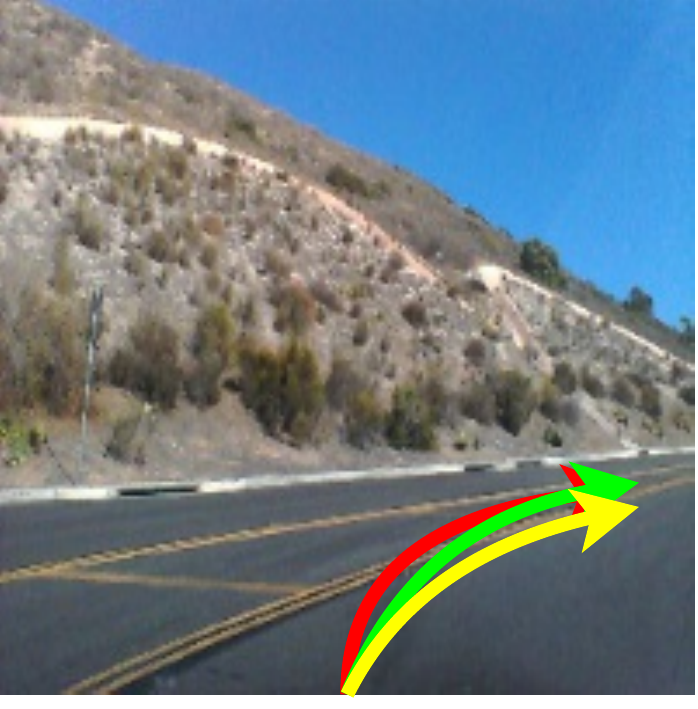}
\label{subfig:light_car_rect3_orig}}
\hfill
\subfloat[all:1]{
\includegraphics[width=0.1\textwidth, height=0.1\textwidth]{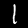}
\label{subfig:18_orig}}
\subfloat[all:3]{
\includegraphics[width=0.1\textwidth, height=0.1\textwidth]{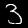}
\label{subfig:35_orig}}
\subfloat[all:5]{
\includegraphics[width=0.1\textwidth, height=0.1\textwidth]{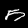}
\label{subfig:57_orig}}
\hfill
\subfloat[all:diver]{
\includegraphics[width=0.1\textwidth, height=0.1\textwidth]{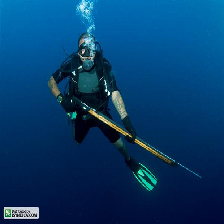}
\label{subfig:scuba_diver_to_ski_orig}}
\subfloat[all:cheeseburger]{
\includegraphics[width=0.1\textwidth, height=0.1\textwidth]{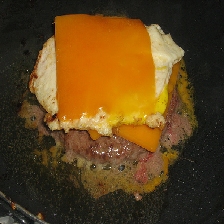}
\label{subfig:cheeseburger_to_ice_cream_orig}}
\subfloat[all:flamingo]{
\includegraphics[width=0.1\textwidth, height=0.1\textwidth]{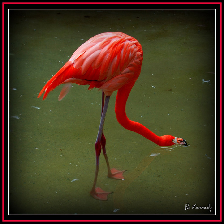}
\label{subfig:flamingo_to_goldfish_orig}}
\vspace{-0.41cm}
\subfloat[DRV\_C1:left]{
\includegraphics[width=0.1\textwidth, height=0.1\textwidth]{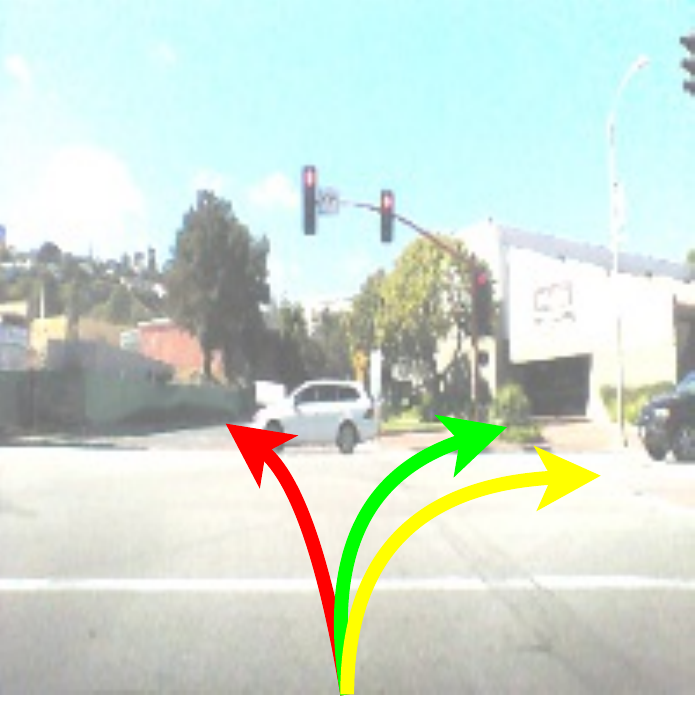}
\label{subfig:light_car_rect1}}
\subfloat[DRV\_C2:left]{
\includegraphics[width=0.1\textwidth, height=0.1\textwidth]{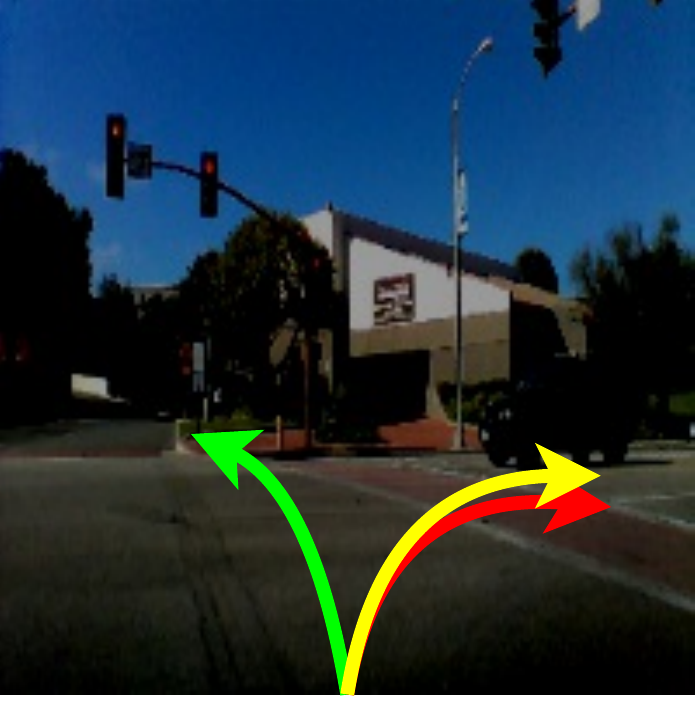}
\label{subfig:light_car_rect2}}
\subfloat[DRV\_C3:left]{
\includegraphics[width=0.1\textwidth, height=0.1\textwidth]{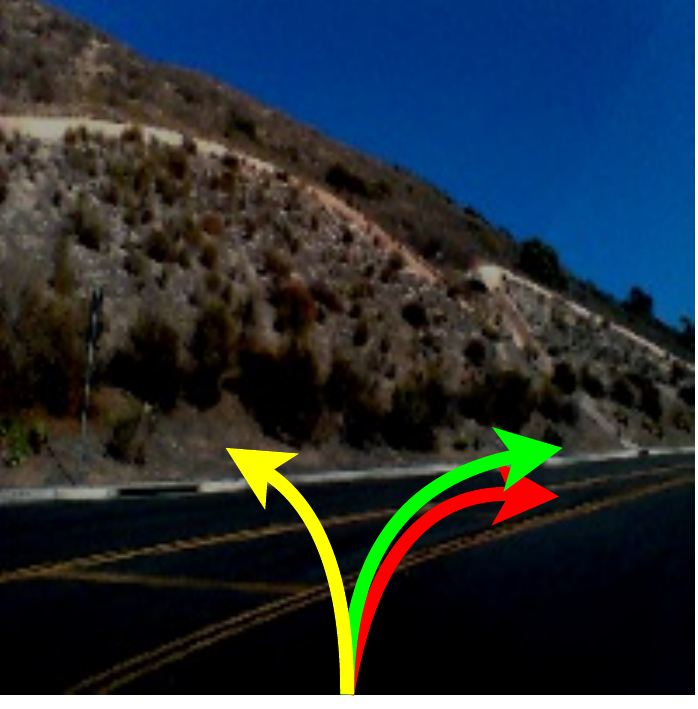}
\label{subfig:light_car_rect3}}
\hfill
\subfloat[MNI\_C1:8]{
\includegraphics[width=0.1\textwidth, height=0.1\textwidth]{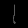}
\label{subfig:18}}
\subfloat[MNI\_C2:5]{
\includegraphics[width=0.1\textwidth, height=0.1\textwidth]{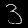}
\label{subfig:35}}
\subfloat[MNI\_C3:7]{
\includegraphics[width=0.1\textwidth, height=0.1\textwidth]{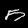}
\label{subfig:57}}
\hfill
\subfloat[IMG\_C1:ski]{
\includegraphics[width=0.1\textwidth, height=0.1\textwidth]{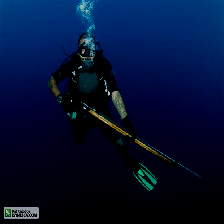}
\label{subfig:scuba_diver_to_ski}}
\subfloat[IMG\_C2:icecream]{
\includegraphics[width=0.1\textwidth, height=0.1\textwidth]{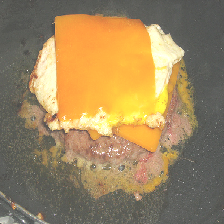}
\label{subfig:cheeseburger_to_ice_cream}}
\subfloat[IMG\_C3:goldfish]{
\includegraphics[width=0.1\textwidth, height=0.1\textwidth]{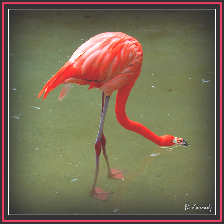}
\label{subfig:flamingo_to_goldfish}}

{\bf \scriptsize \hspace{0.025in} Occlusion with a single small rectangle:\vspace{-0.1in}}

\subfloat[all:right]{
\includegraphics[width=0.1\textwidth, height=0.1\textwidth]{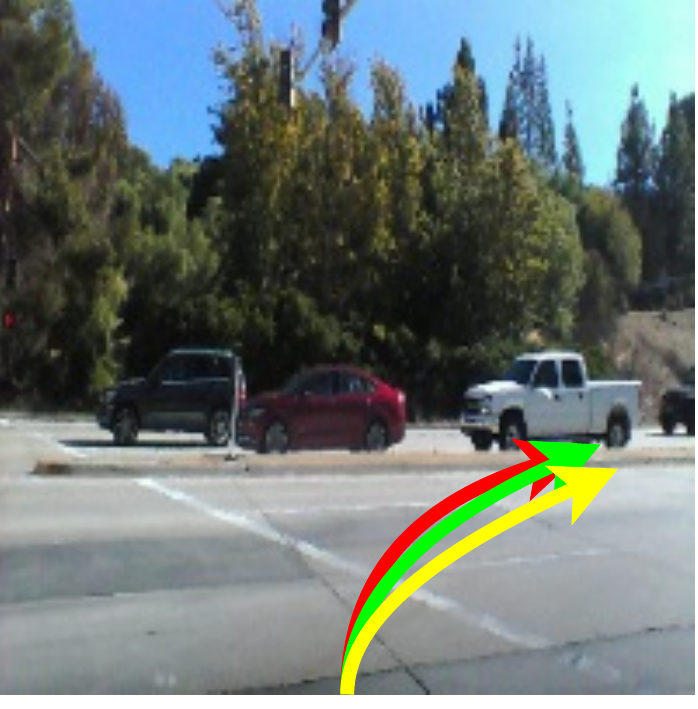}
\label{subfig:light_car_rect5_orig}}
\subfloat[all:right]{
\includegraphics[width=0.1\textwidth, height=0.1\textwidth]{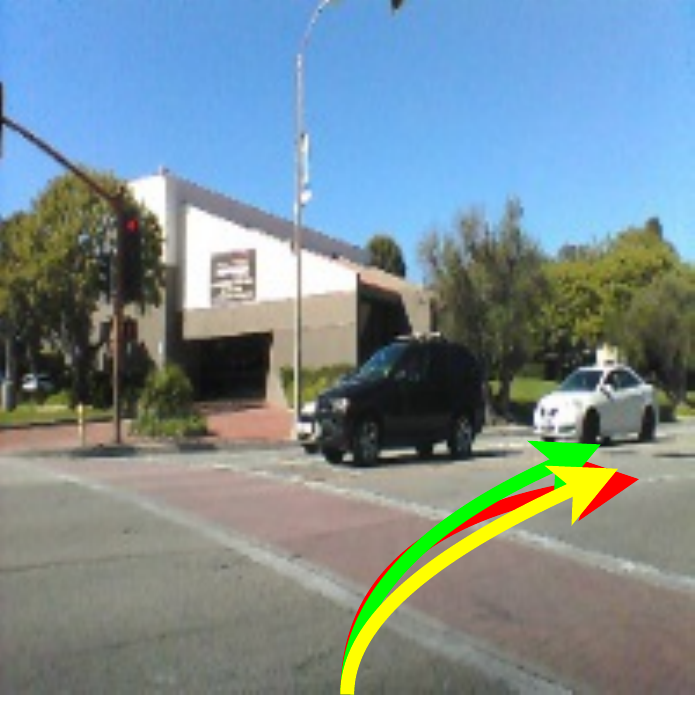}
\label{subfig:light_car_rect6_orig}}
\subfloat[all:left]{
\includegraphics[width=0.1\textwidth, height=0.1\textwidth]{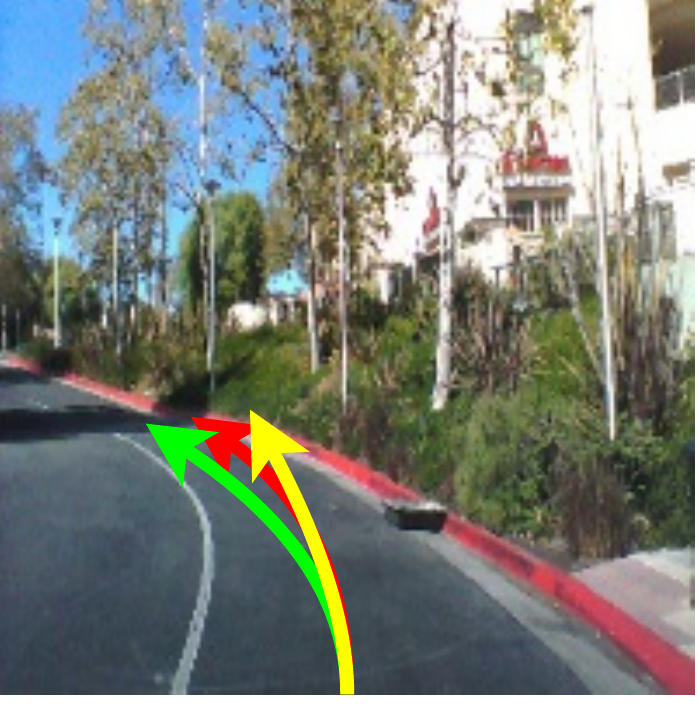}
\label{subfig:light_car_rect7_orig}}
\hfill
\subfloat[all:5]{
\includegraphics[width=0.1\textwidth, height=0.1\textwidth]{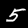}
\label{subfig:53_orig}}
\subfloat[all:7]{
\includegraphics[width=0.1\textwidth, height=0.1\textwidth]{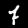}
\label{subfig:74_orig}}
\subfloat[all: 9]{
\includegraphics[width=0.1\textwidth, height=0.1\textwidth]{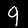}
\label{subfig:92_orig}}
\hfill
\subfloat[all:cauliflower]{
\includegraphics[width=0.1\textwidth, height=0.1\textwidth]{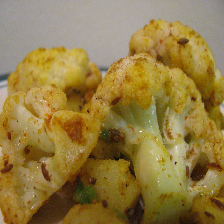}
\label{subfig:cauliflower_to_carbonara_orig}}
\subfloat[all:dhole]{
\includegraphics[width=0.1\textwidth, height=0.1\textwidth]{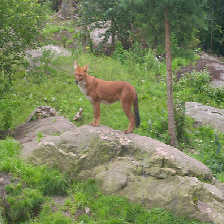}
\label{subfig:dhole_to_umbrella_orig}}
\subfloat[all:hay]{
\includegraphics[width=0.1\textwidth, height=0.1\textwidth]{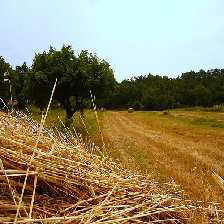}
\label{subfig:hay_to_stupa_orig}}
\vspace{-0.41cm}
\subfloat[DRV\_C1:left]{
\includegraphics[width=0.1\textwidth, height=0.1\textwidth]{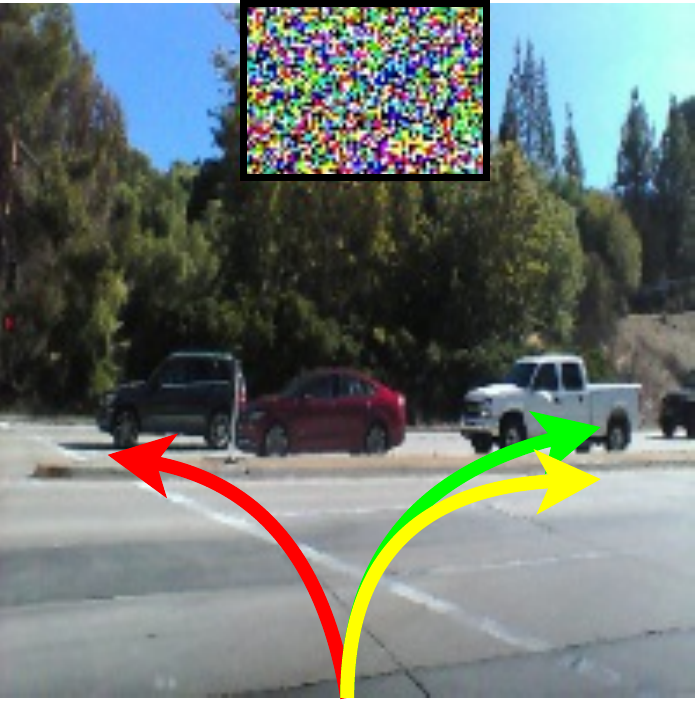}
\label{subfig:light_car_rect5}}
\subfloat[DRV\_C2:left]{
\includegraphics[width=0.1\textwidth, height=0.1\textwidth]{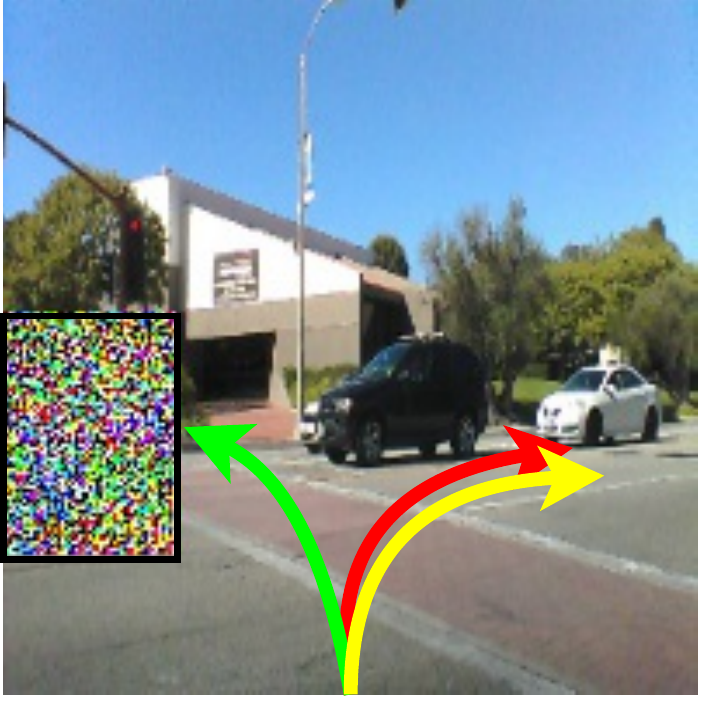}
\label{subfig:light_car_rect6}}
\subfloat[DRV\_C3:right]{
\includegraphics[width=0.1\textwidth, height=0.1\textwidth]{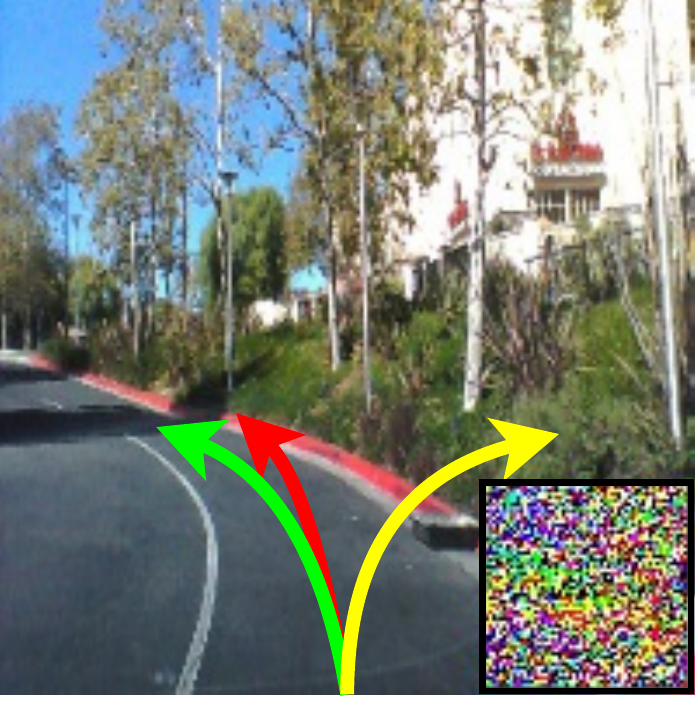}
\label{subfig:light_car_rect7}}
\hfill
\subfloat[MNI\_C1:3]{
\includegraphics[width=0.1\textwidth, height=0.1\textwidth]{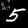}
\label{subfig:53}}
\subfloat[MNI\_C2:4]{
\includegraphics[width=0.1\textwidth, height=0.1\textwidth]{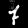}
\label{subfig:74}}
\subfloat[MNI\_C3:2]{
\includegraphics[width=0.1\textwidth, height=0.1\textwidth]{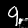}
\label{subfig:92}}
\hfill
\subfloat[IMG\_C1:carbonara]{
\includegraphics[width=0.1\textwidth, height=0.1\textwidth]{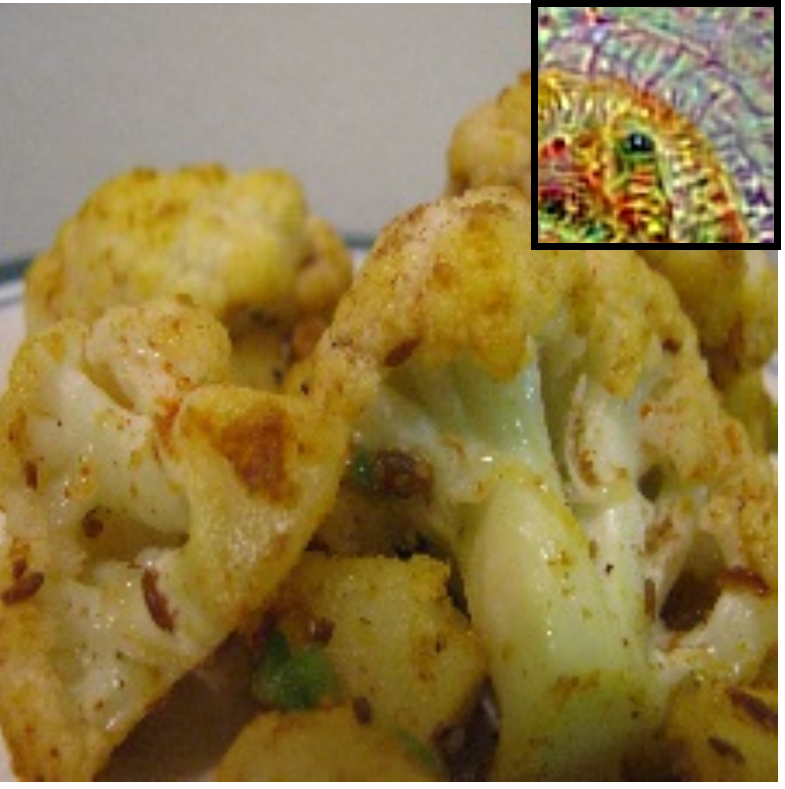}
\label{subfig:cauliflower_to_carbonara}}
\subfloat[IMG\_C2:umbrella]{
\includegraphics[width=0.1\textwidth, height=0.1\textwidth]{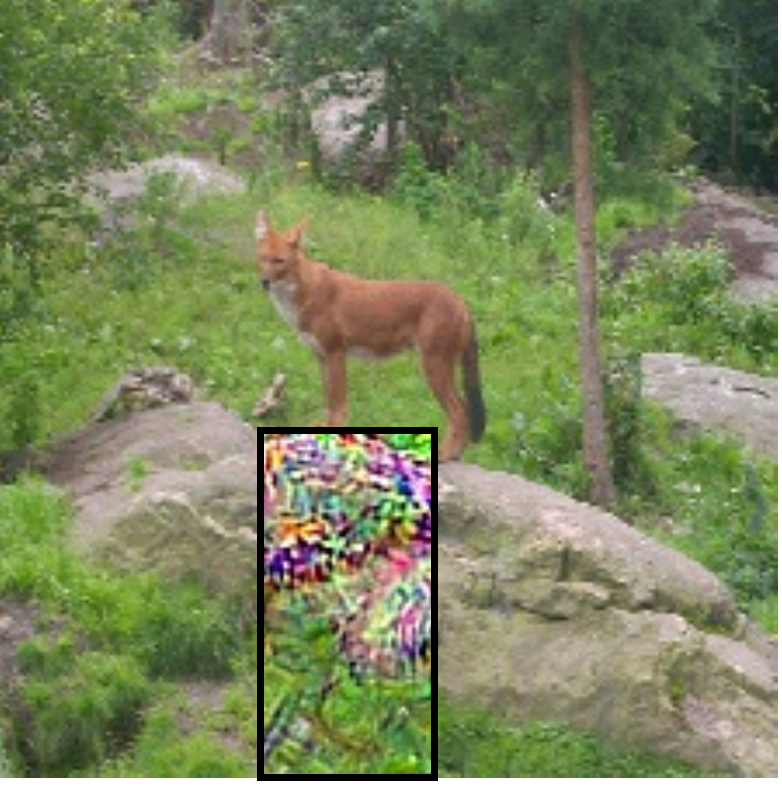}
\label{subfig:dhole_to_umbrella}}
\subfloat[IMG\_C3:stupa]{
\includegraphics[width=0.1\textwidth, height=0.1\textwidth]{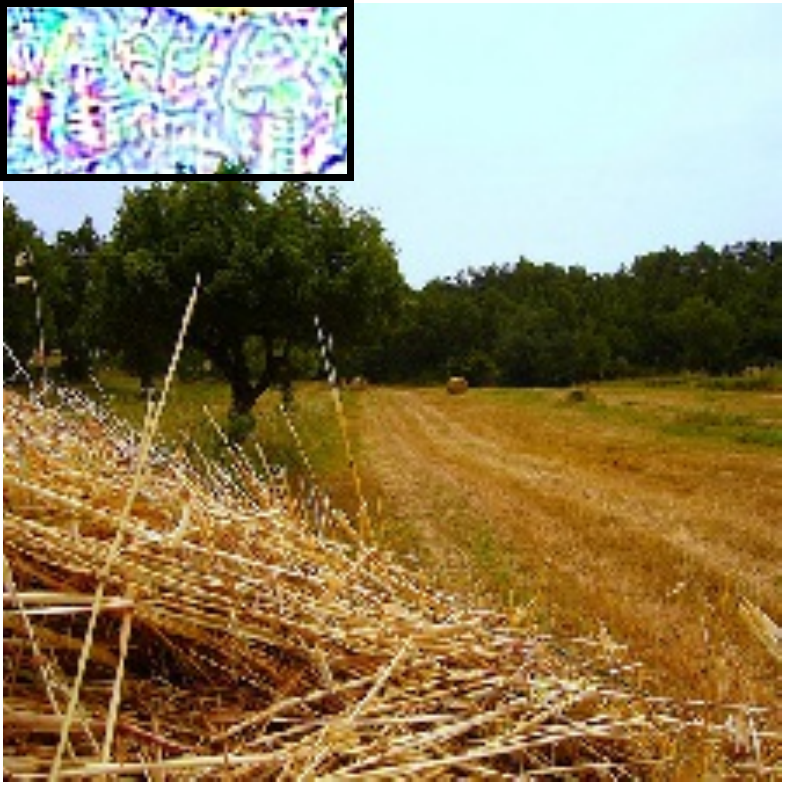}
\label{subfig:hay_to_stupa}}

{\bf \scriptsize \hspace{0.025in} Occlusion with multiple tiny black rectangles:\vspace{-0.1in}}

\subfloat[all:left]{
\includegraphics[width=0.1\textwidth, height=0.1\textwidth]{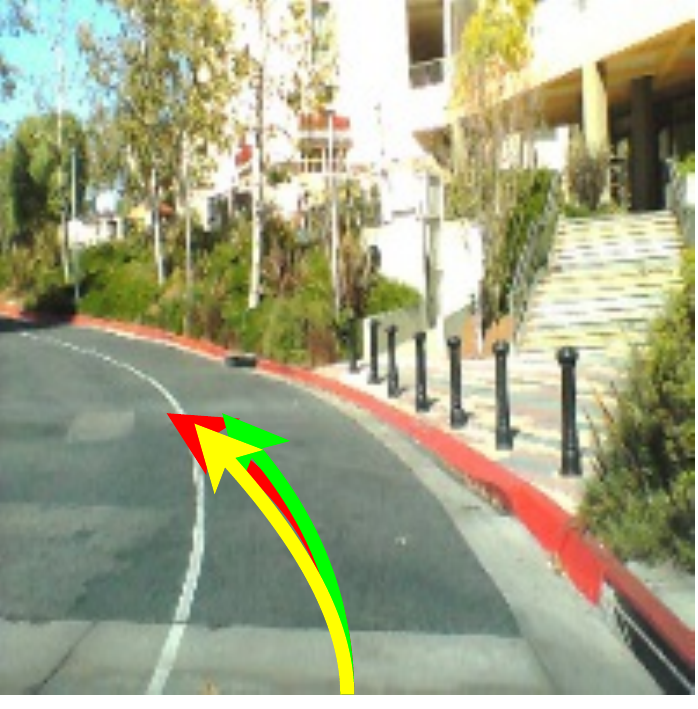}
\label{subfig:car_rect11_orig}}
\subfloat[all:left]{
\includegraphics[width=0.1\textwidth, height=0.1\textwidth]{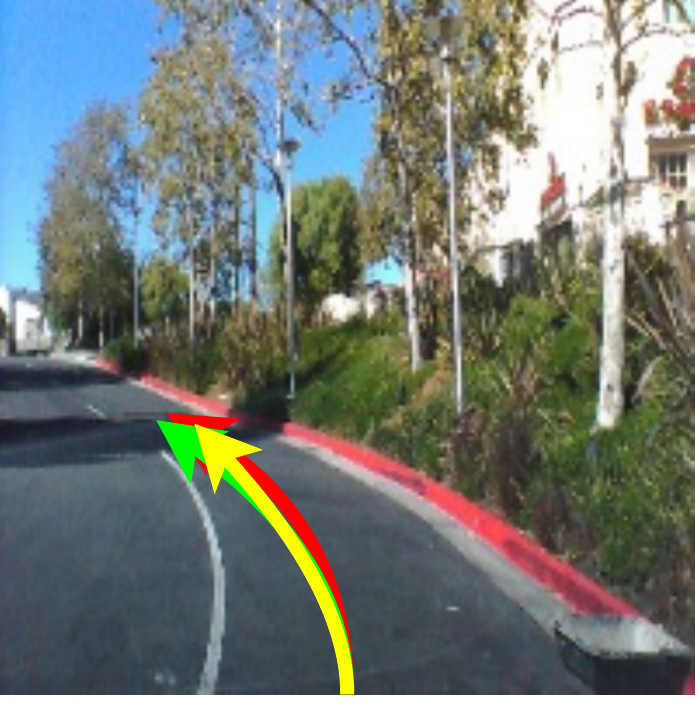}
\label{subfig:car_rect12_orig}}
\subfloat[all:left]{
\includegraphics[width=0.1\textwidth, height=0.1\textwidth]{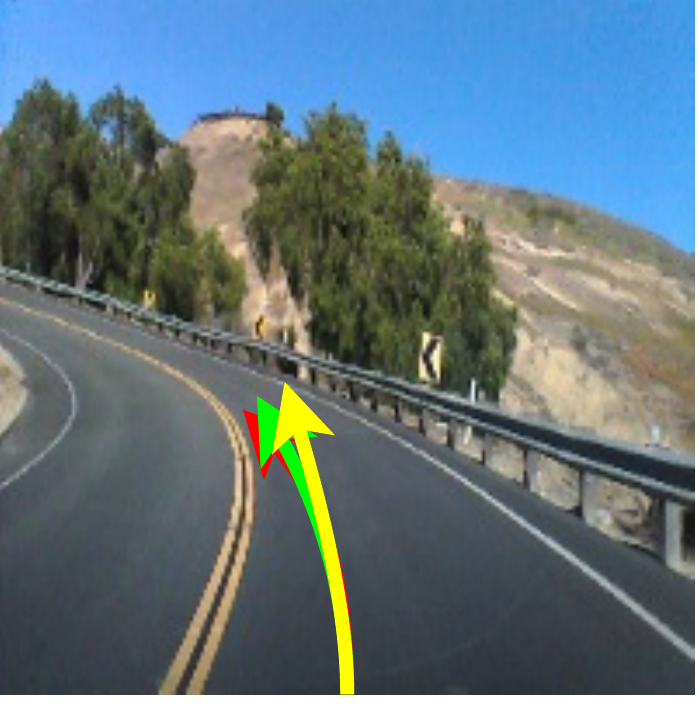}
\label{subfig:car_rect13_orig}}
\hfill
\subfloat[all:1]{
\includegraphics[width=0.1\textwidth, height=0.1\textwidth]{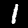}
\label{subfig:122_orig}}
\subfloat[all:5]{
\includegraphics[width=0.1\textwidth, height=0.1\textwidth]{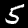}
\label{subfig:544_orig}}
\subfloat[all:7]{
\includegraphics[width=0.1\textwidth, height=0.1\textwidth]{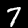}
\label{subfig:744_orig}}
\hfill
\subfloat[all:castle]{
\includegraphics[width=0.1\textwidth, height=0.1\textwidth]{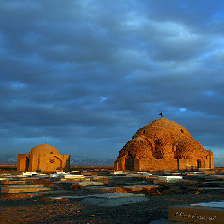}
\label{subfig:castle_to_beacon_orig}}
\subfloat[all:cock]{
\includegraphics[width=0.1\textwidth, height=0.1\textwidth]{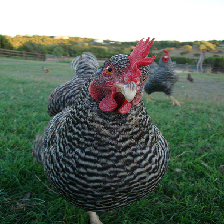}
\label{subfig:cock_to_hen_orig}}
\subfloat[all:groom]{
\includegraphics[width=0.1\textwidth, height=0.1\textwidth]{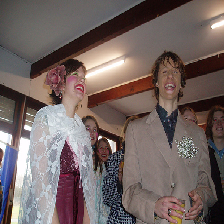}
\label{subfig:groom_to_vestment_orig}}
\vspace{-0.41cm}
\subfloat[DRV\_C1:right]{
\includegraphics[width=0.1\textwidth, height=0.1\textwidth]{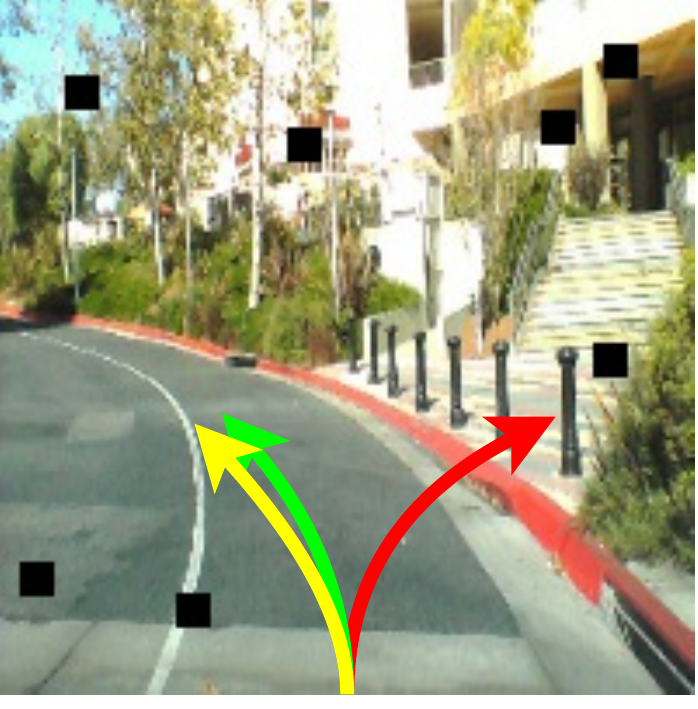}
\label{subfig:car_rect11}}
\subfloat[DRV\_C2:right]{
\includegraphics[width=0.1\textwidth, height=0.1\textwidth]{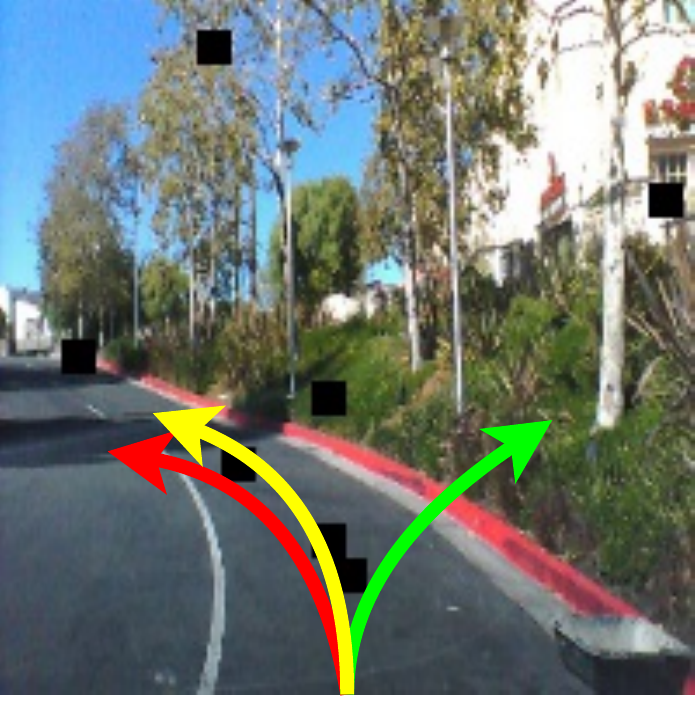}
\label{subfig:car_rect12}}
\subfloat[DRV\_C3:right]{
\includegraphics[width=0.1\textwidth, height=0.1\textwidth]{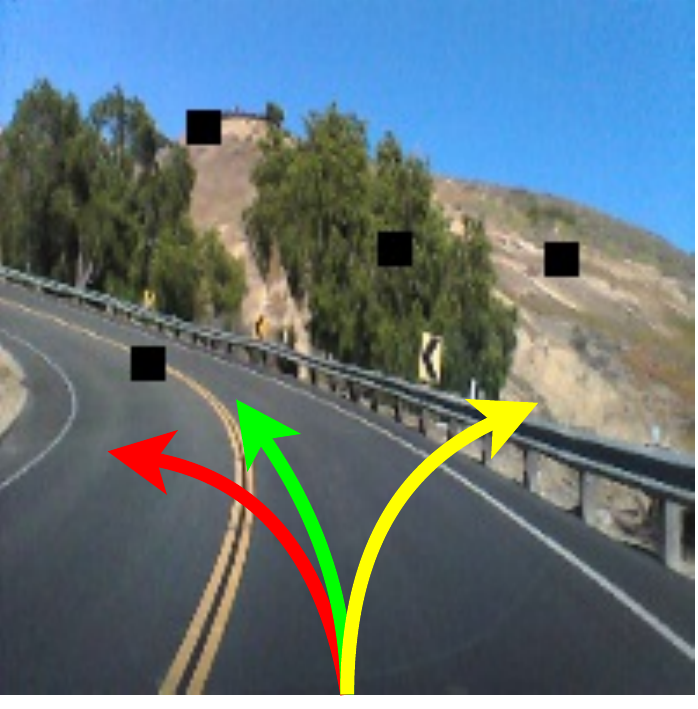}
\label{subfig:car_rect13}}
\hfill
\subfloat[MNI\_C1:2]{
\includegraphics[width=0.1\textwidth, height=0.1\textwidth]{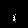}
\label{subfig:122}}
\subfloat[MNI\_C2:4]{
\includegraphics[width=0.1\textwidth, height=0.1\textwidth]{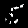}
\label{subfig:544}}
\subfloat[MNI\_C3:4]{
\includegraphics[width=0.1\textwidth, height=0.1\textwidth]{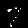}
\label{subfig:744}}
\hfill
\subfloat[IMG\_C1:beacon]{
\includegraphics[width=0.1\textwidth, height=0.1\textwidth]{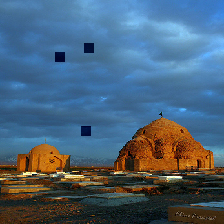}
\label{subfig:castle_to_beacon}}
\subfloat[IMG\_C2:hen]{
\includegraphics[width=0.1\textwidth, height=0.1\textwidth]{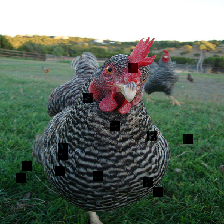}
\label{subfig:cock_to_hen}}
\subfloat[IMG\_C3:vestment]{
\includegraphics[width=0.1\textwidth, height=0.1\textwidth]{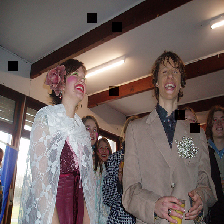}
\label{subfig:groom_to_vestment}}
\caption{Odd rows show the seed test inputs and even rows show the difference-inducing test inputs generated by \prname. The left three columns show inputs for self-driving car, the middle three are for MNIST, and the right three are for ImageNet.}
\vspace{-.1cm}
\label{fig:results}
\end{figure*}


\noindent \textbf{Image constraints (MNIST, ImageNet, and Driving).}
\prname used three different types of constraints for simulating different environment conditions of images: (1) lighting effects for simulating different intensities of lights, (2) occlusion by a single small rectangle for simulating an attacker potentially blocking some parts of a camera, and (3) occlusion by multiple tiny black rectangles for simulating effects of dirt on camera lens.  

The first constraint restricts image modifications so that \prname can only make the image darker or brighter without changing its content. Specifically, the modification can only increase or decrease all pixel values by the same amount (\eg $1*stepsize$ in line 14 of Algorithm~\ref{alg:testgen})---the decision to increase or decrease depends on the value of $mean(G)$ where $G$ denotes the gradients calculated at each iteration of gradient ascent. Note that $mean(G)$ simply denotes the mean of all entries in the multi-dimensional array $G$. The first and second rows of Figure~\ref{fig:results} show some examples of the difference-inducing inputs generated by \prname with these constraints. 

The second constraint simulates the effect of the camera lens that may be accidentally or deliberately occluded by a single small rectangle $R$ ($m \times n$ pixels).
Specifically, we apply only $G_{i:i+m,j:j+n}$ to the original image ($I$) where $I_{i:i+m,j:j+n}$ is the location of $R$. Note that \prname is free to choose any values of $i$ and $j$ to place the rectangle $R$ at any arbitrary position within the image.
The third and fourth rows of Figure~\ref{fig:results} show some examples \prname  generated while running with such occlusion constraints.



The third constraint restricts the modifications so that \prname only selects a tiny $m\times m$ size patch, $G_{i:i+m,j:j+m}$, from $G$ with upper-left corner at $(i,j)$ during each iteration of the gradient ascent. If the average value $mean(G_{i:i+m,j:j+m})$ of this patch is greater than $0$, we set $G_{i:i+m,j:j+m}=0$, \ie we only allow the pixel values to be decreased. Unlike the second constraint described above, here \prname will pick multiple positions (i.e., multiple $(i,j)$ pairs) to place the black rectangles simulating pieces of dirt on the camera lens. The fifth and sixth rows of Figure~\ref{fig:results} show some generated examples with these constraints.


\noindent \textbf{Other constraints (Drebin and Contagio/VirusTotal).} For Drebin dataset, \prname enforces a constraint that only allows modifying features related to the Android manifest file and thus ensures that the application code is unaffected. Moreover, \prname only allows adding features (changing from zero to one) but do not allow deleting features (changing from one to zero) from the manifest files to ensure that no application functionality is changed due to insufficient permissions. Thus, after computing the gradient, \prname only modifies the manifest features whose corresponding gradients are greater than zero.

For Contagio/VirusTotal dataset, \prname follows the restrictions on each feature as described by \v{S}rndic et al.~\cite{6956565}. 

\section{Results}


\begin{table}[!htb]
\setlength{\tabcolsep}{6pt}
\footnotesize
\renewcommand{\arraystretch}{.95}
\centering
\caption{Number of difference-inducing inputs found by \prname for each tested DNN obtained by randomly selecting 2,000 seeds from the corresponding test set for each run.}
\label{tab:violations}
\begin{tabular}{|c|cccc|c|}
\hline
\multirow{2}{*}{\textbf{DNN name}} & \multicolumn{4}{c|}{\textbf{Hyperparams (Algorithm~\ref{alg:testgen})}} & \multirow{2}{*}{\textbf{\begin{tabular}[c]{@{}c@{}}\# Differences\\ Found\end{tabular}}} \\
\cline{2-5}
 & $\lambda_1$ & $\lambda_2$ & $s$ & $t$ &  \\ \hline
\textbf{MNI\_C1} & \multirow{3}{*}{1} & \multirow{3}{*}{0.1} & \multirow{3}{*}{10} & \multirow{3}{*}{0} & 1,073 \\
\textbf{MNI\_C2} &  &  &  &  & 1,968 \\
\textbf{MNI\_C3} &  &  &  &  & 827 \\ \hline
\textbf{IMG\_C1} & \multirow{3}{*}{1} & \multirow{3}{*}{0.1} & \multirow{3}{*}{10} & \multirow{3}{*}{0} & 1,969 \\
\textbf{IMG\_C2} &  &  &  &  & 1,976 \\
\textbf{IMG\_C3} &  &  &  &  & 1,996 \\ \hline
\textbf{DRV\_C1} & \multirow{3}{*}{1} & \multirow{3}{*}{0.1} & \multirow{3}{*}{10} & \multirow{3}{*}{0} & 1,720 \\
\textbf{DRV\_C2} &  &  &  &  & 1,866 \\
\textbf{DRV\_C3} &  &  &  &  & 1,930 \\ \hline
\textbf{PDF\_C1} & \multirow{3}{*}{2} & \multirow{3}{*}{0.1} & \multirow{3}{*}{0.1} & \multirow{3}{*}{0} & 1,103 \\
\textbf{PDF\_C2} &  &  &  &  & 789 \\
\textbf{PDF\_C3} &  &  &  &  & 1,253 \\ \hline
\textbf{APP\_C1} & \multirow{3}{*}{1} & \multirow{3}{*}{0.5} & \multirow{3}{*}{N/A} & \multirow{3}{*}{0} & 2,000 \\
\textbf{APP\_C2} &  &  &  &  & 2,000 \\
\textbf{APP\_C3} &  &  &  &  & 2,000 \\
\hline
\end{tabular}
\end{table}

\noindent
{\bf Summary.} 
\prname found thousands of erroneous behaviors in all the tested DNNs. Table~\ref{tab:violations} summarizes the numbers of erroneous behaviors found by \prname for each tested DNN while using 2,000 randomly selected seed inputs from the corresponding test sets. Note that as the testing set has similar number of samples for each class, these randomly-chosen 2,000 samples also follow that distribution.
The hyperparameter values for these experiments, as shown in Table~\ref{tab:violations}, are empirically chosen to maximize both the rate of finding difference-inputs as well as the neuron coverage achieved by these inputs.

For the experimental results shown in Figure~\ref{fig:results}, we apply three domain-specific constraints (lighting effects, occlusion by a single rectangle, and occlusion by multiple rectangles) as described in \S~\ref{subsubsec:physical-realizablity}. For all other experiments involving vision-related tasks, we only use the lighting effects as the domain-specific constraints. For all malware-related experiments, we apply all the relevant domain-specific constraints described in \S~\ref{subsubsec:physical-realizablity}. We use the hyperparameter values listed in Table~\ref{tab:violations} in all the experiments unless otherwise specified. 
%
%
%

  Figure~\ref{fig:results} shows some difference-inducing inputs generated by \prname (with different domain-specific constraints) for MNIST, ImageNet, and Driving dataset along with the corresponding erroneous behaviors. Table~\ref{tab:drebin_feat} (Drebin) and Table~\ref{tab:contagio_feat} (Contagio/VirusTotal) show two sample difference-inducing inputs generated by \prname that caused erroneous behaviors in the tested DNNs. We highlight the differences between the seed input features and the features modified by \prname. Note that we only list the top three modified features due to space limitations.

\begin{table}
\setlength{\tabcolsep}{2.5pt}
\renewcommand{\arraystretch}{1}
\centering
\footnotesize
\caption{The features added to the manifest file by \prname for generating two sample malware inputs which Android app classifiers (Drebin) incorrectly mark as benign.}
\label{tab:drebin_feat}
\begin{tabular}{|l|l|l|l|l|}
\hline
\multirow{3}{*}{\textbf{input 1}} & \textbf{feature} & \textit{\begin{tabular}[c]{@{}l@{}}feature::\\ bluetooth\end{tabular}} & \textit{\begin{tabular}[c]{@{}l@{}}activity::\\ .SmartAlertTerms\end{tabular}} & \textit{\begin{tabular}[c]{@{}l@{}}service\_receiver::\\ .rrltpsi\end{tabular}} \\ \cline{2-5} 
 & \textbf{before} & 0 & 0 & 0 \\ \cline{2-5} 
 & \textbf{after} & 1 & 1 & 1 \\ \hhline{|=|=|=|=|=|}
\multirow{3}{*}{\textbf{input 2}} & \textbf{feature} & \textit{\begin{tabular}[c]{@{}l@{}}provider::\\ xclockprovider\end{tabular}} & \textit{\begin{tabular}[c]{@{}l@{}}permission::\\ CALL\_PHONE\end{tabular}} & \textit{\begin{tabular}[c]{@{}l@{}}provider::\\ contentprovider\end{tabular}} \\ \cline{2-5} 
 & \textbf{before} & 0 & 0 & 0 \\ \cline{2-5} 
 & \textbf{after} & 1 & 1 & 1 \\ 
\hline
\end{tabular}
\end{table}

\begin{table}
\setlength{\tabcolsep}{5pt}
\renewcommand{\arraystretch}{1}
\centering
\footnotesize
\caption{The top-3 most in(de)cremented features for generating two sample malware inputs which PDF classifiers incorrectly mark as benign.}
\label{tab:contagio_feat}
\begin{tabular}{|l|l|l|l|l|}
\hline
\multirow{3}{*}{\textbf{input 1}} & \textbf{feature} & \textit{size} & \textit{count\_action} & \textit{count\_endobj} \\ \cline{2-5} 
 & \textbf{before} & 1 & 0 & 1 \\ \cline{2-5} 
 & \textbf{after} & 34 & 21 & 20 \\ 
\hhline{|=|=|=|=|=|}
\multirow{3}{*}{\textbf{input 2}} & \textbf{feature} & \textit{size} & \textit{count\_font} & \textit{author\_num} \\ \cline{2-5} 
 & \textbf{before} & 1 & 0 & 10 \\ \cline{2-5} 
 & \textbf{after} & 27 & 15 & 5 \\ 
\hline
\end{tabular}
\end{table}

\subsection{Benefits of neuron coverage} \label{subsec:nc}
In this subsection, we evaluate how effective, neuron coverage, our new metric, is in measuring the comprehensiveness of DNN testing. It has recently been shown that each neuron in a DNN tends to independently extract a specific feature of the input instead of collaborating with other neurons for feature extraction~\cite{yosinskiunderstanding, radford2017learning}. Essentially, each neuron tends to learn a different set of rules than others. This finding intuitively explains why neuron coverage is a good metric for DNN testing comprehensiveness. To empirically confirm this observation, we perform two different experiments as described below.  

First, we show that neuron coverage is a significantly better metric than code coverage for measuring comprehensiveness of the DNN test inputs. More specifically, we find that a small number of test inputs can achieve 100\% code coverage for all DNNs where neuron coverage is actually less than 34\%. Second, we evaluate neuron activations for test inputs from different classes.  Our results show that inputs from different classes tend to activate more unique neurons than inputs from the same class. Both findings confirm that neuron coverage provides a good estimation of the numbers and types of DNN rules exercised by an input.



\noindent
{\bf Neuron coverage vs. code coverage.}  We compare both code and neuron coverages achieved by the same number of inputs by evaluating the test DNNs on ten randomly picked testing samples as described in \S~\ref{subsec:dataset-neuralnet}. We measure a DNN's code coverage in terms of the line coverage of the Python code used in the training and testing process. We set the threshold $t$ in neuron coverage $0.75$, i.e., a neuron is considered covered only if its output is greater than $0.75$ for at least one input. 

Note that for the DNNs where the outputs of intermediate layers produce values in a different range than those of the final layers, we scale the neuron outputs to be within $[0,1]$ by computing $(\bm{out}-min(\bm{out})/(max(\bm{out})-min(\bm{out}))$ where $\bm{out}$ is the vector denoting the output of all neurons of a given layer.

The results, as shown in Table~\ref{tab:codecov-performance}, clearly demonstrate that neuron coverage is a significantly better metric than code coverage for measuring DNN testing comprehensiveness. Even 10 randomly picked inputs result in 100\% code coverage for all DNNs while the neuron coverage never goes above 34\% for any of the DNNs. Moreover, neuron coverage changes significantly based on the tested DNNs and the test inputs.  For example, the neuron coverage for the complete MNIST testing set (i.e., 10,000 testing samples) only reaches 57.7\%, 76.4\%, and 83.6\% for C1, C2, and C3 respectively. By contrast, the neuron coverage for the complete Contagio/Virustotal test set reaches 100\%. 

\noindent\textbf{Effect of neuron coverage on the difference-inducing inputs found by \prname.}
The primary goal behind maximizing neuron coverage as one of the objectives during the joint optimization process is to generate \textit{diverse} difference-inducing inputs as discussed in \S~\ref{sec:overview}.
In this experiment, we evaluate the effectiveness of neuron coverage at achieving this goal.

We randomly pick 2,000 seed inputs from MNIST test dataset and use \prname to generate difference-inducing inputs with and without neuron coverage by setting $\lambda_2$ in Equation~\ref{eq:objective 1} to $1$ and $0$ respectively. 
We measure the \textit{diversity} of the generated difference-inducing inputs in terms of averaged L1 distance between all difference-inducing inputs generated from the same seed and the original seed.
The L1-distance calculates the sum of absolute differences of each pixel values between the generated image and the original one. Table~\ref{tab:nc_effect} shows the results of three such experiments. The results clearly show that neuron coverage helps in increasing the diversity of generated inputs. 

Note that even though the absolute value of the increase in neuron coverage achieved by setting $\lambda_2=1$ instead of $\lambda_2=0$ may seem small (e.g., 1-2 percentage points), it has a significant effect on increasing the diversity of the generated difference-inducing images as shown in Table~\ref{tab:nc_effect}. These results show that increasing neuron coverage,  similar to code coverage, becomes increasingly harder for higher values but even small increases in neuron coverage can improve the test diversity significantly. Also, the numbers of difference-inducing inputs generated with $\lambda_2=1$ are less than those for $\lambda_2=0$ as setting $\lambda_2=1$ causes \prname to focus on finding diverse differences rather than simply increasing the number of differences with the same underlying root cause. In general, the number of difference-inducing inputs alone is a not a good metric for measuring the quality of the generated tests for vision-related tasks as one can create a large number of difference-inducing images with the same root cause by making tiny changes to an existing difference-inducing image.


\begin{table}
\setlength{\tabcolsep}{2pt}
\footnotesize
\renewcommand{\arraystretch}{1.1}
\centering
\caption{The increase in \textit{diversity} (L1-distance) in the difference-inducing inputs found by \prname while using neuron coverage as part of the optimization goal (Equation~\ref{eq:objective 1}). This experiment 
uses 2,000 randomly picked seed inputs from the MNIST dataset. Higher values denote larger diversity. NC denotes the neuron coverage (with $t=0.25$) achieved under each setting. } 
\label{tab:nc_effect}
\begin{tabular}{|c|c|c|c|c|c|c|}
\hline
\multirow{2}{*}{\textbf{Exp. \#}} & \multicolumn{3}{c|}{\textbf{$\bm{\lambda_2}=0$ (w/o neuron coverage)}} & \multicolumn{3}{c|}{\textbf{$\bm{\lambda_2}=1$ (with neuron coverage)}} \\ \cline{2-7} 
 & \textit{Avg. diversity} & \textit{NC} & \textit{\# Diffs} & \textit{Avg. diversity} & \textit{NC} & \textit{\# Diffs} \\ \hline
1 & 237.9 & 69.4\% & 871 & 283.3 & 70.6\% & 776 \\ \hline
2 & 194.6 & 66.7\% & 789 & 253.2 & 67.8\% & 680 \\ \hline
3 & 170.8 & 68.9\% & 734 & 182.7 & 70.2\% & 658 \\ \hline
\end{tabular}
\end{table}



\begin{table}
\setlength{\tabcolsep}{5pt}
\footnotesize
\renewcommand{\arraystretch}{1}
\centering
\caption{Comparison of code coverage and neuron coverage for $10$ randomly selected inputs from the original test set of each DNN.}
\label{tab:codecov-performance}
\begin{tabular}{|l|l|l|l|l|l|l|l|}
\hline
\multirow{2}{*}{\textbf{Dataset}} & \multicolumn{3}{l|}{\textbf{Code Coverage}} & \multicolumn{3}{l|}{\textbf{Neuron Coverage}} \\ \cline{2-7} 
 & \textit{C1} & \textit{C2} & \textit{C3} & \textit{C1} & \textit{C2} & \textit{C3} \\ \hhline{|=|=|=|=|=|=|=|}
\textbf{MNIST} & 100\% & 100\% & 100\% & 32.7\% & 33.1\% & 25.7\% \\ \hline
\textbf{ImageNet} & 100\% & 100\% & 100\% & 1.5\% & 1.1\% & 0.3\% \\ \hline
\textbf{Driving} & 100\% & 100\% & 100\% & 2.5\% & 3.1\% & 3.9\% \\ \hline
\textbf{VirusTotal} & 100\% & 100\% & 100\% & 19.8\% & 17.3\% & 17.3\% \\ \hline
\textbf{Drebin} & 100\% & 100\% & 100\% & 16.8\% & 10\% & 28.6\% \\ \hline
\end{tabular}
\end{table}

\noindent
{\bf Activation of neurons for different classes of inputs.} In this experiment, we measure the number of active neurons that are common across the LeNet-5 DNN running on pairs of MNIST inputs of the same and different classes respectively. In particular, we randomly select 200 input pairs where 100 pairs have the same label (e.g., labeled as $8$) and 100 pairs have different labels (e.g., labeled as $8$ and $4$).  Then, we calculate the number of common (overlapped) active neurons for these input pairs.
 Table~\ref{tab:overlap} shows the results, which confirm our hypothesis that inputs coming from the same class share more activated neurons than those coming from different classes. As inputs from different classes tend to get detected through matching of different DNN rules, our result also confirms that neuron coverage can effectively estimate the numbers of different rules activated during DNN testing.


 

\begin{table}
\setlength{\tabcolsep}{4pt}
\footnotesize
\renewcommand{\arraystretch}{1}
\centering
\caption{Average number of overlaps among activated neurons for a pair of inputs of the same class and different classes. Inputs of different classes tend to activate different neurons.}
\label{tab:overlap}
\begin{tabular}{|l|l|l|l|}
\hline
\textbf{} & \textit{\textbf{Total neurons}} & \textit{\textbf{\begin{tabular}[c]{@{}l@{}}Avg. no. of\\ activated neurons\end{tabular}}} & \textit{\textbf{Avg. overlap}} \\ \hline
\textbf{Diff. class} & 268 & 83.6 & 45.9 \\ \hline
\textbf{Same class} & 268 & 84.1 & 74.2 \\ \hline
\end{tabular}
\end{table}



\vspace{-0.2cm}
\subsection{Performance}
\label{subsec:performance}
We evaluate \prname's performance using two metrics: neuron coverage of the generated tests and execution time for generating difference-inducing inputs. 

\begin{figure*}
\centering
\captionsetup[subfloat]{captionskip=3pt}
\subfloat[MNIST]{
\includegraphics[width=0.18\textwidth]{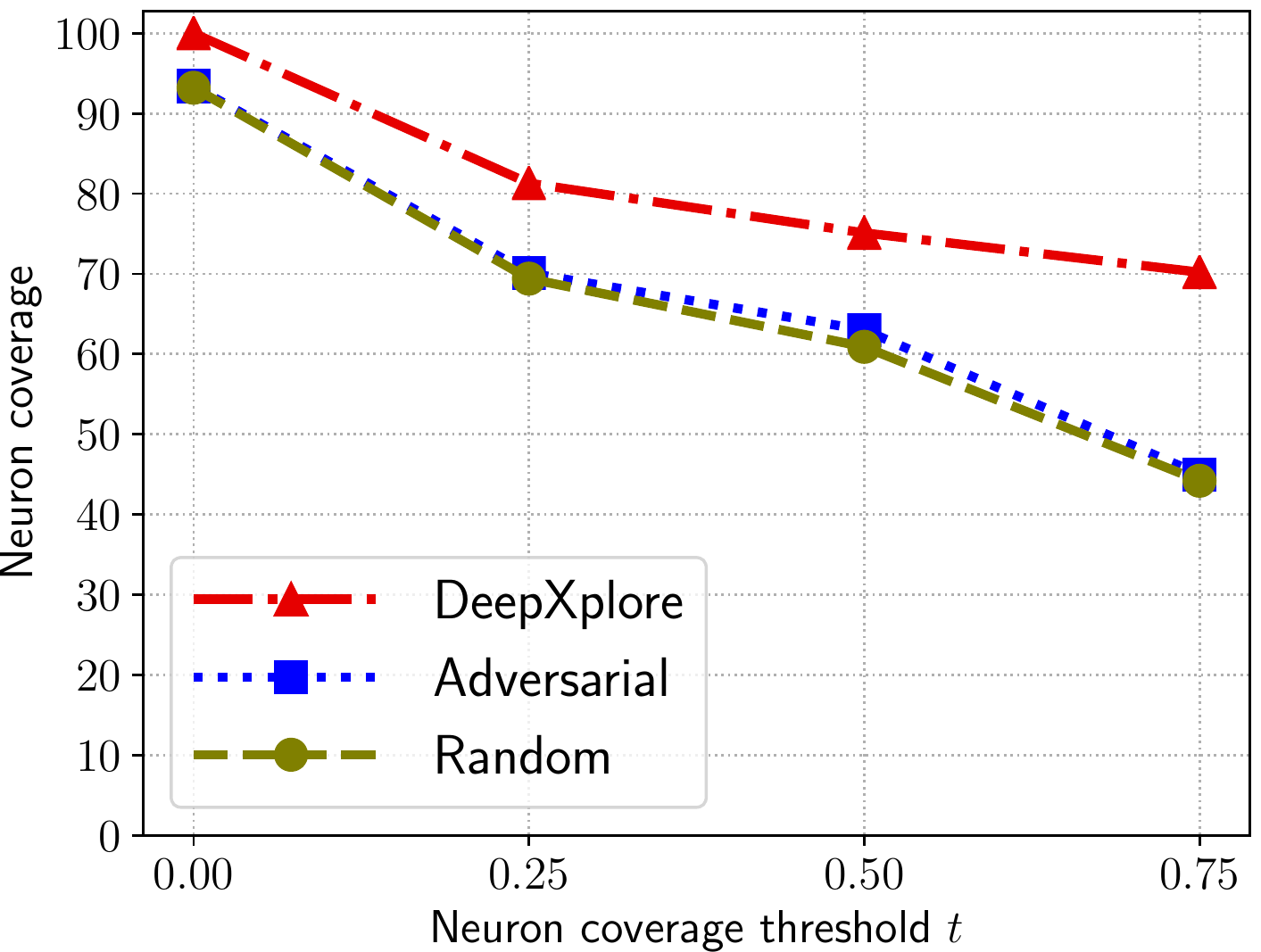}
\label{subfig:mnist}}
\hfill
\subfloat[ImageNet]{
\includegraphics[width=0.18\textwidth]{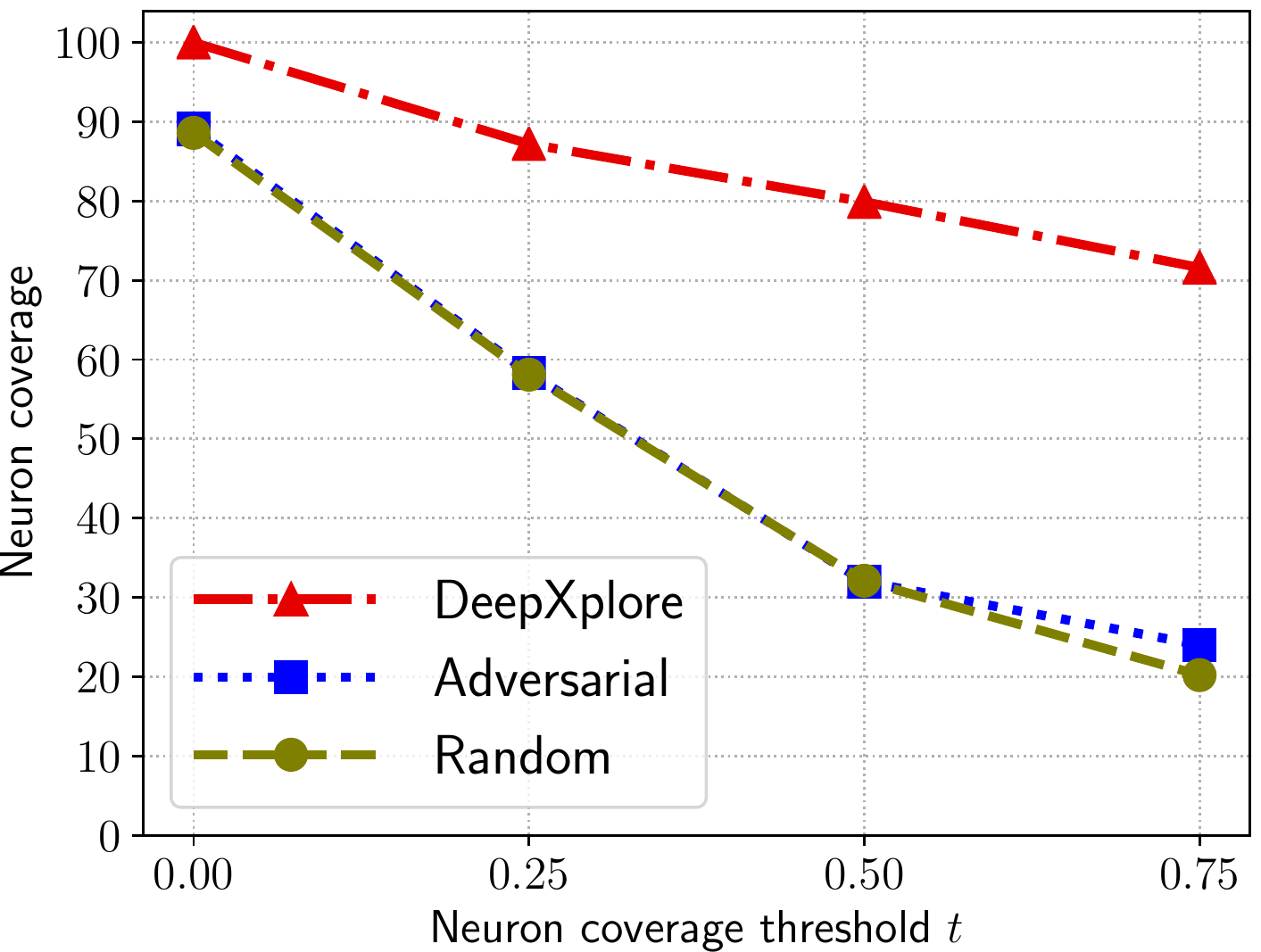}
\label{subfig:imageNet}}
\hfill
\subfloat[Driving]{
\includegraphics[width=0.18\textwidth]{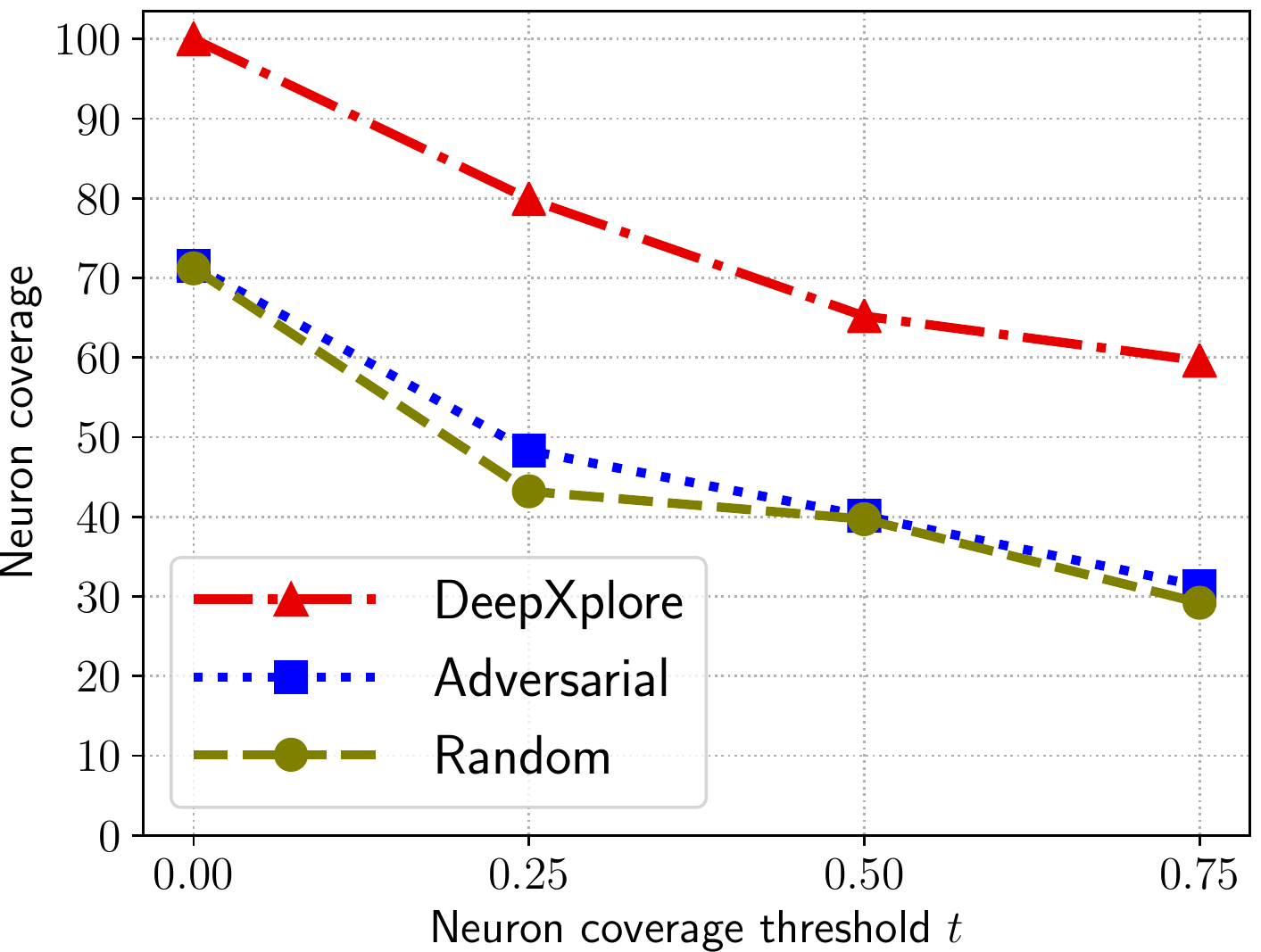}
\label{subfig:driving}}
\hfill
\subfloat[VirusTotal]{
\includegraphics[width=0.18\textwidth]{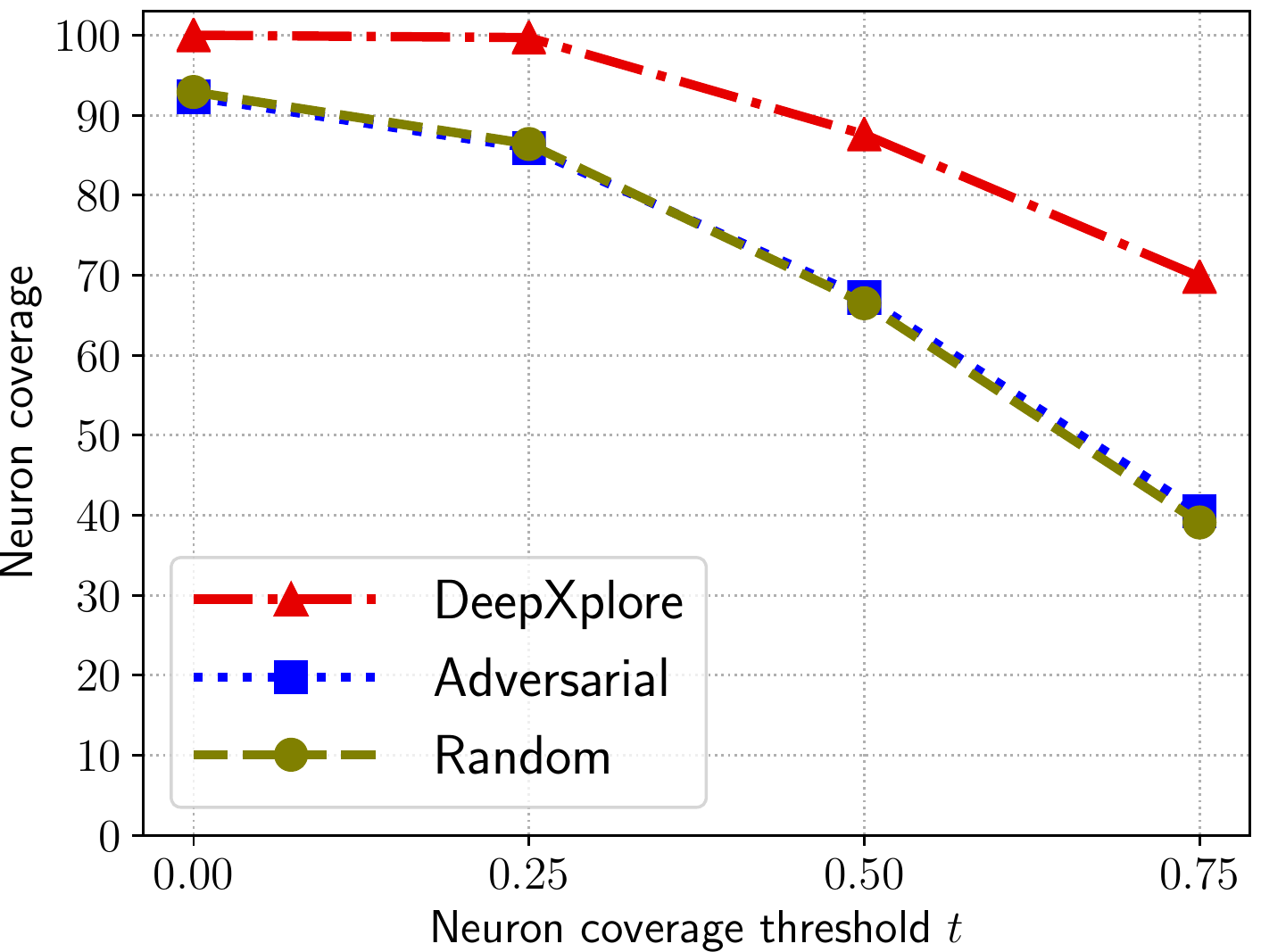}
\label{subfig:contagio}}
\hfill
\subfloat[Drebin]{
\includegraphics[width=0.18\textwidth]{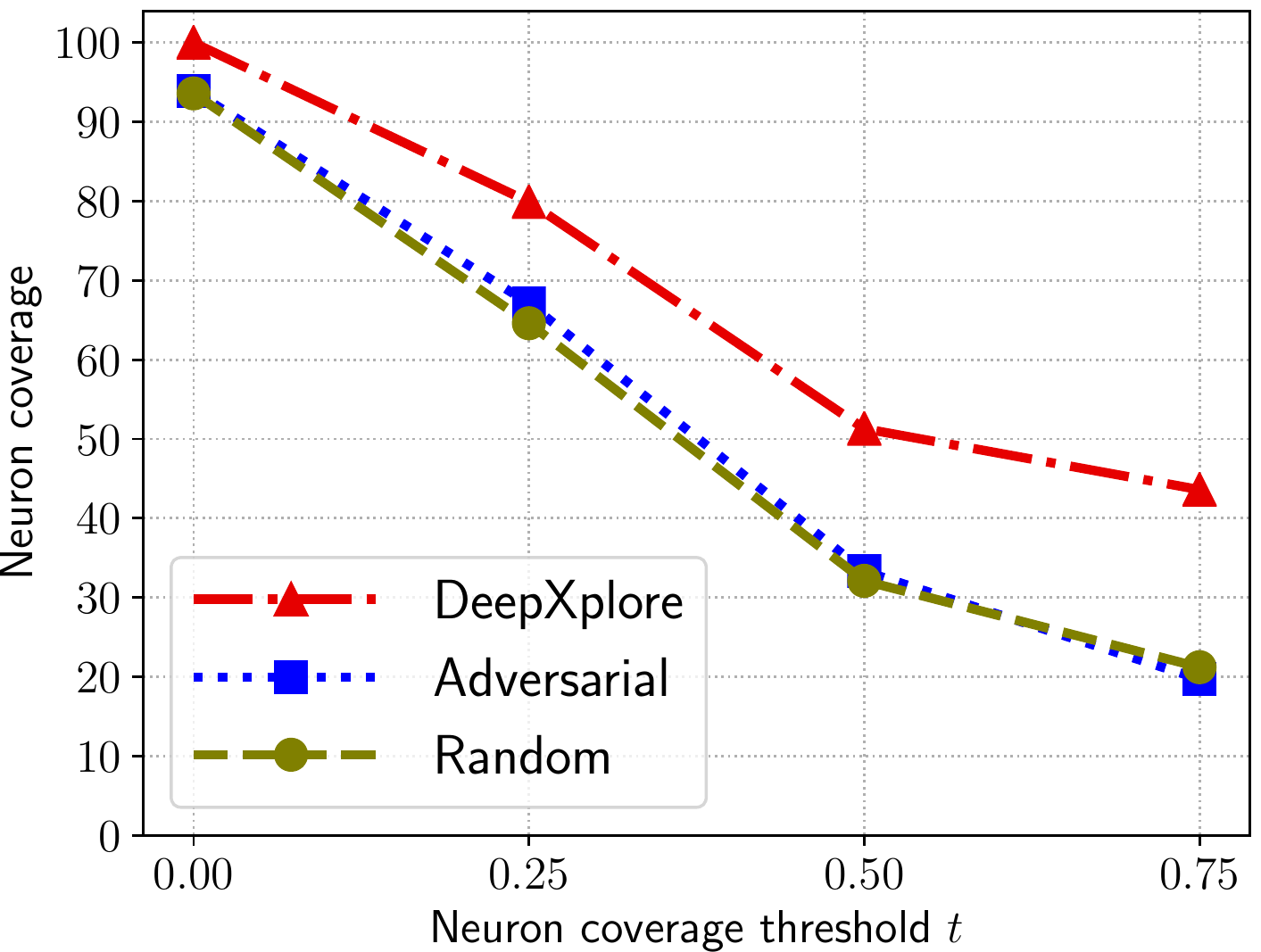}
\label{subfig:drebin}}
\caption{The neuron coverage achieved by the same number of inputs (1\% of the original test set) produced by \prname, adversarial testing~\cite{goodfellow2014explaining}, and random selection from the original test set. The plots show the changes in neuron coverage for all three methods as the threshold $t$ (defined in \S~\ref{sec:approach}) increases. \prname, on average, covers 34.4\% and 33.2\% more neurons than random testing and adversarial testing.}
\label{fig:percent_coverage}
\vspace{-.2cm}
\end{figure*}

\noindent
{\bf Neuron coverage.} In this experiment, we compare the neuron coverage achieved by the same number of tests generated by three different approaches: (1) \prname, (2) adversarial testing~\cite{goodfellow2014explaining}, and (3) random selection from the original test set. 
 The results are shown in Table~\ref{tab:performance} and Figure~\ref{fig:percent_coverage}.  
 
 We can make two key observations from the results.  First, \prname, on average, covers 34.4\% and 33.2\% more neurons than random testing and adversarial testing as demonstrated in Figure~\ref{fig:percent_coverage}. Second, the neuron coverage threshold $t$ (defined in \S~\ref{sec:approach}), which decides when a neuron has been activated, greatly affects the achieved neuron coverage.  As the threshold $t$ increases, all three approaches cover fewer neurons. This is intuitive as a higher value of $t$  makes it increasingly harder to activate neurons using simple modifications.

\noindent
{\bf Execution time and number of seed inputs.} For this experiment, we measure the execution time of \prname to generate difference-inducing inputs with 100\% neuron coverage for all the tested DNNs. 
We note that some neurons in fully-connected layers of DNNs on MNIST, ImageNet and Driving are very hard to activate, we thus only consider neuron coverage on layers except fully-connected layers.
Table~\ref{tab:performance} shows the results, which indicate that \prname is very efficient in terms of finding difference-inducing inputs as well as increasing neuron coverage.


\begin{table}
\setlength{\tabcolsep}{6pt}
\footnotesize
\renewcommand{\arraystretch}{1}
\centering
\caption{Total time taken by \prname to achieve 100\% neuron coverage for different DNNs averaged over 10 runs. The last column shows the number of seed inputs.}
\label{tab:performance}
\begin{tabular}{|c|c|c|c|c|}
\hline
\textbf{} & \textbf{C1} & \textbf{C2} & \textbf{C3} & \textbf{\# seeds} \\ \hline
\textbf{MNIST} & 6.6 s & 6.8 s & 7.6 s & 9 \\ \hline
\textbf{ImageNet} & 43.6 s & 45.3 s & 42.7 s & 35 \\ \hline
\textbf{Driving} & 11.7 s & 12.3 s & 9.8 s & 12 \\ \hline
\textbf{VirusTotal} & 31.1 s & 29.7 s & 23.2 s & 6 \\ \hline
\textbf{Drebin} & 180.2 s & 196.4 s & 152.9 s & 16 \\ \hline
\end{tabular}
\end{table}

\noindent
{\bf Different choices of hyperparameters.}  We further evaluate how the choices of different hyperparameters of \prname ($s$, $\lambda_1$, $\lambda_2$, and $t$ as described in \S~\ref{subsec:algo}) influence \prname's performance. The effects of changing neuron activation threshold $t$ was shown in Figure~\ref{fig:percent_coverage} as described earlier.
%
Tables~\ref{tab:step-performance},~\ref{tab:lambda1-performance},~and~\ref{tab:lambda2-performance} show the variations in \prname runtime with changes in $s$, $\lambda_1$, and $\lambda_2$ respectively. Our results show that the optimal values of $s$ and $\lambda_1$ vary across the DNNs and datasets, while $\lambda_2=0.5$ tend to be optimal for all the datasets. 

We use the time taken by \prname to find the first difference-inducing input as the metric for comparing different choices of hyperparameters in Tables~\ref{tab:step-performance},~\ref{tab:lambda1-performance}, and~\ref{tab:lambda2-performance}. We choose this metric as we observe that finding the first difference-inducing input for a given seed tend to be significantly harder than increasing the number of difference-inducing inputs.


\begin{table}
\setlength{\tabcolsep}{6pt}
\footnotesize
\renewcommand{\arraystretch}{1}
\centering
\caption{The variation in \prname runtime (in seconds) while generating the first difference-inducing input for the tested DNNs with different step size choice ($s$ for gradient ascent shown in Algorithm~\ref{alg:testgen} line 14). All numbers averaged over 10 runs. The fastest times for each dataset is highlighted in gray.}
\label{tab:step-performance}
\begin{tabular}{|l|l|l|l|l|l|}
\hline
\textbf{} & \textit{s=0.01} & \textit{s=0.1} & \textit{s=1} & \textit{s=10} & \textit{s=100} \\ \hhline{|=|=|=|=|=|=|}
\textbf{MNIST} & \cellcolor[HTML]{C0C0C0}0.19 s & 0.26 s & 0.65 s & 0.62 s & 0.65 s \\ \hline
\textbf{ImageNet} & 9.3 s & 4.78 s & 1.7 s & \cellcolor[HTML]{C0C0C0}1.06 s & 6.66 s \\ \hline
\textbf{Driving} & 0.51 s & \cellcolor[HTML]{C0C0C0}0.5 s & 0.53 s & 0.55 s & 0.54 s \\ \hline
\textbf{VirusTotal} & 0.17 s & \cellcolor[HTML]{C0C0C0}0.16 s & 0.21 s & 0.21 s & 0.22 s \\ \hline
\textbf{Drebin} & \cellcolor[HTML]{C0C0C0}7.65 s & 7.65 s & 7.65 s & 7.65 s & 7.65 s \\ \hline
\end{tabular}
\end{table}

\begin{table}
\setlength{\tabcolsep}{6pt}
\footnotesize
\renewcommand{\arraystretch}{1}
\centering
\caption{The variation in \prname runtime (in seconds) while generating the first difference-inducing input for the tested DNNs with different $\lambda_1$, a parameter in Equation~\ref{eq:objective 1}.  Higher $\lambda_1$ values indicate prioritization of minimizing a DNNs' outputs over maximizing the outputs of other DNNs showing differential behavior. The fastest times for each dataset is highlighted in gray.}
\label{tab:lambda1-performance}
\begin{tabular}{|l|l|l|l|l|}
\hline
\textbf{} & $\lambda_1=0.5$ & $\lambda_1=1$ & $\lambda_1=2$ & $\lambda_1=3$ \\ \hline
\textbf{MNIST} & 0.28 s & 0.25 s & 0.2 s & \cellcolor[HTML]{C0C0C0}0.17 s \\ \hline
\textbf{ImageNet} & 1.38 s & 1.26 s & \cellcolor[HTML]{C0C0C0}1.21 s & 1.72 s \\ \hline
\textbf{Driving} & 0.62 s & 0.59 s & \cellcolor[HTML]{C0C0C0}0.57 s & 0.58 s \\ \hline
\textbf{VirusTotal} & 0.13 s & 0.12 s & \cellcolor[HTML]{C0C0C0}0.05 s & 0.09 s \\ \hline
\textbf{Drebin} & 6.4 s & \cellcolor[HTML]{C0C0C0}5.8 s & 6.12 s & 7.5 s \\ \hline
\end{tabular}
\end{table}

\begin{table}
\setlength{\tabcolsep}{6pt}
\footnotesize
\renewcommand{\arraystretch}{1}
\centering
\caption{The variation in \prname runtime (in seconds) while generating the first difference-inducing input for the tested DNNs with different $\lambda_2$, a parameter in Equation~\ref{eq:objectives}. Higher $\lambda_2$ values indicate higher priority for increasing coverage. All numbers averaged over 10 runs. The fastest times for each dataset is highlighted in gray.}
\label{tab:lambda2-performance}
\begin{tabular}{|l|l|l|l|l|}
\hline
\textbf{} & \textit{$\lambda_2=0.5$} & \textit{$\lambda_2=1$} & \textit{$\lambda_2=2$} & \textit{$\lambda_2=3$} \\ \hline
\textbf{MNIST} & \cellcolor[HTML]{C0C0C0}0.19 s & 0.22 s & 0.23 s & 0.26 s \\ \hline
\textbf{ImageNet} & \cellcolor[HTML]{C0C0C0}1.55 s & 1.6 s & 1.62 s & 1.85 s \\ \hline
\textbf{Driving} & \cellcolor[HTML]{C0C0C0}0.56 s & 0.61 s & 0.67 s & 0.74 s \\ \hline
\textbf{VirusTotal} & \cellcolor[HTML]{C0C0C0}0.09 s & 0.15 s & 0.21 s & 0.25 s \\ \hline
\textbf{Drebin} & \cellcolor[HTML]{C0C0C0}6.14 s & 6.75 s & 6.82 s & 6.83 s \\ \hline
\end{tabular}
\end{table}

\noindent\textbf{Testing very similar models with \prname.}
Note that while \prname's gradient-guided test generation process works very well in practice, it may fail to find any difference-inducing inputs within a reasonable time for some cases especially for DNNs with very similar decision boundaries. To estimate how similar two DNNs have to be in order to make \prname to fail in practice, we control three types of differences between two DNNs and measure the changes in iterations required to generate the first difference-inducing inputs in each case.

We use MNIST training set (60,000 samples) and LeNet-1 trained with 10 epochs as the control group.
We change the (1) number of training samples, (2) number of filters per convolutional layer, and (3) number of training epochs respectively to create variants of LeNet-1. Table~\ref{tab:differences} summarizes the averaged number of iterations (over 100 seed inputs) needed by \prname to find the first difference inducing inputs between these LeNet-1 variants and the original version. Overall, we find that \prname is very good at finding differences even between DNNs with minute variations (only failing once as shown in Table~\ref{tab:differences}). As the number of differences goes down, the number of iterations to find a difference-inducing input goes up, \ie it gets increasingly harder to find difference-inducing tests between DNNs with smaller differences.


\begin{table}
\setlength{\tabcolsep}{5pt}
\footnotesize
\renewcommand{\arraystretch}{1.1}
\centering
\caption{Changes in the number of iterations \prname takes, on average, to find the first difference inducing inputs as the type and numbers of differences between the test DNNs increase.}
\label{tab:differences}
\begin{tabular}{|l|l|l|l|l|l|l|}
\hline
\multirow{2}{*}{\textbf{\begin{tabular}[c]{@{}l@{}}Training\\ Samples\end{tabular}}} & \textit{\# diff} & \textit{0} & 1 & \textit{100} & \textit{1000} & \textit{10000} \\ \cline{2-7} 
& \textit{\# iter} & -$^{*}$& -$^{*}$ & 616.1 & 503.7 & 256.9 \\ \hhline{|=|=|=|=|=|=|=|}
\multirow{2}{*}{\textbf{\begin{tabular}[c]{@{}l@{}}Neurons\\ per layer\end{tabular}}} & \textit{\# diff} & \textit{0} & 1 & \textit{2} & \textit{3} & \textit{4} \\ \cline{2-7} 
& \textit{\# iter} & -$^{*}$ & 69.6 & 53.9 & 33.1 & 18.7 \\ \hhline{|=|=|=|=|=|=|=|}
\multirow{2}{*}{\textbf{\begin{tabular}[c]{@{}l@{}}Training\\ Epochs\end{tabular}}} & \textit{\# diff} & \textit{0} & 5 & \textit{10} & \textit{20} & \textit{40} \\ \cline{2-7} 
& \textit{\# iter} & -$^{*}$ & 453.8 & 433.9 & 348.7 & 210 \\ \hline
\multicolumn{6}{l}{\scriptsize\begin{tabular}[c]{@{}l@{}}*- indicates timeout after $1000$ iterations
\end{tabular}}
\end{tabular}
\end{table}

\vspace{-0.2cm}
\subsection{Improving DNNs with \prname}
\label{subsec:other_usage}

In this section, we demonstrate two additional applications of the error-inducing inputs generated by \prname: augmenting training set and then improve DNN's accuracy and detecting potentially corrupted training data.


\noindent \textbf{Augmenting training data to improve accuracy.} We augment the original training data of a DNN with the error-inducing inputs generated by \prname for retraining the DNN to fix the erroneous behaviors and therefore improve its accuracy.  Note that such strategy has also been adopted for fixing a DNN's behavior for adversarial inputs~\cite{goodfellow2014explaining}---but the key difference is that adversarial testing requires manual labeling while \prname can adopt majority voting~\cite{freund1995desicion} to automatically generate labels for the generated test inputs. Note that the underlying assumption is that the decision made by the majority of the DNNs are more likely to be correct.

\begin{figure}[t]
\centering
\subfloat[LeNet-1]{
\includegraphics[width=0.32\columnwidth]{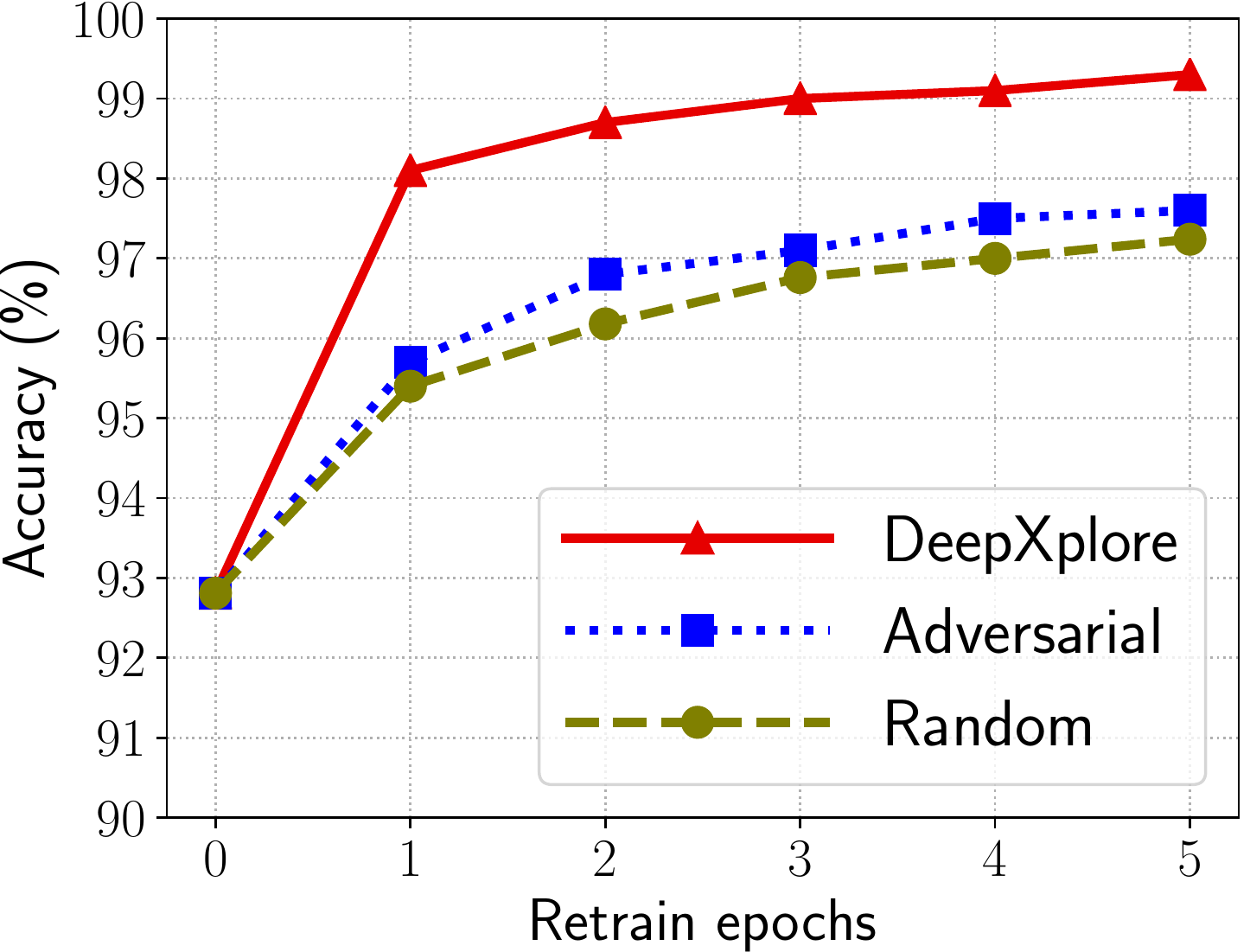}}
\hfill
\subfloat[LeNet-4]{
\includegraphics[width=0.32\columnwidth]{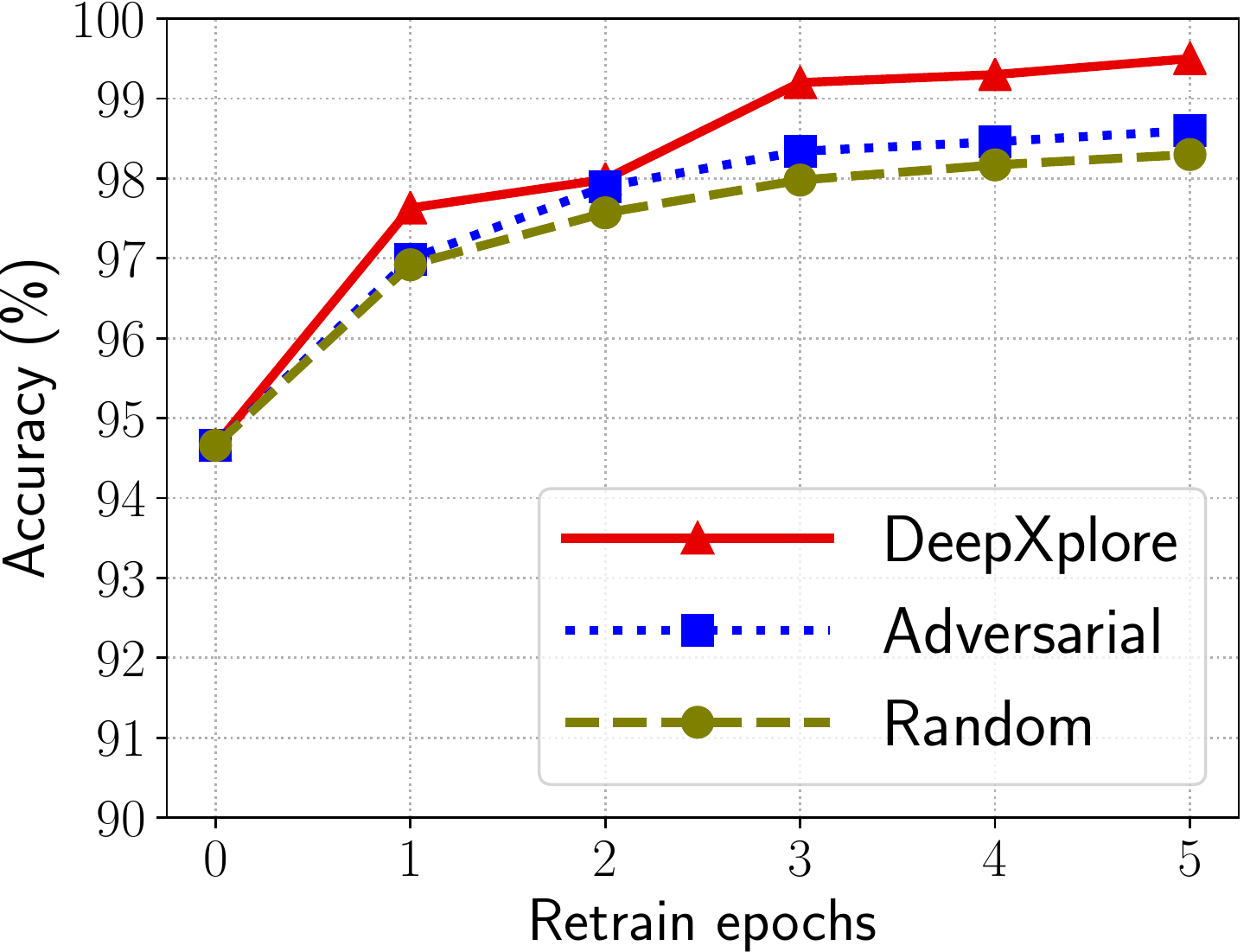}}
\hfill
\subfloat[LeNet-5]{
\includegraphics[width=0.32\columnwidth]{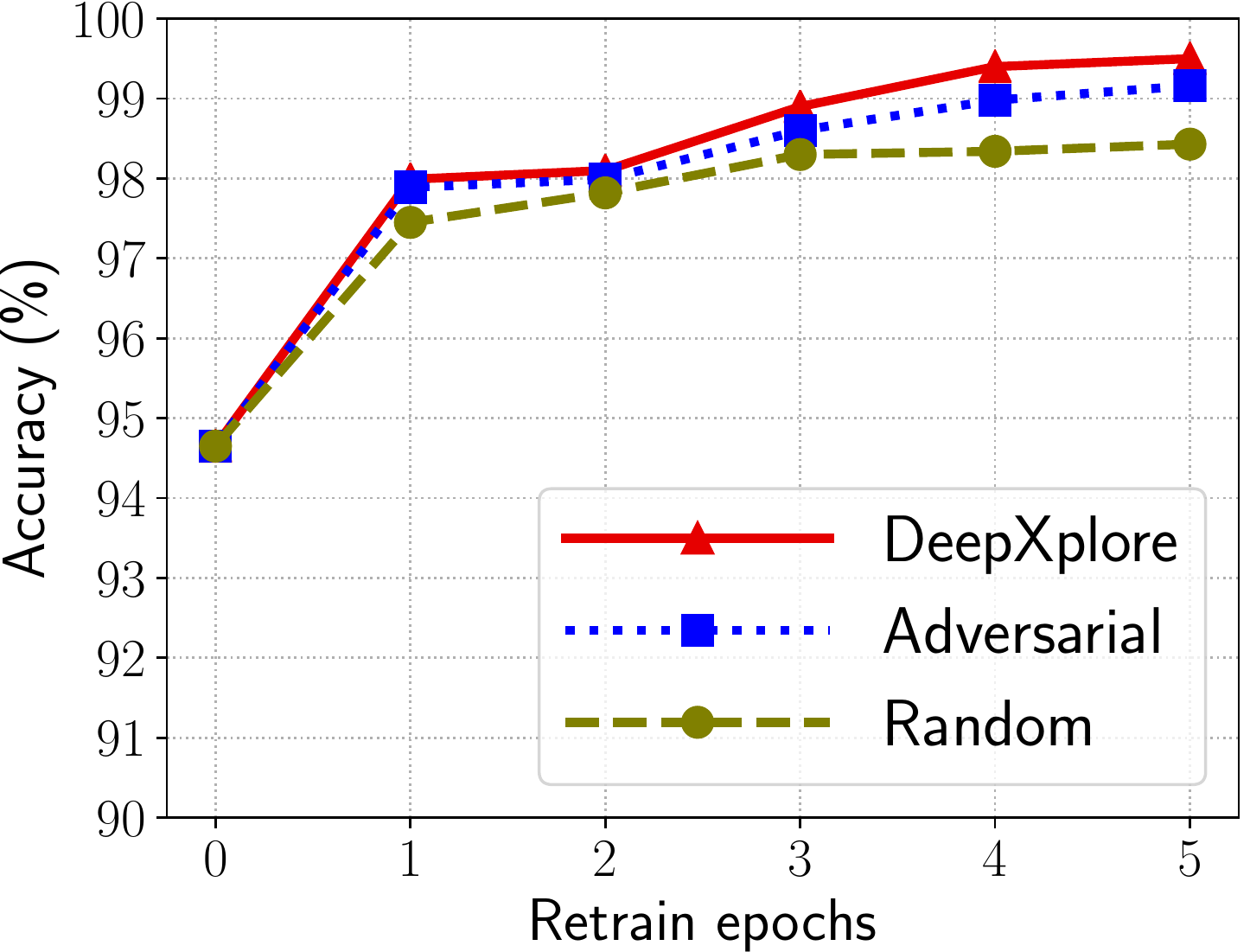}}
\caption{Improvement in accuracy of three LeNet DNNs when the training set is augmented with the same number of inputs generated by random selection (``random''), adversarial testing (``adversarial'')~\cite{goodfellow2014explaining}, and \prname. 
}
\label{fig:retrain}
\end{figure}

To evaluate this approach, we train LeNet-1, LeNet-4, and LeNet-5 as shown in Table~\ref{tab:architecture} with $60,000$ original samples. We further augment the training data by adding $100$ new error-inducing samples and retrain the DNNs by 5 epochs. Our experiment results---comparing three approaches, \ie random selection (``random''), adversarial testing (``adversarial'') and \prname---are shown in Figure~\ref{fig:retrain}.  
The results show that \prname achieved 1--3\% more average accuracy improvement over adversarial and random augmentation. 



\noindent \textbf{Detecting training data pollution attack.} As another application of \prname, we demonstrate how it can be used to detect training data pollution attacks with an experiment on two LeNet-5 DNNs: one trained on $60,000$ hand-written digits from MNIST dataset and the other trained on an artificially polluted version of the same dataset where 30\% of the images originally labeled as digit $9$ are mislabeled as $1$.  We use \prname to generate error-inducing inputs that are classified as the digit $9$ and $1$ by the unpolluted and polluted versions of the LeNet-5 DNN respectively. We then search for samples in the training set that are closest to the inputs generated by \prname in terms of structural similarity~\cite{wang2004image} and identify them as polluted data.  Using this process, we are able to correctly identify 95.6\% of the polluted samples.

\vspace{-0.3cm}
\section{Discussion}
\label{sec:discussion}
\noindent\textbf{Causes of differences between DNNs.}
The underlying root cause behind prediction differences between two DNNs for the same input is differences in their decision logic/boundaries. As described in \S~\ref{subsec:dl_systems}, a DNN's decision logic is determined by multiple factors including training data, the DNN architecture, hyperparameters, etc.  Therefore, any differences in the choice of these factors will result in subtle changes in the decision logic of the resulting DNN. As we empirically demonstrated in Table~\ref{tab:differences}, the more similar the decision boundaries of two DNNs are, the harder it is to find difference-inducing inputs. However, all the real-world DNNs that we tested tend to have significant differences and therefore \prname can efficiently find erroneous behaviors in all of the tested DNNs.

\noindent\textbf{Overhead of training vs. testing DNNs.}
There is a significant performance asymmetry between the prediction/gradient computation and training of large real-world DNNs. For example, training a state-of-the-art DNN like VGG-16~\cite{simonyan2014very} (one of the tested DNNs in this paper) on 1.2 million images in ImageNet dataset~\cite{ILSVRC15} competitions) can take up to $7$ days on a single GTX 1080 Ti GPU. By contrast, the prediction and gradient computations on the same GPU take around 120 milliseconds in total per image. Such massive performance difference between training and prediction for large DNNs make \prname especially suitable for testing large, pre-trained DNNs. 

\noindent\textbf{Limitations.}
\prname adopts the technique of differential testing from software analysis and thus  inherits the limitations of differential testing. We summarize them briefly below. 

First, differential testing requires at least two different DNNs with the same functionality. Further, if two DNNs only differ slightly (i.e., by a few neurons), \prname will take longer to find difference-inducing inputs than if the DNNs were significantly different from each other as shown in Table~\ref{tab:differences}. However, our evaluation shows that in most cases multiple different DNNs, for a given problem, are easily available as developers often define and train their own DNNs for customization and improved accuracy. 

%

Second, differential testing can only detect an erroneous behavior if at least one DNN produces different results than other DNNs.  If all the tested DNNs make the same mistake, \prname cannot generate the corresponding test case.  However, we found this to be not a significant issue in practice as most DNNs are independently constructed and trained, the odds of all of them making the same mistake is low. 



\vspace{-0.3cm}
\section{Related Work}
\label{sec:related}
\noindent{\bf Adversarial deep learning.}  \hspace{0.02in}
Recently, the security and privacy aspects of machine learning have drawn significant attention from the researchers in both machine learning~\cite{szegedy2013intriguing,goodfellow2014explaining,nguyen2015deep} and security~\cite{cao2015towards, adversarial:ccs16, wu2016methodology, perdisci2006misleading, fredrikson2015model, fredrikson2014privacy, 197128} communities. Many of these works have demonstrated that a DNN can be fooled by applying minute perturbations to an input image, which was originally classified correctly by the DNN, even though the modified image looks visibly indistinguishable from the original image to the human eye. 


Adversarial images demonstrate a particular type of erroneous behaviors of DNNs. However, they suffer from two major limitations: (1) they have low neuron coverage (similar to the randomly selected test inputs as shown in Figure~\ref{fig:percent_coverage}) and therefore, unlike \prname, can not expose different types of erroneous behaviors; and (2) the adversarial image generation process is inherently limited to only use the tiny, undetectable perturbations as any visible change will require manual inspection. \prname bypasses this issue by using differential testing and therefore can perturb inputs to create many realistic visible differences (\eg different lighting, occlusion, etc.) and automatically detect erroneous behaviors of DNNs under these circumstances.   

\noindent{\bf Testing and verification of DNNs.}
Traditional practices in evaluating machine learning systems primarily measure their accuracy on randomly drawn test inputs from manually labeled datasets~\cite{witten2016data}. Some machine learning systems like autonomous vehicles leverage ad hoc unguided simulations~\cite{waymo_simulation, waymoreport}. 
However, without the knowledge of the model's internals, such blackbox testing paradigms are not able to find different corner cases that induce erroneous behaviors~\cite{verify-test}. This observation has inspired several researchers to try to improve the robustness and reliability of DNNs~\cite{zhang2013toward, gu2014towards, shaham2015understanding, bastani2016measuring, zheng2016improving, huang2017safety, carlini2017towards, cisse2017parseval, papernot2016distillation, xu2017feature, metzen2017detecting, papernot2017extending}. However, all of these projects only focus on adversarial inputs and rely on the ground truth labels provided manually. By contrast, our technique can systematically test the robustness and reliability of DL systems for a broad range of flaws in a fully automated manner without any manual labeling.


Another recent line of work has explored the possibility of formally verifying DNNs against different safety properties~\cite{pulina2010abstraction, huang2017safety, katz2017reluplex}. None of these techniques scale well to find violations of interesting safety properties for real-world DNNs. By contrast, \prname can find interesting erroneous behaviors in large, state-of-the-art DNNs but cannot provide any guarantee about whether a specific DNN satisfies a given safety property.


\noindent{\bf Other applications of DNN gradients. } 
Gradients have been used in the past for visualizing activation of different intermediate layers of a DNN for tasks like object segmentation~\cite{mahendran2015understanding, simonyan2014very}, artistic style transfer between two images~\cite{gatys2015neural, li2016combining, ruder2016artistic}, etc. By contrast, in this paper, we apply gradient ascent for solving the joint optimization problem that maximizes both neuron coverage and the number of differential behaviors among tested DNNs.

 

\noindent{\bf Differential testing of traditional software.} 
Differential testing has been widely used for successfully testing various types of traditional software
including JVMs~\cite{chen_2016_pldi_difftest_jvm}, C compilers~\cite{mckeeman1998differential, yang2011finding}, 
SSL/TLS certification validation logic~\cite{frankencert, mucert, jana:nezha, jana:hvlearn}, PDF viewers~\cite{jana:nezha}, space flight software~\cite{groce2007randomized}, mobile applications~\cite{jung2008privacy}, and Web application firewalls~\cite{sfadiff}.  

The key advantage of applying differential testing to DNNs over traditional software is that the problem of finding a large number of difference-inducing inputs while simultaneously maximizing neuron coverage can be expressed as a well defined joint optimization problem. Moreover, the gradient of a DNN with respect to the input can be utilized for efficiently solving the optimization problem using gradient ascent.
 

\vspace{-0.4cm}
\section{Conclusion}
\label{sec:conclusion}
We designed and implemented \prname, the first whitebox system for systematically testing DL systems and automatically identify erroneous behaviors without manual labels. We introduced a new metric, neuron coverage, for measuring how many rules in a DNN are exercised by a set of inputs. \prname performs gradient ascent to solve a joint optimization problem that maximizes both neuron coverage and the number of potentially erroneous behaviors. \prname was able to find thousands of erroneous behaviors in fifteen state-of-the-art DNNs trained on five real-world datasets. 

\vspace{-0.4cm}
\begin{acks}
We would like to thank Byung-Gon Chun (our shepherd), Yoav Hollander, Daniel Hsu, Murat Demirbas, Dave Evans, and the anonymous reviewers for their helpful feedback. This work was supported in part by NSF grants CNS-16-17670, CNS-15-63843, and CNS-15-64055; ONR grants N00014-17-1-2010, N00014-16-1-2263, and N00014-17-1-2788; and a Google Faculty Fellowship.
\end{acks}
%

\bibliographystyle{ACM-Reference-Format}
\bibliography{ref}


\begin{thebibliography}{90}


\ifx \showCODEN    \undefined \def \showCODEN     #1{\unskip}     \fi
\ifx \showDOI      \undefined \def \showDOI       #1{#1}\fi
\ifx \showISBNx    \undefined \def \showISBNx     #1{\unskip}     \fi
\ifx \showISBNxiii \undefined \def \showISBNxiii  #1{\unskip}     \fi
\ifx \showISSN     \undefined \def \showISSN      #1{\unskip}     \fi
\ifx \showLCCN     \undefined \def \showLCCN      #1{\unskip}     \fi
\ifx \shownote     \undefined \def \shownote      #1{#1}          \fi
\ifx \showarticletitle \undefined \def \showarticletitle #1{#1}   \fi
\ifx \showURL      \undefined \def \showURL       {\relax}        \fi
\providecommand\bibfield[2]{#2}
\providecommand\bibinfo[2]{#2}
\providecommand\natexlab[1]{#1}
\providecommand\showeprint[2][]{arXiv:#2}

\bibitem[\protect\citeauthoryear{??}{ima}{2010}]%
        {imagenet-crowdsource}
 \bibinfo{year}{2010}\natexlab{}.
\newblock \bibinfo{title}{ImageNet crowdsourcing, benchmarking \& other cool
  things}.
\newblock
  \bibinfo{howpublished}{\url{http://www.image-net.org/papers/ImageNet_2010.pdf}}.
    (\bibinfo{year}{2010}).
\newblock


\bibitem[\protect\citeauthoryear{??}{way}{2016a}]%
        {waymoreport}
 \bibinfo{year}{2016}\natexlab{a}.
\newblock \bibinfo{title}{Google auto Waymo disengagement report for autonomous
  driving}.
\newblock
  \bibinfo{howpublished}{\url{https://www.dmv.ca.gov/portal/wcm/connect/946b3502-c959-4e3b-b119-91319c27788f/GoogleAutoWaymo_disengage_report_2016.pdf?MOD=AJPERES}}.
    (\bibinfo{year}{2016}).
\newblock


\bibitem[\protect\citeauthoryear{??}{way}{2016b}]%
        {waymo-report}
 \bibinfo{year}{2016}\natexlab{b}.
\newblock \bibinfo{title}{Report on autonomous mode disengagements for waymo
  self-driving vehicles in california}.
\newblock
  \bibinfo{howpublished}{\url{https://www.dmv.ca.gov/portal/wcm/connect/946b3502-c959-4e3b-b119-91319c27788f/GoogleAutoWaymo_disengage_report_2016.pdf?MOD=AJPERES}}.
    (\bibinfo{year}{2016}).
\newblock


\bibitem[\protect\citeauthoryear{??}{way}{2017}]%
        {waymo_simulation}
 \bibinfo{year}{2017}\natexlab{}.
\newblock \bibinfo{title}{Inside Waymo's secret world for training self-driving
  cars}.
\newblock
  \bibinfo{howpublished}{\url{https://www.theatlantic.com/technology/archive/2017/08/inside-waymos-secret-testing-and-simulation-facilities/537648/}}.
    (\bibinfo{year}{2017}).
\newblock


\bibitem[\protect\citeauthoryear{Abadi, Barham, Chen, Chen, Davis, Dean, Devin,
  Ghemawat, Irving, Isard, et~al\mbox{.}}{Abadi et~al\mbox{.}}{2016}]%
        {abadi2016tensorflow}
\bibfield{author}{\bibinfo{person}{Mart{\'\i}n Abadi}, \bibinfo{person}{Paul
  Barham}, \bibinfo{person}{Jianmin Chen}, \bibinfo{person}{Zhifeng Chen},
  \bibinfo{person}{Andy Davis}, \bibinfo{person}{Jeffrey Dean},
  \bibinfo{person}{Matthieu Devin}, \bibinfo{person}{Sanjay Ghemawat},
  \bibinfo{person}{Geoffrey Irving}, \bibinfo{person}{Michael Isard},
  {et~al\mbox{.}}} \bibinfo{year}{2016}\natexlab{}.
\newblock \showarticletitle{TensorFlow: A system for large-scale machine
  learning}. In \bibinfo{booktitle}{{\em Proceedings of the 12th USENIX
  Symposium on Operating Systems Design and Implementation}}.
\newblock


\bibitem[\protect\citeauthoryear{Argyros, Stais, Jana, Keromytis, and
  Kiayias}{Argyros et~al\mbox{.}}{2016}]%
        {sfadiff}
\bibfield{author}{\bibinfo{person}{George Argyros}, \bibinfo{person}{Ioannis
  Stais}, \bibinfo{person}{Suman Jana}, \bibinfo{person}{Angelos~D Keromytis},
  {and} \bibinfo{person}{Aggelos Kiayias}.} \bibinfo{year}{2016}\natexlab{}.
\newblock \showarticletitle{SFADiff: Automated evasion attacks and
  fingerprinting using black-box differential automata learning}. In
  \bibinfo{booktitle}{{\em Proceedings of the 23rd ACM SIGSAC Conference on
  Computer and Communications Security}}.
\newblock


\bibitem[\protect\citeauthoryear{Arp, Spreitzenbarth, Hubner, Gascon, Rieck,
  and Siemens}{Arp et~al\mbox{.}}{2014}]%
        {arp2014drebin}
\bibfield{author}{\bibinfo{person}{Daniel Arp}, \bibinfo{person}{Michael
  Spreitzenbarth}, \bibinfo{person}{Malte Hubner}, \bibinfo{person}{Hugo
  Gascon}, \bibinfo{person}{Konrad Rieck}, {and} \bibinfo{person}{CERT
  Siemens}.} \bibinfo{year}{2014}\natexlab{}.
\newblock \showarticletitle{DREBIN: Effective and Explainable Detection of
  Android Malware in Your Pocket.}. In \bibinfo{booktitle}{{\em Proceedings of
  the 21st Annual Network and Distributed System Security Symposium}}.
\newblock


\bibitem[\protect\citeauthoryear{autopilot:dave}{autopilot:dave}{2016}]%
        {autopilot:dave}
autopilot:dave \bibinfo{year}{2016}\natexlab{}.
\newblock \bibinfo{title}{Nvidia-Autopilot-Keras}.
\newblock
  \bibinfo{howpublished}{\url{https://github.com/0bserver07/Nvidia-Autopilot-Keras}}.
    (\bibinfo{year}{2016}).
\newblock


\bibitem[\protect\citeauthoryear{Bastani, Ioannou, Lampropoulos, Vytiniotis,
  Nori, and Criminisi}{Bastani et~al\mbox{.}}{2016}]%
        {bastani2016measuring}
\bibfield{author}{\bibinfo{person}{Osbert Bastani}, \bibinfo{person}{Yani
  Ioannou}, \bibinfo{person}{Leonidas Lampropoulos}, \bibinfo{person}{Dimitrios
  Vytiniotis}, \bibinfo{person}{Aditya Nori}, {and} \bibinfo{person}{Antonio
  Criminisi}.} \bibinfo{year}{2016}\natexlab{}.
\newblock \showarticletitle{Measuring neural net robustness with constraints}.
  In \bibinfo{booktitle}{{\em Proceedings of the 29th Advances in Neural
  Information Processing Systems}}.
\newblock


\bibitem[\protect\citeauthoryear{Bojarski, Del~Testa, Dworakowski, Firner,
  Flepp, Goyal, Jackel, Monfort, Muller, Zhang, et~al\mbox{.}}{Bojarski
  et~al\mbox{.}}{2016}]%
        {bojarski2016end}
\bibfield{author}{\bibinfo{person}{Mariusz Bojarski}, \bibinfo{person}{Davide
  Del~Testa}, \bibinfo{person}{Daniel Dworakowski}, \bibinfo{person}{Bernhard
  Firner}, \bibinfo{person}{Beat Flepp}, \bibinfo{person}{Prasoon Goyal},
  \bibinfo{person}{Lawrence~D Jackel}, \bibinfo{person}{Mathew Monfort},
  \bibinfo{person}{Urs Muller}, \bibinfo{person}{Jiakai Zhang},
  {et~al\mbox{.}}} \bibinfo{year}{2016}\natexlab{}.
\newblock \showarticletitle{End to end learning for self-driving cars}.
\newblock \bibinfo{journal}{{\em arXiv preprint arXiv:1604.07316\/}}
  (\bibinfo{year}{2016}).
\newblock


\bibitem[\protect\citeauthoryear{Brubaker, Jana, Ray, Khurshid, and
  Shmatikov}{Brubaker et~al\mbox{.}}{2014}]%
        {frankencert}
\bibfield{author}{\bibinfo{person}{Chad Brubaker}, \bibinfo{person}{Suman
  Jana}, \bibinfo{person}{Baishakhi Ray}, \bibinfo{person}{Sarfraz Khurshid},
  {and} \bibinfo{person}{Vitaly Shmatikov}.} \bibinfo{year}{2014}\natexlab{}.
\newblock \showarticletitle{Using Frankencerts for Automated Adversarial
  Testing of Certificate Validation in {SSL/TLS} Implementations}. In
  \bibinfo{booktitle}{{\em Proceedings of the 35th IEEE Symposium on Security
  and Privacy}}.
\newblock


\bibitem[\protect\citeauthoryear{Cao and Yang}{Cao and Yang}{2015}]%
        {cao2015towards}
\bibfield{author}{\bibinfo{person}{Yinzhi Cao} {and} \bibinfo{person}{Junfeng
  Yang}.} \bibinfo{year}{2015}\natexlab{}.
\newblock \showarticletitle{Towards Making Systems Forget with Machine
  Unlearning}. In \bibinfo{booktitle}{{\em Proceedings of the 36th IEEE
  Symposium on Security and Privacy}}.
\newblock


\bibitem[\protect\citeauthoryear{Carlini and Wagner}{Carlini and
  Wagner}{2017}]%
        {carlini2017towards}
\bibfield{author}{\bibinfo{person}{Nicholas Carlini} {and}
  \bibinfo{person}{David Wagner}.} \bibinfo{year}{2017}\natexlab{}.
\newblock \showarticletitle{Towards evaluating the robustness of neural
  networks}. In \bibinfo{booktitle}{{\em Proceedings of the 38th IEEE Symposium
  on Security and Privacy}}.
\newblock


\bibitem[\protect\citeauthoryear{Chen, Su, Sun, Su, and Zhao}{Chen
  et~al\mbox{.}}{2016}]%
        {chen_2016_pldi_difftest_jvm}
\bibfield{author}{\bibinfo{person}{Yuting Chen}, \bibinfo{person}{Ting Su},
  \bibinfo{person}{Chengnian Sun}, \bibinfo{person}{Zhendong Su}, {and}
  \bibinfo{person}{Jianjun Zhao}.} \bibinfo{year}{2016}\natexlab{}.
\newblock \showarticletitle{Coverage-directed differential testing of {JVM}
  implementations}. In \bibinfo{booktitle}{{\em Proceedings of the 37th ACM
  SIGPLAN Conference on Programming Language Design and Implementation}}.
\newblock


\bibitem[\protect\citeauthoryear{Chen and Su}{Chen and Su}{2015}]%
        {mucert}
\bibfield{author}{\bibinfo{person}{Yuting Chen} {and} \bibinfo{person}{Zhendong
  Su}.} \bibinfo{year}{2015}\natexlab{}.
\newblock \showarticletitle{Guided differential testing of certificate
  validation in SSL/TLS implementations}. In \bibinfo{booktitle}{{\em
  Proceedings of the 10th Joint Meeting on Foundations of Software
  Engineering}}.
\newblock


\bibitem[\protect\citeauthoryear{Chollet}{Chollet}{2015}]%
        {chollet2015keras}
\bibfield{author}{\bibinfo{person}{Fran{\c{c}}ois Chollet}.}
  \bibinfo{year}{2015}\natexlab{}.
\newblock \bibinfo{title}{Keras}.
\newblock \bibinfo{howpublished}{\url{https://github.com/fchollet/keras}}.
  (\bibinfo{year}{2015}).
\newblock


\bibitem[\protect\citeauthoryear{Cisse, Bojanowski, Grave, Dauphin, and
  Usunier}{Cisse et~al\mbox{.}}{2017}]%
        {cisse2017parseval}
\bibfield{author}{\bibinfo{person}{Moustapha Cisse}, \bibinfo{person}{Piotr
  Bojanowski}, \bibinfo{person}{Edouard Grave}, \bibinfo{person}{Yann Dauphin},
  {and} \bibinfo{person}{Nicolas Usunier}.} \bibinfo{year}{2017}\natexlab{}.
\newblock \showarticletitle{Parseval networks: Improving robustness to
  adversarial examples}. In \bibinfo{booktitle}{{\em Proceedings of the 34th
  International Conference on Machine Learning}}.
\newblock


\bibitem[\protect\citeauthoryear{clone:dave}{clone:dave}{2016}]%
        {clone:dave}
clone:dave \bibinfo{year}{2016}\natexlab{}.
\newblock \bibinfo{title}{Behavioral cloning: end-to-end learning for
  self-driving cars}.
\newblock
  \bibinfo{howpublished}{\url{https://github.com/navoshta/behavioral-cloning}}.
    (\bibinfo{year}{2016}).
\newblock


\bibitem[\protect\citeauthoryear{contagio}{contagio}{2010}]%
        {contagio}
contagio \bibinfo{year}{2010}\natexlab{}.
\newblock \bibinfo{title}{Contagio, PDF malware dump}.
\newblock \bibinfo{howpublished}{\url{http://
  contagiodump.blogspot.de/2010/08/malicious-documents-archive-for.html}}.
  (\bibinfo{year}{2010}).
\newblock


\bibitem[\protect\citeauthoryear{Deng, Dong, Socher, Li, Li, and Fei-Fei}{Deng
  et~al\mbox{.}}{2009}]%
        {deng2009imagenet}
\bibfield{author}{\bibinfo{person}{Jia Deng}, \bibinfo{person}{Wei Dong},
  \bibinfo{person}{Richard Socher}, \bibinfo{person}{Li-Jia Li},
  \bibinfo{person}{Kai Li}, {and} \bibinfo{person}{Li Fei-Fei}.}
  \bibinfo{year}{2009}\natexlab{}.
\newblock \showarticletitle{Imagenet: A large-scale hierarchical image
  database}. In \bibinfo{booktitle}{{\em Proceedings of the 22nd IEEE
  Conference on Computer Vision and Pattern Recognition}}.
\newblock


\bibitem[\protect\citeauthoryear{Fredrikson, Jha, and Ristenpart}{Fredrikson
  et~al\mbox{.}}{2015}]%
        {fredrikson2015model}
\bibfield{author}{\bibinfo{person}{Matt Fredrikson}, \bibinfo{person}{Somesh
  Jha}, {and} \bibinfo{person}{Thomas Ristenpart}.}
  \bibinfo{year}{2015}\natexlab{}.
\newblock \showarticletitle{Model inversion attacks that exploit confidence
  information and basic countermeasures}. In \bibinfo{booktitle}{{\em
  Proceedings of the 22nd ACM SIGSAC Conference on Computer and Communications
  Security}}.
\newblock


\bibitem[\protect\citeauthoryear{Fredrikson, Lantz, Jha, Lin, Page, and
  Ristenpart}{Fredrikson et~al\mbox{.}}{2014}]%
        {fredrikson2014privacy}
\bibfield{author}{\bibinfo{person}{Matthew Fredrikson}, \bibinfo{person}{Eric
  Lantz}, \bibinfo{person}{Somesh Jha}, \bibinfo{person}{Simon Lin},
  \bibinfo{person}{David Page}, {and} \bibinfo{person}{Thomas Ristenpart}.}
  \bibinfo{year}{2014}\natexlab{}.
\newblock \showarticletitle{Privacy in pharmacogenetics: An end-to-end case
  study of personalized warfarin dosing.}. In \bibinfo{booktitle}{{\em
  Proceedings of the 23rd {USENIX} Security Symposium ({USENIX} Security 14)}}.
\newblock


\bibitem[\protect\citeauthoryear{Freund and Schapire}{Freund and
  Schapire}{1995}]%
        {freund1995desicion}
\bibfield{author}{\bibinfo{person}{Yoav Freund} {and} \bibinfo{person}{Robert~E
  Schapire}.} \bibinfo{year}{1995}\natexlab{}.
\newblock \showarticletitle{A desicion-theoretic generalization of on-line
  learning and an application to boosting}. In \bibinfo{booktitle}{{\em
  European conference on computational learning theory}}.
\newblock


\bibitem[\protect\citeauthoryear{Gatys, Ecker, and Bethge}{Gatys
  et~al\mbox{.}}{2015}]%
        {gatys2015neural}
\bibfield{author}{\bibinfo{person}{Leon~A Gatys}, \bibinfo{person}{Alexander~S
  Ecker}, {and} \bibinfo{person}{Matthias Bethge}.}
  \bibinfo{year}{2015}\natexlab{}.
\newblock \showarticletitle{A neural algorithm of artistic style}.
\newblock \bibinfo{journal}{{\em arXiv preprint arXiv:1508.06576\/}}
  (\bibinfo{year}{2015}).
\newblock


\bibitem[\protect\citeauthoryear{Goodfellow and Papernot}{Goodfellow and
  Papernot}{2017}]%
        {verify-test}
\bibfield{author}{\bibinfo{person}{Ian Goodfellow} {and}
  \bibinfo{person}{Nicolas Papernot}.} \bibinfo{year}{2017}\natexlab{}.
\newblock \bibinfo{title}{The challenge of verification and testing of machine
  learning}.
\newblock
  \bibinfo{howpublished}{\url{http://www.cleverhans.io/security/privacy/ml/2017/06/14/verification.html}}.
    (\bibinfo{year}{2017}).
\newblock


\bibitem[\protect\citeauthoryear{Goodfellow, Shlens, and Szegedy}{Goodfellow
  et~al\mbox{.}}{2015}]%
        {goodfellow2014explaining}
\bibfield{author}{\bibinfo{person}{Ian Goodfellow}, \bibinfo{person}{Jonathon
  Shlens}, {and} \bibinfo{person}{Christian Szegedy}.}
  \bibinfo{year}{2015}\natexlab{}.
\newblock \showarticletitle{Explaining and Harnessing Adversarial Examples}. In
  \bibinfo{booktitle}{{\em Proceedings of the 3rd International Conference on
  Learning Representations}}.
\newblock
\showURL{%
\url{http://arxiv.org/abs/1412.6572}}


\bibitem[\protect\citeauthoryear{google-accident}{google-accident}{2016}]%
        {google-accident}
google-accident \bibinfo{year}{2016}\natexlab{}.
\newblock \bibinfo{title}{A Google self-driving car caused a crash for the
  first time}.
\newblock
  \bibinfo{howpublished}{\url{http://www.theverge.com/2016/2/29/11134344/google-self-driving-car-crash-report}}.
    (\bibinfo{year}{2016}).
\newblock


\bibitem[\protect\citeauthoryear{Groce, Holzmann, and Joshi}{Groce
  et~al\mbox{.}}{2007}]%
        {groce2007randomized}
\bibfield{author}{\bibinfo{person}{Alex Groce}, \bibinfo{person}{Gerard
  Holzmann}, {and} \bibinfo{person}{Rajeev Joshi}.}
  \bibinfo{year}{2007}\natexlab{}.
\newblock \showarticletitle{Randomized differential testing as a prelude to
  formal verification}. In \bibinfo{booktitle}{{\em Proceedings of the 29th
  international conference on Software Engineering}}.
\newblock


\bibitem[\protect\citeauthoryear{Grosse, Papernot, Manoharan, Backes, and
  McDaniel}{Grosse et~al\mbox{.}}{2016}]%
        {grosse2016adversarial}
\bibfield{author}{\bibinfo{person}{Kathrin Grosse}, \bibinfo{person}{Nicolas
  Papernot}, \bibinfo{person}{Praveen Manoharan}, \bibinfo{person}{Michael
  Backes}, {and} \bibinfo{person}{Patrick McDaniel}.}
  \bibinfo{year}{2016}\natexlab{}.
\newblock \showarticletitle{Adversarial perturbations against deep neural
  networks for malware classification}.
\newblock \bibinfo{journal}{{\em arXiv preprint arXiv:1606.04435\/}}
  (\bibinfo{year}{2016}).
\newblock


\bibitem[\protect\citeauthoryear{Gu and Rigazio}{Gu and Rigazio}{2015}]%
        {gu2014towards}
\bibfield{author}{\bibinfo{person}{Shixiang Gu} {and} \bibinfo{person}{Luca
  Rigazio}.} \bibinfo{year}{2015}\natexlab{}.
\newblock \showarticletitle{Towards deep neural network architectures robust to
  adversarial examples}. In \bibinfo{booktitle}{{\em Proceedings of the 3rd
  International Conference on Learning Representations}}.
\newblock


\bibitem[\protect\citeauthoryear{He, Zhang, Ren, and Sun}{He
  et~al\mbox{.}}{2016}]%
        {he2015deep}
\bibfield{author}{\bibinfo{person}{Kaiming He}, \bibinfo{person}{Xiangyu
  Zhang}, \bibinfo{person}{Shaoqing Ren}, {and} \bibinfo{person}{Jian Sun}.}
  \bibinfo{year}{2016}\natexlab{}.
\newblock \showarticletitle{Deep residual learning for image recognition}. In
  \bibinfo{booktitle}{{\em Proceedings of the 29th IEEE Conference on Computer
  Vision and Pattern Recognition}}. \bibinfo{pages}{770--778}.
\newblock


\bibitem[\protect\citeauthoryear{Huang, Kwiatkowska, Wang, and Wu}{Huang
  et~al\mbox{.}}{2017}]%
        {huang2017safety}
\bibfield{author}{\bibinfo{person}{Xiaowei Huang}, \bibinfo{person}{Marta
  Kwiatkowska}, \bibinfo{person}{Sen Wang}, {and} \bibinfo{person}{Min Wu}.}
  \bibinfo{year}{2017}\natexlab{}.
\newblock \showarticletitle{Safety verification of deep neural networks}. In
  \bibinfo{booktitle}{{\em Proceedings of the 29th International Conference on
  Computer Aided Verification}}.
\newblock


\bibitem[\protect\citeauthoryear{Ioffe and Szegedy}{Ioffe and Szegedy}{2015}]%
        {ioffe2015batch}
\bibfield{author}{\bibinfo{person}{Sergey Ioffe} {and}
  \bibinfo{person}{Christian Szegedy}.} \bibinfo{year}{2015}\natexlab{}.
\newblock \showarticletitle{Batch normalization: Accelerating deep network
  training by reducing internal covariate shift}.
\newblock \bibinfo{journal}{{\em arXiv preprint arXiv:1502.03167\/}}
  (\bibinfo{year}{2015}).
\newblock


\bibitem[\protect\citeauthoryear{Jouppi, Young, Patil, Patterson, Agrawal,
  Bajwa, Bates, Bhatia, Boden, Borchers, Boyle, et~al\mbox{.}}{Jouppi
  et~al\mbox{.}}{2017}]%
        {tpu}
\bibfield{author}{\bibinfo{person}{Norman~P. Jouppi}, \bibinfo{person}{Cliff
  Young}, \bibinfo{person}{Nishant Patil}, \bibinfo{person}{David Patterson},
  \bibinfo{person}{Gaurav Agrawal}, \bibinfo{person}{Raminder Bajwa},
  \bibinfo{person}{Sarah Bates}, \bibinfo{person}{Suresh Bhatia},
  \bibinfo{person}{Nan Boden}, \bibinfo{person}{Al Borchers},
  \bibinfo{person}{Rick Boyle}, {et~al\mbox{.}}}
  \bibinfo{year}{2017}\natexlab{}.
\newblock \showarticletitle{In-Datacenter Performance Analysis of a Tensor
  Processing Unit}. In \bibinfo{booktitle}{{\em Proceedings of the 44th Annual
  International Symposium on Computer Architecture}}.
\newblock


\bibitem[\protect\citeauthoryear{Julian, Lopez, Brush, Owen, and
  Kochenderfer}{Julian et~al\mbox{.}}{2016}]%
        {julian2016policy}
\bibfield{author}{\bibinfo{person}{Kyle~D Julian}, \bibinfo{person}{Jessica
  Lopez}, \bibinfo{person}{Jeffrey~S Brush}, \bibinfo{person}{Michael~P Owen},
  {and} \bibinfo{person}{Mykel~J Kochenderfer}.}
  \bibinfo{year}{2016}\natexlab{}.
\newblock \showarticletitle{Policy compression for aircraft collision avoidance
  systems}. In \bibinfo{booktitle}{{\em Proceedings of the 35th IEEE/AIAA
  Digital Avionics Systems Conference}}.
\newblock


\bibitem[\protect\citeauthoryear{Jung, Sheth, Greenstein, Wetherall, Maganis,
  and Kohno}{Jung et~al\mbox{.}}{2008}]%
        {jung2008privacy}
\bibfield{author}{\bibinfo{person}{Jaeyeon Jung}, \bibinfo{person}{Anmol
  Sheth}, \bibinfo{person}{Ben Greenstein}, \bibinfo{person}{David Wetherall},
  \bibinfo{person}{Gabriel Maganis}, {and} \bibinfo{person}{Tadayoshi Kohno}.}
  \bibinfo{year}{2008}\natexlab{}.
\newblock \showarticletitle{Privacy oracle: a system for finding application
  leaks with black box differential testing}. In \bibinfo{booktitle}{{\em
  Proceedings of the 15th ACM Conference on Computer and Communications
  Security}}.
\newblock


\bibitem[\protect\citeauthoryear{Katz, Barrett, Dill, Julian, and
  Kochenderfer}{Katz et~al\mbox{.}}{2017}]%
        {katz2017reluplex}
\bibfield{author}{\bibinfo{person}{Guy Katz}, \bibinfo{person}{Clark Barrett},
  \bibinfo{person}{David~L. Dill}, \bibinfo{person}{Kyle Julian}, {and}
  \bibinfo{person}{Mykel~J. Kochenderfer}.} \bibinfo{year}{2017}\natexlab{}.
\newblock \showarticletitle{Reluplex: An Efficient SMT Solver for Verifying
  Deep Neural Networks}. In \bibinfo{booktitle}{{\em Proceedings of the 29th
  International Conference On Computer Aided Verification}}.
\newblock


\bibitem[\protect\citeauthoryear{Krizhevsky}{Krizhevsky}{2009}]%
        {krizhevsky2009learning}
\bibfield{author}{\bibinfo{person}{Alex Krizhevsky}.}
  \bibinfo{year}{2009}\natexlab{}.
\newblock \bibinfo{booktitle}{{\em Learning multiple layers of features from
  tiny images}}.
\newblock \bibinfo{type}{{T}echnical {R}eport}.
\newblock


\bibitem[\protect\citeauthoryear{Krizhevsky, Sutskever, and Hinton}{Krizhevsky
  et~al\mbox{.}}{2012}]%
        {krizhevsky2012imagenet}
\bibfield{author}{\bibinfo{person}{Alex Krizhevsky}, \bibinfo{person}{Ilya
  Sutskever}, {and} \bibinfo{person}{Geoffrey~E. Hinton}.}
  \bibinfo{year}{2012}\natexlab{}.
\newblock \showarticletitle{ImageNet Classification with Deep Convolutional
  Neural Networks}. In \bibinfo{booktitle}{{\em Proceedings of the 25th
  International Conference on Neural Information Processing Systems}}.
\newblock


\bibitem[\protect\citeauthoryear{LeCun, Bottou, Bengio, and Haffner}{LeCun
  et~al\mbox{.}}{1998a}]%
        {lecun1998gradient}
\bibfield{author}{\bibinfo{person}{Yann LeCun}, \bibinfo{person}{L{\'e}on
  Bottou}, \bibinfo{person}{Yoshua Bengio}, {and} \bibinfo{person}{Patrick
  Haffner}.} \bibinfo{year}{1998}\natexlab{a}.
\newblock \showarticletitle{Gradient-based learning applied to document
  recognition}.
\newblock \bibinfo{journal}{{\it Proc. IEEE}} (\bibinfo{year}{1998}).
\newblock


\bibitem[\protect\citeauthoryear{LeCun, Cortes, and Burges}{LeCun
  et~al\mbox{.}}{1998b}]%
        {lecun1998mnist}
\bibfield{author}{\bibinfo{person}{Yann LeCun}, \bibinfo{person}{Corinna
  Cortes}, {and} \bibinfo{person}{Christopher~JC Burges}.}
  \bibinfo{year}{1998}\natexlab{b}.
\newblock \bibinfo{title}{The MNIST database of handwritten digits}.
\newblock   (\bibinfo{year}{1998}).
\newblock


\bibitem[\protect\citeauthoryear{LeCun, Cortes, and Burges}{LeCun
  et~al\mbox{.}}{2010}]%
        {lecun2010mnist}
\bibfield{author}{\bibinfo{person}{Yann LeCun}, \bibinfo{person}{Corinna
  Cortes}, {and} \bibinfo{person}{Christopher~JC Burges}.}
  \bibinfo{year}{2010}\natexlab{}.
\newblock \showarticletitle{MNIST handwritten digit database}.
\newblock \bibinfo{journal}{{\em AT\&T Labs [Online]. Available:
  http://yann.lecun.com/exdb/mnist\/}}  \bibinfo{volume}{2}
  (\bibinfo{year}{2010}).
\newblock


\bibitem[\protect\citeauthoryear{Li and Wand}{Li and Wand}{2016}]%
        {li2016combining}
\bibfield{author}{\bibinfo{person}{Chuan Li} {and} \bibinfo{person}{Michael
  Wand}.} \bibinfo{year}{2016}\natexlab{}.
\newblock \showarticletitle{Combining markov random fields and convolutional
  neural networks for image synthesis}. In \bibinfo{booktitle}{{\em Proceedings
  of the 29th IEEE Conference on Computer Vision and Pattern Recognition}}.
\newblock


\bibitem[\protect\citeauthoryear{Mahendran and Vedaldi}{Mahendran and
  Vedaldi}{2015}]%
        {mahendran2015understanding}
\bibfield{author}{\bibinfo{person}{Aravindh Mahendran} {and}
  \bibinfo{person}{Andrea Vedaldi}.} \bibinfo{year}{2015}\natexlab{}.
\newblock \showarticletitle{Understanding deep image representations by
  inverting them}. In \bibinfo{booktitle}{{\em Proceedings of the 28th IEEE
  Conference on Computer Vision and Pattern Recognition}}.
\newblock


\bibitem[\protect\citeauthoryear{McKeeman}{McKeeman}{1998}]%
        {mckeeman1998differential}
\bibfield{author}{\bibinfo{person}{William~M McKeeman}.}
  \bibinfo{year}{1998}\natexlab{}.
\newblock \showarticletitle{Differential testing for software}.
\newblock \bibinfo{journal}{{\em Digital Technical Journal\/}}
  (\bibinfo{year}{1998}).
\newblock


\bibitem[\protect\citeauthoryear{Metzen, Genewein, Fischer, and
  Bischoff}{Metzen et~al\mbox{.}}{2017}]%
        {metzen2017detecting}
\bibfield{author}{\bibinfo{person}{Jan~Hendrik Metzen}, \bibinfo{person}{Tim
  Genewein}, \bibinfo{person}{Volker Fischer}, {and} \bibinfo{person}{Bastian
  Bischoff}.} \bibinfo{year}{2017}\natexlab{}.
\newblock \showarticletitle{On detecting adversarial perturbations}. In
  \bibinfo{booktitle}{{\em Proceedings of the 6th International Conference on
  Learning Representations}}.
\newblock


\bibitem[\protect\citeauthoryear{Miller}{Miller}{1995}]%
        {miller1995wordnet}
\bibfield{author}{\bibinfo{person}{George~A Miller}.}
  \bibinfo{year}{1995}\natexlab{}.
\newblock \showarticletitle{WordNet: a lexical database for English}.
\newblock \bibinfo{journal}{{\it Commun. ACM}} (\bibinfo{year}{1995}).
\newblock


\bibitem[\protect\citeauthoryear{Nair and Hinton}{Nair and Hinton}{2010}]%
        {nair2010rectified}
\bibfield{author}{\bibinfo{person}{Vinod Nair} {and}
  \bibinfo{person}{Geoffrey~E Hinton}.} \bibinfo{year}{2010}\natexlab{}.
\newblock \showarticletitle{Rectified linear units improve restricted boltzmann
  machines}. In \bibinfo{booktitle}{{\em Proceedings of the 27th International
  Conference on Machine Learning}}. \bibinfo{pages}{807--814}.
\newblock


\bibitem[\protect\citeauthoryear{Nguyen, Yosinski, and Clune}{Nguyen
  et~al\mbox{.}}{2015}]%
        {nguyen2015deep}
\bibfield{author}{\bibinfo{person}{Anh Nguyen}, \bibinfo{person}{Jason
  Yosinski}, {and} \bibinfo{person}{Jeff Clune}.}
  \bibinfo{year}{2015}\natexlab{}.
\newblock \showarticletitle{Deep neural networks are easily fooled: High
  confidence predictions for unrecognizable images}. In
  \bibinfo{booktitle}{{\em Proceedings of the 28th IEEE Conference on Computer
  Vision and Pattern Recognition}}.
\newblock


\bibitem[\protect\citeauthoryear{Nvidia}{Nvidia}{2008}]%
        {nvidia2008programming}
\bibfield{author}{\bibinfo{person}{Nvidia}.} \bibinfo{year}{2008}\natexlab{}.
\newblock \bibinfo{title}{CUDA Programming guide}.
\newblock   (\bibinfo{year}{2008}).
\newblock


\bibitem[\protect\citeauthoryear{Papernot and McDaniel}{Papernot and
  McDaniel}{2017}]%
        {papernot2017extending}
\bibfield{author}{\bibinfo{person}{Nicolas Papernot} {and}
  \bibinfo{person}{Patrick McDaniel}.} \bibinfo{year}{2017}\natexlab{}.
\newblock \showarticletitle{Extending defensive distillation}.
\newblock \bibinfo{journal}{{\em arXiv preprint arXiv:1705.05264\/}}
  (\bibinfo{year}{2017}).
\newblock


\bibitem[\protect\citeauthoryear{Papernot, McDaniel, Jha, Fredrikson, Celik,
  and Swami}{Papernot et~al\mbox{.}}{2016a}]%
        {papernot2016limitations}
\bibfield{author}{\bibinfo{person}{Nicolas Papernot}, \bibinfo{person}{Patrick
  McDaniel}, \bibinfo{person}{Somesh Jha}, \bibinfo{person}{Matt Fredrikson},
  \bibinfo{person}{Z~Berkay Celik}, {and} \bibinfo{person}{Ananthram Swami}.}
  \bibinfo{year}{2016}\natexlab{a}.
\newblock \showarticletitle{The limitations of deep learning in adversarial
  settings}. In \bibinfo{booktitle}{{\em Proceedings of the 37th IEEE European
  Symposium on Security and Privacy}}.
\newblock


\bibitem[\protect\citeauthoryear{Papernot, McDaniel, Wu, Jha, and
  Swami}{Papernot et~al\mbox{.}}{2016b}]%
        {papernot2016distillation}
\bibfield{author}{\bibinfo{person}{Nicolas Papernot}, \bibinfo{person}{Patrick
  McDaniel}, \bibinfo{person}{Xi Wu}, \bibinfo{person}{Somesh Jha}, {and}
  \bibinfo{person}{Ananthram Swami}.} \bibinfo{year}{2016}\natexlab{b}.
\newblock \showarticletitle{Distillation as a defense to adversarial
  perturbations against deep neural networks}. In \bibinfo{booktitle}{{\em
  Proceedings of the 37th IEEE Symposium on Security and Privacy}}.
\newblock


\bibitem[\protect\citeauthoryear{pdfrate}{pdfrate}{2012}]%
        {pdfrate}
pdfrate \bibinfo{year}{2012}\natexlab{}.
\newblock \bibinfo{title}{PDFRate, A machine learning based classifier
  operating on document metadata and structure}.
\newblock \bibinfo{howpublished}{\url{http://pdfrate.com/}}.
  (\bibinfo{year}{2012}).
\newblock


\bibitem[\protect\citeauthoryear{Perdisci, Dagon, Lee, Fogla, and
  Sharif}{Perdisci et~al\mbox{.}}{2006}]%
        {perdisci2006misleading}
\bibfield{author}{\bibinfo{person}{Roberto Perdisci}, \bibinfo{person}{David
  Dagon}, \bibinfo{person}{Wenke Lee}, \bibinfo{person}{P. Fogla}, {and}
  \bibinfo{person}{M. Sharif}.} \bibinfo{year}{2006}\natexlab{}.
\newblock \showarticletitle{Misleading worm signature generators using
  deliberate noise injection}. In \bibinfo{booktitle}{{\em Proceedings of the
  27th IEEE Symposium on Security and Privacy}}.
\newblock


\bibitem[\protect\citeauthoryear{Petsios, Tang, Stolfo, Keromytis, and
  Jana}{Petsios et~al\mbox{.}}{2017}]%
        {jana:nezha}
\bibfield{author}{\bibinfo{person}{Theofilos Petsios}, \bibinfo{person}{Adrian
  Tang}, \bibinfo{person}{Salvatore~J. Stolfo}, \bibinfo{person}{Angelos~D.
  Keromytis}, {and} \bibinfo{person}{Suman Jana}.}
  \bibinfo{year}{2017}\natexlab{}.
\newblock \showarticletitle{{NEZHA: Efficient Domain-independent Differential
  Testing}}. In \bibinfo{booktitle}{{\em Proceedings of the 38th IEEE Symposium
  on Security and Privacy}}.
\newblock


\bibitem[\protect\citeauthoryear{Pulina and Tacchella}{Pulina and
  Tacchella}{2010}]%
        {pulina2010abstraction}
\bibfield{author}{\bibinfo{person}{Luca Pulina} {and} \bibinfo{person}{Armando
  Tacchella}.} \bibinfo{year}{2010}\natexlab{}.
\newblock \showarticletitle{An abstraction-refinement approach to verification
  of artificial neural networks}. In \bibinfo{booktitle}{{\em Proceedings of
  the 22nd International Conference on Computer Aided Verification}}.
\newblock


\bibitem[\protect\citeauthoryear{Radford, Jozefowicz, and Sutskever}{Radford
  et~al\mbox{.}}{2017}]%
        {radford2017learning}
\bibfield{author}{\bibinfo{person}{Alec Radford}, \bibinfo{person}{Rafal
  Jozefowicz}, {and} \bibinfo{person}{Ilya Sutskever}.}
  \bibinfo{year}{2017}\natexlab{}.
\newblock \showarticletitle{Learning to generate reviews and discovering
  sentiment}.
\newblock \bibinfo{journal}{{\em arXiv preprint arXiv:1704.01444\/}}
  (\bibinfo{year}{2017}).
\newblock


\bibitem[\protect\citeauthoryear{Ruder, Dosovitskiy, and Brox}{Ruder
  et~al\mbox{.}}{2016}]%
        {ruder2016artistic}
\bibfield{author}{\bibinfo{person}{Manuel Ruder}, \bibinfo{person}{Alexey
  Dosovitskiy}, {and} \bibinfo{person}{Thomas Brox}.}
  \bibinfo{year}{2016}\natexlab{}.
\newblock \showarticletitle{Artistic style transfer for videos}. In
  \bibinfo{booktitle}{{\em German Conference on Pattern Recognition}}.
\newblock


\bibitem[\protect\citeauthoryear{Rumelhart, Hinton, and Williams}{Rumelhart
  et~al\mbox{.}}{1988}]%
        {rumelhart1988learning}
\bibfield{author}{\bibinfo{person}{David~E Rumelhart},
  \bibinfo{person}{Geoffrey~E Hinton}, {and} \bibinfo{person}{Ronald~J
  Williams}.} \bibinfo{year}{1988}\natexlab{}.
\newblock \showarticletitle{Learning representations by back-propagating
  errors}.
\newblock \bibinfo{journal}{{\em Cognitive modeling\/}} (\bibinfo{year}{1988}).
\newblock


\bibitem[\protect\citeauthoryear{Russakovsky, Deng, Su, Krause, Satheesh, Ma,
  Huang, Karpathy, Khosla, Bernstein, Berg, and Fei-Fei}{Russakovsky
  et~al\mbox{.}}{2015}]%
        {ILSVRC15}
\bibfield{author}{\bibinfo{person}{Olga Russakovsky}, \bibinfo{person}{Jia
  Deng}, \bibinfo{person}{Hao Su}, \bibinfo{person}{Jonathan Krause},
  \bibinfo{person}{Sanjeev Satheesh}, \bibinfo{person}{Sean Ma},
  \bibinfo{person}{Zhiheng Huang}, \bibinfo{person}{Andrej Karpathy},
  \bibinfo{person}{Aditya Khosla}, \bibinfo{person}{Michael Bernstein},
  \bibinfo{person}{Alexander~C. Berg}, {and} \bibinfo{person}{Li Fei-Fei}.}
  \bibinfo{year}{2015}\natexlab{}.
\newblock \showarticletitle{{ImageNet Large Scale Visual Recognition
  Challenge}}.
\newblock \bibinfo{journal}{{\em International Journal of Computer Vision\/}}
  (\bibinfo{year}{2015}).
\newblock


\bibitem[\protect\citeauthoryear{Shaham, Yamada, and Negahban}{Shaham
  et~al\mbox{.}}{2015}]%
        {shaham2015understanding}
\bibfield{author}{\bibinfo{person}{Uri Shaham}, \bibinfo{person}{Yutaro
  Yamada}, {and} \bibinfo{person}{Sahand Negahban}.}
  \bibinfo{year}{2015}\natexlab{}.
\newblock \showarticletitle{Understanding adversarial training: Increasing
  local stability of neural nets through robust optimization}.
\newblock \bibinfo{journal}{{\em arXiv preprint arXiv:1511.05432\/}}
  (\bibinfo{year}{2015}).
\newblock


\bibitem[\protect\citeauthoryear{Sharif, Bhagavatula, Bauer, and Reiter}{Sharif
  et~al\mbox{.}}{2016}]%
        {adversarial:ccs16}
\bibfield{author}{\bibinfo{person}{Mahmood Sharif}, \bibinfo{person}{Sruti
  Bhagavatula}, \bibinfo{person}{Lujo Bauer}, {and} \bibinfo{person}{Michael~K.
  Reiter}.} \bibinfo{year}{2016}\natexlab{}.
\newblock \showarticletitle{Accessorize to a crime: {R}eal and stealthy attacks
  on state-of-the-art face recognition}. In \bibinfo{booktitle}{{\em
  Proceedings of the 23rd ACM SIGSAC Conference on Computer and Communications
  Security}}.
\newblock


\bibitem[\protect\citeauthoryear{Silver, Huang, Maddison, Guez, Sifre, van~den
  Driessche, Schrittwieser, Antonoglou, Panneershelvam, Lanctot, Dieleman,
  Grewe, Nham, Kalchbrenner, Sutskever, Lillicrap, Leach, Kavukcuoglu, Graepel,
  and Hassabis}{Silver et~al\mbox{.}}{2016}]%
        {silver2016mastering}
\bibfield{author}{\bibinfo{person}{David Silver}, \bibinfo{person}{Aja Huang},
  \bibinfo{person}{Christopher~J. Maddison}, \bibinfo{person}{Arthur Guez},
  \bibinfo{person}{Laurent Sifre}, \bibinfo{person}{George van~den Driessche},
  \bibinfo{person}{Julian Schrittwieser}, \bibinfo{person}{Ioannis Antonoglou},
  \bibinfo{person}{Veda Panneershelvam}, \bibinfo{person}{Marc Lanctot},
  \bibinfo{person}{Sander Dieleman}, \bibinfo{person}{Dominik Grewe},
  \bibinfo{person}{John Nham}, \bibinfo{person}{Nal Kalchbrenner},
  \bibinfo{person}{Ilya Sutskever}, \bibinfo{person}{Timothy Lillicrap},
  \bibinfo{person}{Madeleine Leach}, \bibinfo{person}{Koray Kavukcuoglu},
  \bibinfo{person}{Thore Graepel}, {and} \bibinfo{person}{Demis Hassabis}.}
  \bibinfo{year}{2016}\natexlab{}.
\newblock \showarticletitle{Mastering the game of Go with deep neural networks
  and tree search}.
\newblock \bibinfo{journal}{{\em Nature\/}} (\bibinfo{year}{2016}).
\newblock


\bibitem[\protect\citeauthoryear{Simonyan, Vedaldi, and Zisserman}{Simonyan
  et~al\mbox{.}}{2013}]%
        {simonyan2013deep}
\bibfield{author}{\bibinfo{person}{Karen Simonyan}, \bibinfo{person}{Andrea
  Vedaldi}, {and} \bibinfo{person}{Andrew Zisserman}.}
  \bibinfo{year}{2013}\natexlab{}.
\newblock \showarticletitle{Deep inside convolutional networks: Visualising
  image classification models and saliency maps}.
\newblock \bibinfo{journal}{{\em arXiv preprint arXiv:1312.6034\/}}
  (\bibinfo{year}{2013}).
\newblock


\bibitem[\protect\citeauthoryear{Simonyan and Zisserman}{Simonyan and
  Zisserman}{2014}]%
        {simonyan2014very}
\bibfield{author}{\bibinfo{person}{Karen Simonyan} {and}
  \bibinfo{person}{Andrew Zisserman}.} \bibinfo{year}{2014}\natexlab{}.
\newblock \showarticletitle{Very deep convolutional networks for large-scale
  image recognition}.
\newblock \bibinfo{journal}{{\em arXiv preprint arXiv:1409.1556\/}}
  (\bibinfo{year}{2014}).
\newblock


\bibitem[\protect\citeauthoryear{Sivakorn, Argyros, Pei, Keromytis, and
  Jana}{Sivakorn et~al\mbox{.}}{2017}]%
        {jana:hvlearn}
\bibfield{author}{\bibinfo{person}{Suphannee Sivakorn}, \bibinfo{person}{George
  Argyros}, \bibinfo{person}{Kexin Pei}, \bibinfo{person}{Angelos~D.
  Keromytis}, {and} \bibinfo{person}{Suman Jana}.}
  \bibinfo{year}{2017}\natexlab{}.
\newblock \showarticletitle{{HVLearn: Automated black-box analysis of hostname
  verification in SSL/TLS implementations}}. In \bibinfo{booktitle}{{\em
  Proceedings of the 38th IEEE Symposium on Security and Privacy}}.
  \bibinfo{address}{San Jose, CA}.
\newblock


\bibitem[\protect\citeauthoryear{Smutz and Stavrou}{Smutz and Stavrou}{2012}]%
        {smutz2012malicious}
\bibfield{author}{\bibinfo{person}{Charles Smutz} {and}
  \bibinfo{person}{Angelos Stavrou}.} \bibinfo{year}{2012}\natexlab{}.
\newblock \showarticletitle{Malicious PDF detection using metadata and
  structural features}. In \bibinfo{booktitle}{{\em Proceedings of the 28th
  Annual Computer Security Applications Conference}}.
\newblock


\bibitem[\protect\citeauthoryear{Spreitzenbarth, Freiling, Echtler, Schreck,
  and Hoffmann}{Spreitzenbarth et~al\mbox{.}}{2013}]%
        {spreitzenbarth2013mobile}
\bibfield{author}{\bibinfo{person}{Michael Spreitzenbarth},
  \bibinfo{person}{Felix Freiling}, \bibinfo{person}{Florian Echtler},
  \bibinfo{person}{Thomas Schreck}, {and} \bibinfo{person}{Johannes Hoffmann}.}
  \bibinfo{year}{2013}\natexlab{}.
\newblock \showarticletitle{Mobile-sandbox: having a deeper look into android
  applications}. In \bibinfo{booktitle}{{\em Proceedings of the 28th Annual ACM
  Symposium on Applied Computing}}.
\newblock


\bibitem[\protect\citeauthoryear{Srivastava, Hinton, Krizhevsky, Sutskever, and
  Salakhutdinov}{Srivastava et~al\mbox{.}}{2014}]%
        {srivastava2014dropout}
\bibfield{author}{\bibinfo{person}{Nitish Srivastava},
  \bibinfo{person}{Geoffrey~E Hinton}, \bibinfo{person}{Alex Krizhevsky},
  \bibinfo{person}{Ilya Sutskever}, {and} \bibinfo{person}{Ruslan
  Salakhutdinov}.} \bibinfo{year}{2014}\natexlab{}.
\newblock \showarticletitle{Dropout: a simple way to prevent neural networks
  from overfitting.}
\newblock \bibinfo{journal}{{\em Journal of Machine Learning Research\/}}
  (\bibinfo{year}{2014}).
\newblock


\bibitem[\protect\citeauthoryear{Szegedy, Liu, Jia, Sermanet, Reed, Anguelov,
  Erhan, Vanhoucke, and Rabinovich}{Szegedy et~al\mbox{.}}{2015}]%
        {szegedy2015going}
\bibfield{author}{\bibinfo{person}{Christian Szegedy}, \bibinfo{person}{Wei
  Liu}, \bibinfo{person}{Yangqing Jia}, \bibinfo{person}{Pierre Sermanet},
  \bibinfo{person}{Scott Reed}, \bibinfo{person}{Dragomir Anguelov},
  \bibinfo{person}{Dumitru Erhan}, \bibinfo{person}{Vincent Vanhoucke}, {and}
  \bibinfo{person}{Andrew Rabinovich}.} \bibinfo{year}{2015}\natexlab{}.
\newblock \showarticletitle{Going deeper with convolutions}. In
  \bibinfo{booktitle}{{\em Proceedings of the 28th IEEE Conference on Computer
  Vision and Pattern Recognition}}.
\newblock


\bibitem[\protect\citeauthoryear{Szegedy, Zaremba, Sutskever, Bruna, Erhan,
  Goodfellow, and Fergus}{Szegedy et~al\mbox{.}}{2014}]%
        {szegedy2013intriguing}
\bibfield{author}{\bibinfo{person}{Christian Szegedy},
  \bibinfo{person}{Wojciech Zaremba}, \bibinfo{person}{Ilya Sutskever},
  \bibinfo{person}{Joan Bruna}, \bibinfo{person}{Dumitru Erhan},
  \bibinfo{person}{Ian Goodfellow}, {and} \bibinfo{person}{Rob Fergus}.}
  \bibinfo{year}{2014}\natexlab{}.
\newblock \showarticletitle{Intriguing properties of neural networks}. In
  \bibinfo{booktitle}{{\em Proceedings of the 2nd International Conference on
  Learning Representations}}.
\newblock


\bibitem[\protect\citeauthoryear{tesla-accident}{tesla-accident}{2016}]%
        {tesla-accident}
tesla-accident \bibinfo{year}{2016}\natexlab{}.
\newblock \bibinfo{title}{Understanding the fatal Tesla accident on Autopilot
  and the NHTSA probe}.
\newblock
  \bibinfo{howpublished}{\url{https://electrek.co/2016/07/01/understanding-fatal-tesla-accident-autopilot-nhtsa-probe/}}.
    (\bibinfo{year}{2016}).
\newblock


\bibitem[\protect\citeauthoryear{Tram{\`e}r, Zhang, Juels, Reiter, and
  Ristenpart}{Tram{\`e}r et~al\mbox{.}}{2016}]%
        {197128}
\bibfield{author}{\bibinfo{person}{Florian Tram{\`e}r}, \bibinfo{person}{Fan
  Zhang}, \bibinfo{person}{Ari Juels}, \bibinfo{person}{Michael~K. Reiter},
  {and} \bibinfo{person}{Thomas Ristenpart}.} \bibinfo{year}{2016}\natexlab{}.
\newblock \showarticletitle{Stealing machine learning models via prediction
  APIs}. In \bibinfo{booktitle}{{\em Proceedings of the 25th USENIX Security
  Symposium}}.
\newblock


\bibitem[\protect\citeauthoryear{udacity-challenge}{udacity-challenge}{2016}]%
        {udacity:challenge}
udacity-challenge \bibinfo{year}{2016}\natexlab{}.
\newblock \bibinfo{title}{Using Deep Learning to Predict Steering Angles}.
\newblock
  \bibinfo{howpublished}{\url{https://github.com/udacity/self-driving-car}}.
  (\bibinfo{year}{2016}).
\newblock


\bibitem[\protect\citeauthoryear{Vapnik}{Vapnik}{1998}]%
        {vapnik1998statistical}
\bibfield{author}{\bibinfo{person}{Vladimir~Naumovich Vapnik}.}
  \bibinfo{year}{1998}\natexlab{}.
\newblock \bibinfo{booktitle}{{\em Statistical learning theory}}.
\newblock


\bibitem[\protect\citeauthoryear{virustotal}{virustotal}{2004}]%
        {virustotal}
virustotal \bibinfo{year}{2004}\natexlab{}.
\newblock \bibinfo{title}{VirusTotal, a free service that analyzes suspicious
  files and URLs and facilitates the quick detection of viruses, worms,
  trojans, and all kinds of malware}.
\newblock \bibinfo{howpublished}{\url{https://www.virustotal.com/}}.
  (\bibinfo{year}{2004}).
\newblock


\bibitem[\protect\citeauthoryear{visualize:dave}{visualize:dave}{2016}]%
        {visualize:dave}
visualize:dave \bibinfo{year}{2016}\natexlab{}.
\newblock \bibinfo{title}{Visualizations for understanding the regressed wheel
  steering angle for self driving cars}.
\newblock
  \bibinfo{howpublished}{\url{https://github.com/jacobgil/keras-steering-angle-visualizations}}.
    (\bibinfo{year}{2016}).
\newblock


\bibitem[\protect\citeauthoryear{\v{S}rndic and Laskov}{\v{S}rndic and
  Laskov}{2014}]%
        {6956565}
\bibfield{author}{\bibinfo{person}{Nedim \v{S}rndic} {and}
  \bibinfo{person}{Pavel Laskov}.} \bibinfo{year}{2014}\natexlab{}.
\newblock \showarticletitle{Practical evasion of a learning-based classifier: a
  case study}. In \bibinfo{booktitle}{{\em Proceedings of the 35th IEEE
  Symposium on Security and Privacy}}.
\newblock


\bibitem[\protect\citeauthoryear{Wang, Bovik, Sheikh, and Simoncelli}{Wang
  et~al\mbox{.}}{2004}]%
        {wang2004image}
\bibfield{author}{\bibinfo{person}{Zhou Wang}, \bibinfo{person}{Alan~C Bovik},
  \bibinfo{person}{Hamid~R Sheikh}, {and} \bibinfo{person}{Eero~P Simoncelli}.}
  \bibinfo{year}{2004}\natexlab{}.
\newblock \showarticletitle{Image quality assessment: from error visibility to
  structural similarity}.
\newblock \bibinfo{journal}{{\em IEEE Transactions on Image Processing\/}}
  (\bibinfo{year}{2004}).
\newblock


\bibitem[\protect\citeauthoryear{Witten, Frank, Hall, and Pal}{Witten
  et~al\mbox{.}}{2016}]%
        {witten2016data}
\bibfield{author}{\bibinfo{person}{Ian~H Witten}, \bibinfo{person}{Eibe Frank},
  \bibinfo{person}{Mark~A Hall}, {and} \bibinfo{person}{Christopher~J Pal}.}
  \bibinfo{year}{2016}\natexlab{}.
\newblock \bibinfo{booktitle}{{\em Data Mining: Practical machine learning
  tools and techniques}}.
\newblock \bibinfo{publisher}{Morgan Kaufmann}.
\newblock


\bibitem[\protect\citeauthoryear{Wu, Fredrikson, Jha, and Naughton}{Wu
  et~al\mbox{.}}{2016}]%
        {wu2016methodology}
\bibfield{author}{\bibinfo{person}{Xi Wu}, \bibinfo{person}{Matthew
  Fredrikson}, \bibinfo{person}{Somesh Jha}, {and} \bibinfo{person}{Jeffrey~F
  Naughton}.} \bibinfo{year}{2016}\natexlab{}.
\newblock \showarticletitle{A Methodology for Formalizing Model-Inversion
  Attacks}. In \bibinfo{booktitle}{{\em Proceedings of the 29th IEEE Computer
  Security Foundations Symposium}}.
\newblock


\bibitem[\protect\citeauthoryear{Xiong, Droppo, Huang, Seide, Seltzer, Stolcke,
  Yu, and Zweig}{Xiong et~al\mbox{.}}{2016}]%
        {xiong2016achieving}
\bibfield{author}{\bibinfo{person}{Wayne Xiong}, \bibinfo{person}{Jasha
  Droppo}, \bibinfo{person}{Xuedong Huang}, \bibinfo{person}{Frank Seide},
  \bibinfo{person}{Mike Seltzer}, \bibinfo{person}{Andreas Stolcke},
  \bibinfo{person}{Dong Yu}, {and} \bibinfo{person}{Geoffrey Zweig}.}
  \bibinfo{year}{2016}\natexlab{}.
\newblock \showarticletitle{Achieving human parity in conversational speech
  recognition}.
\newblock \bibinfo{journal}{{\em arXiv preprint arXiv:1610.05256\/}}
  (\bibinfo{year}{2016}).
\newblock


\bibitem[\protect\citeauthoryear{Xu, Evans, and Qi}{Xu et~al\mbox{.}}{2017}]%
        {xu2017feature}
\bibfield{author}{\bibinfo{person}{Weilin Xu}, \bibinfo{person}{David Evans},
  {and} \bibinfo{person}{Yanjun Qi}.} \bibinfo{year}{2017}\natexlab{}.
\newblock \showarticletitle{Feature squeezing: detecting adversarial examples
  in deep neural networks}.
\newblock \bibinfo{journal}{{\em arXiv preprint arXiv:1704.01155\/}}
  (\bibinfo{year}{2017}).
\newblock


\bibitem[\protect\citeauthoryear{Xu, Qi, and Evans}{Xu et~al\mbox{.}}{2016}]%
        {xu2016automatically}
\bibfield{author}{\bibinfo{person}{Weilin Xu}, \bibinfo{person}{Yanjun Qi},
  {and} \bibinfo{person}{David Evans}.} \bibinfo{year}{2016}\natexlab{}.
\newblock \showarticletitle{Automatically evading classifiers}. In
  \bibinfo{booktitle}{{\em Proceedings of the 23rd Network and Distributed
  Systems Symposium}}.
\newblock


\bibitem[\protect\citeauthoryear{Yang, Chen, Eide, and Regehr}{Yang
  et~al\mbox{.}}{2011}]%
        {yang2011finding}
\bibfield{author}{\bibinfo{person}{Xuejun Yang}, \bibinfo{person}{Yang Chen},
  \bibinfo{person}{Eric Eide}, {and} \bibinfo{person}{John Regehr}.}
  \bibinfo{year}{2011}\natexlab{}.
\newblock \showarticletitle{Finding and understanding bugs in C compilers}. In
  \bibinfo{booktitle}{{\em ACM SIGPLAN Notices}}.
\newblock


\bibitem[\protect\citeauthoryear{Yosinski, Clune, Fuchs, and Lipson}{Yosinski
  et~al\mbox{.}}{2015}]%
        {yosinskiunderstanding}
\bibfield{author}{\bibinfo{person}{Jason Yosinski}, \bibinfo{person}{Jeff
  Clune}, \bibinfo{person}{Thomas Fuchs}, {and} \bibinfo{person}{Hod Lipson}.}
  \bibinfo{year}{2015}\natexlab{}.
\newblock \showarticletitle{Understanding neural networks through deep
  visualization}. In \bibinfo{booktitle}{{\em 2015 ICML Workshop on Deep
  Learning}}.
\newblock


\bibitem[\protect\citeauthoryear{Yuan, Lu, Wang, and Xue}{Yuan
  et~al\mbox{.}}{2014}]%
        {yuan2014droid}
\bibfield{author}{\bibinfo{person}{Zhenlong Yuan}, \bibinfo{person}{Yongqiang
  Lu}, \bibinfo{person}{Zhaoguo Wang}, {and} \bibinfo{person}{Yibo Xue}.}
  \bibinfo{year}{2014}\natexlab{}.
\newblock \showarticletitle{Droid-sec: deep learning in android malware
  detection}. In \bibinfo{booktitle}{{\em ACM SIGCOMM Computer Communication
  Review}}.
\newblock


\bibitem[\protect\citeauthoryear{Zhang, Mu, Kuo, and Wright}{Zhang
  et~al\mbox{.}}{2013}]%
        {zhang2013toward}
\bibfield{author}{\bibinfo{person}{Yuqian Zhang}, \bibinfo{person}{Cun Mu},
  \bibinfo{person}{Han-Wen Kuo}, {and} \bibinfo{person}{John Wright}.}
  \bibinfo{year}{2013}\natexlab{}.
\newblock \showarticletitle{Toward guaranteed illumination models for
  non-convex objects}. In \bibinfo{booktitle}{{\em Proceedings of the 26th IEEE
  International Conference on Computer Vision}}. \bibinfo{pages}{937--944}.
\newblock


\bibitem[\protect\citeauthoryear{Zheng, Song, Leung, and Goodfellow}{Zheng
  et~al\mbox{.}}{2016}]%
        {zheng2016improving}
\bibfield{author}{\bibinfo{person}{Stephan Zheng}, \bibinfo{person}{Yang Song},
  \bibinfo{person}{Thomas Leung}, {and} \bibinfo{person}{Ian Goodfellow}.}
  \bibinfo{year}{2016}\natexlab{}.
\newblock \showarticletitle{Improving the robustness of deep neural networks
  via stability training}. In \bibinfo{booktitle}{{\em Proceedings of the 29th
  IEEE Conference on Computer Vision and Pattern Recognition}}.
  \bibinfo{pages}{4480--4488}.
\newblock


\end{thebibliography}

\end{document}